\documentclass[sn-nature]{sn-jnl}

\usepackage[utf8]{inputenc}
\usepackage{graphicx}%
\usepackage{multirow}%
\usepackage{amsmath,amssymb,amsfonts}%
\usepackage{amsthm}%
\usepackage{mathrsfs}%
\usepackage[title]{appendix}%
\usepackage{xcolor}%
\usepackage{textcomp}%
\usepackage{manyfoot}%
\usepackage{booktabs}%
\usepackage{algorithm}%
\usepackage{algorithmicx}%
\usepackage{algpseudocode}%
\usepackage{listings}%

\usepackage{bbm}

\usepackage{hyperref} 
\usepackage{url} 
\usepackage{outlines} 

\usepackage{amsmath,amsthm,amsfonts,amssymb} 
\usepackage{bbm} 
\usepackage{bm} 
\usepackage{dsfont} 
\usepackage{mathtools} 
\usepackage{braket} 

\usepackage{listings} 

\usepackage{csquotes} 
\usepackage[capitalize]{cleveref} 
\usepackage{nameref} 

\usepackage{float} 
\usepackage{graphicx} 
\usepackage[export]{adjustbox} 
\usepackage{svg} 
\usepackage{tikz} 
\usepackage{caption} 
\usepackage{subcaption} 

\usepackage{array,tabularx} 
\usepackage{booktabs} 
\usepackage{longtable} 
\usepackage{multirow}
\usepackage{multicol} 
\usepackage{makecell}

\usepackage{soul} 
\usepackage{indentfirst} 
\usepackage[htt]{hyphenat} 
\usepackage{bold-extra}
\usepackage{nicefrac}
\usepackage{todonotes}
\usepackage{pdflscape}
\usepackage{afterpage}

\usepackage{changes}
\definechangesauthor[name={Victor}, color=orange]{v}

\theoremstyle{thmstyleone}%
\theoremstyle{thmstyletwo}%

\theoremstyle{thmstylethree}%

\def\1{\bm{1}}

\def\vh{{\bm{h}}}

\def\vv{{\bm{v}}}

\def\vy{{\bm{y}}}

\DeclareMathAlphabet{\mathsfit}{\encodingdefault}{\sfdefault}{m}{sl}
\SetMathAlphabet{\mathsfit}{bold}{\encodingdefault}{\sfdefault}{bx}{n}

\newcommand{\R}{\mathbb{R}}
\newcommand{\K}{\mathbb{K}}
\newcommand{\Ht}{\mathcal{H}}

\newcommand{\btheta}{\boldsymbol{\theta}}

\newcommand{\D}{\mathcal{D}}

\DeclarePairedDelimiterX{\infdivx}[2]{(}{)}{%
	#1\;\delimsize\|\;#2%
}

\let\mc\mathcal                                             
                                                 % ...
\let\mb\mathbb                                                  
\let\tt\texttt                                              

\begin{document}

\title[Article Title]{Distribution-Free Conformal Joint Prediction Regions for Neural Marked Temporal Point Processes}


\author*[1]{\fnm{Victor} \sur{Dheur}}\email{victor.dheur@umons.ac.be}
\equalcont{These authors contributed equally to this work.}

\author[1]{\fnm{Tanguy} \sur{Bosser}}\email{tanguy.bosser@umons.ac.be}
\equalcont{These authors contributed equally to this work.}

\author[2]{\fnm{Rafael} \sur{Izbicki}}\email{rizbicki@ufscar.br}


\author[1]{\fnm{Souhaib} \sur{Ben Taieb}}\email{souhaib.bentaieb@umons.ac.be}


\affil*[1]{\orgdiv{Department of Computer Science}, \orgname{University of Mons}, \orgaddress{ \city{Mons}, \postcode{7000}, \country{Belgium}}}

\affil*[2]{\orgdiv{Departamento de Estatistica}, \orgname{Universidade Federal de São Carlos}, \orgaddress{\city{São Carlos}, \postcode{SP 13565-905}, \country{Brazil}}}




\abstract{

Sequences of labeled events observed at irregular intervals in continuous time are ubiquitous across various fields. Temporal Point Processes (TPPs) provide a mathematical framework for modeling these sequences, enabling inferences such as predicting the arrival time of future events and their associated label, called mark. However, due to model misspecification or lack of training data, these probabilistic models may provide a poor approximation of the true, unknown underlying process, with prediction regions extracted from them being unreliable estimates of the underlying uncertainty. This paper develops more reliable methods for uncertainty quantification in neural TPP models via the framework of conformal prediction. A primary objective is to generate a distribution-free joint prediction region for an event's arrival time and mark, with a finite-sample marginal coverage guarantee. A key challenge is to handle both a strictly positive, continuous response and a categorical response, without distributional assumptions. We first consider a simple but overly conservative approach that combines individual prediction regions for the event's arrival time and mark. Then, we introduce a more effective method based on bivariate highest density regions derived from the joint predictive density of arrival times and marks. By leveraging the dependencies between these two variables, this method excludes unlikely combinations of the two, resulting in sharper prediction regions while still attaining the pre-specified coverage level. We also explore the generation of individual univariate prediction regions for events' arrival times and marks through conformal regression and classification techniques. Moreover, we evaluate the stronger notion of conditional coverage. Finally, through extensive experimentation on both simulated and real-world datasets, we assess the validity and efficiency of these methods.

}

\keywords{temporal point processes, conformal prediction, bivariate predition region, highest density regions}



\maketitle

\clearpage
\section{Introduction \label{sec1}}

\begin{wrapfigure}{r}{0.45\textwidth}
\includegraphics[width=\linewidth]{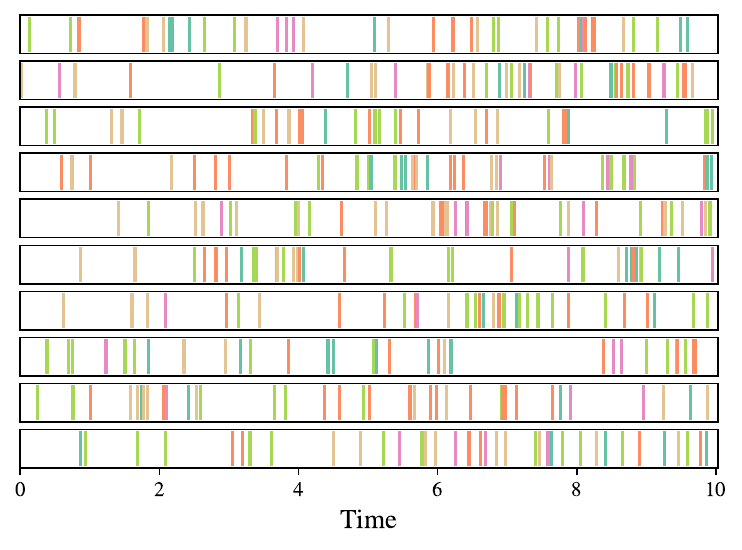} 
\caption{Examples of marked events sequences.
The vertical lines represent the arrival times with the colors representing the marks.}
\label{fig:sequence}
\end{wrapfigure}

Continuous-time event data often involve sequences of labeled events occurring at irregular intervals, with the number, timing, and mark of these events being random. This type of data is ubiquitous across various fields, ranging from healthcare \citep{EHR} and neuroscience to finance \citep{bacry2013hawkes}, social media \citep{Farajtabar}, and seismology \citep{Ogata}. In these domains, examples of event sequences include electronic health records, financial transactions, social media activities, and earthquake occurrences. An important task involves predicting not only the timing of future events based on a sequence of observed historical events but also identifying the likely associated label or type of these events, often referred to as 'marks'. Figure \ref{fig:sequence} shows an example of ten marked event sequences.

Temporal Point Processes (TPPs) provide a principled mathematical framework for modeling these event sequences. The main challenge is to learn a TPP model which effectively captures the underlying complex interactions between past event occurrences and future ones. However, classical TPP models, such as the Hawkes process \citep{Hawkes}, are often constrained by strong assumptions, which can restrict their ability to capture complex real-world event dynamics \citep{NeuralHawkes}.  To address this shortcoming, a range of neural TPPs have been developed \citep{SchurSurvey, bosser2023predictive}. These models leverage the flexibility and efficiency of neural networks with diverse temporal architectures to better capture complex event dynamics.

Using any trained probabilistic TPP model, we can derive a prediction region for the next arrival time, mark, or both, based on a sequence of observed historical events. This region should typically include a subset of potential values that are highly likely to occur, aligned with a predetermined probability coverage level. However, due to model misspecification or lack of training data, the model may provide a poor approximation of the true unknown underlying process. Consequently, prediction regions derived solely from the model's estimates may be unreliable, failing to accurately reflect the true underlying uncertainty. Building on the framework of conformal prediction (CP) \citep{Vovk2005-ib}, this paper develops more reliable methods for uncertainty quantification in neural TPP models. CP enables the construction of distribution-free prediction regions, offering a finite-sample coverage guarantee even when the base model is unreliable


Although conformal prediction has been considered in the closely related field of survival analysis \citep{candès2023conformalized, gui2023conformalized}, these studies have primarily focused on univariate survival times. To our knowledge, this study represents the first attempt to connect the field of neural TPPs to conformal prediction.

In line with the standard assumption prevalent in the neural TPP literature, we consider the setting in which a set of observed event sequences is assumed to be drawn exchangeably from the ground-truth process. Furthermore, for each event sequence, we construct an input-output pair, where the input is a neural vector representation of the event sequence history, and the output is a bivariate response representing the last event, characterized by its arrival time and mark. This aligns our approach with the scenario considered in \cite{ctsf}; however, in that context, the authors focused on regular time series forecasting. 

\begin{figure}[t!]
    \centering
    \includegraphics[width=0.7\textwidth]{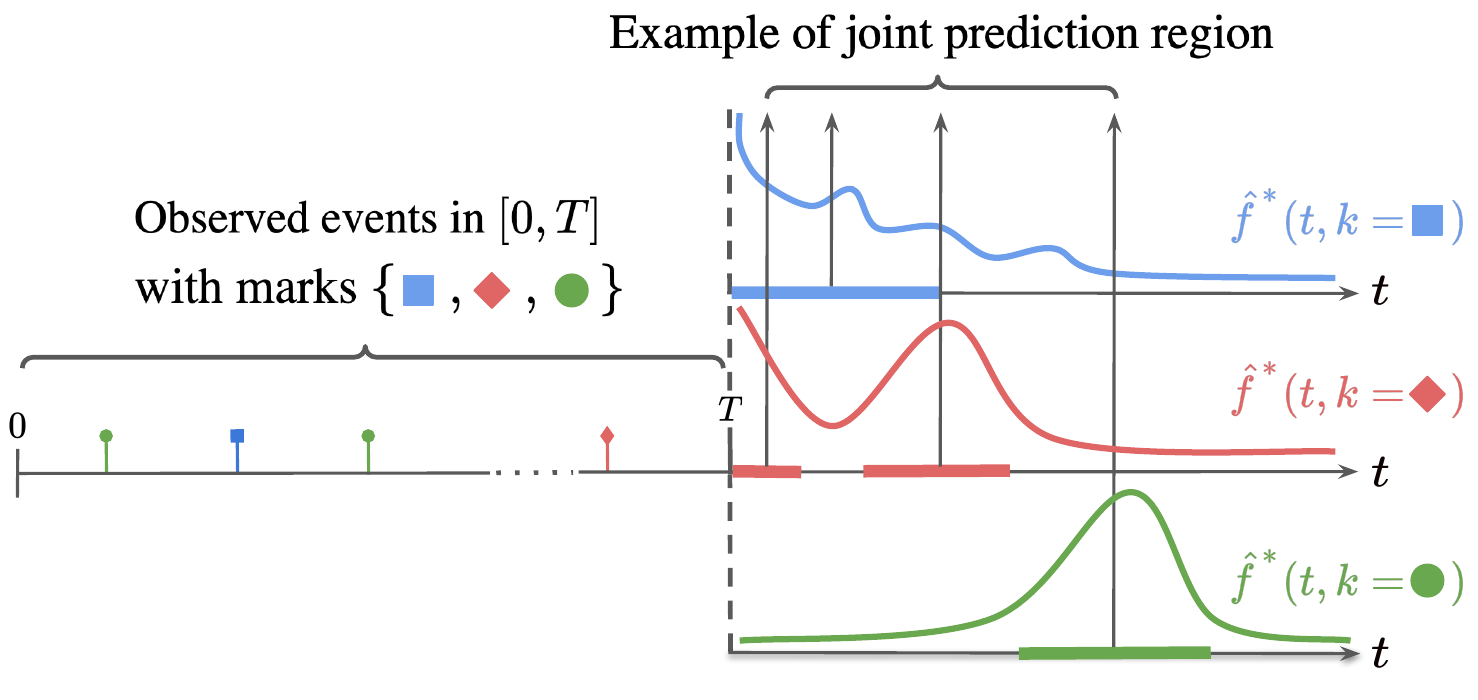}
    \caption{Toy illustration of a joint prediction region constructed from the joint density of arrival times and marks. The three colored curves represent predictive density functions while the horizontal bars represent prediction intervals.}
    \label{fig:pred_example}
\end{figure}

Our primary objective is to generate joint prediction regions for both the event arrival time and mark that are distribution-free and come with a finite-sample coverage guarantee. This entails developing a bivariate conformal prediction region, capable of accommodating both a strictly positive, continuous response and a categorical response with numerous categories, all without depending on distributional assumptions. Figure \ref{fig:pred_example} gives a toy example of such prediction region. Unfortunately, the existing literature on conformal prediction for scenarios involving multi-response or mixed response types is rather limited. Moreover, many neural TPP models typically focus on either estimating the joint density of arrival time and mark, or the conditional intensity function from which it is derived. Methods addressing these aspects are comparatively scarce. Notable contributions in the field of multi-response conformal prediction include \cite{Feldman2023-cc} and \cite{Lei2013-ol}. However, \cite{Feldman2023-cc} proposed a method centered on multi-output quantile regression for continuous random vectors, which does not easily align with our context. On the other hand, while \cite{Lei2013-ol} offers a density-based conformal method, it falls short in addressing estimation problems that involve covariates.

We will first propose a naive method that, despite its simplicity, still offers a finite-sample coverage guarantee. This approach involves combining separate prediction regions for the event arrival time and the mark. However, by neglecting potential dependencies between these variables, this method may be overly conservative. Consequently, it could lead to inflexible and large prediction regions. Such regions, while guaranteeing coverage, may not be tight, failing to accurately reflect the true underlying uncertainty. Next, we will adopt a more effective strategy that accounts for the dependencies between the arrival time and the mark. Specifically, we will construct a bivariate highest density (HDR) region \cite{Hyndman1996-wx} based on their joint predictive density. To achieve a conformal coverage guarantee, we will consider a generalization of the univariate HPD-split method \citep{Izbicki2022-ru} for bivariate responses. In contrast to the naive approach, this method has the advantage of efficiently excluding unlikely combinations of the two variables, while still maintaining the pre-specified coverage level.

Our second objective is to explore conformal prediction methods to generate univariate prediction regions, independently for the arrival time and the mark. Considering the continuous nature of the arrival time, our focus will be on conformal regression techniques. We plan to examine both symmetric and asymmetric prediction intervals using conformal quantile regression \citep{Romano2019-kp}, as well as prediction regions derived from conformal density-based methods \citep{Izbicki2022-ru}. Conversely, for the mark --- a categorical variable --- we explore conformal classification methods. Here, we intend to investigate conformal methods that involve thresholding the mark-conditional probabilities, thereby creating adaptive prediction sets \citep{Romano2020-ed, angelopoulos2022uncertainty}.

While achieving finite-sample marginal coverage is both desirable and practically feasible, we are also interested in the stronger notion of conditional coverage which requires the desired coverage level to be met conditionally. Although this is not attainable without imposing strong distributional assumptions \citep{vovk2012conditional, Foygel_Barber2021-ig}, we will also assess the conformal prediction regions in terms of approximate notions of conditional coverage.

Finally, we will evaluate the validity and efficiency of both the bivariate and univariate conformal prediction methods through an extensive series of experiments on simulated and real-world event sequence datasets. Additionally, we will explore heuristic versions of these methods, which involve substituting the model estimate in the corresponding oracle prediction region. Our evaluation will employ metrics that quantify both the probability coverage and the sharpness of the region, as determined by its length.  Our study is fully reproducible and implemented in a common code base \footnote{\url{https://github.com/tanguybosser/conf\_tpp}}.

\section{Related Work}
\label{sec:relatedwork}

\textbf{Temporal Point Processes.} Capturing the dynamics of marked events occurrences in continuous time has been extensively investigated through the framework of TPP. Early studies, such as the Hawkes \citep{Hawkes} or the self-correcting process \citep{SelfCorrecting}, focused on designing simple parametrizations of TPP models, that were successfully applied to diverse application domains \citep{Ogata, bacry2014estimation, farajtabar2014shaping, du2015timesensitive}. However, these early models often rely on strong modeling assumptions, which inherently limit their flexibility in capturing arbitrary dependencies among events occurrences \citep{NeuralHawkes}. To address these limitations, subsequent studies leveraged recent advances in deep learning, creating a new class of models called Neural TPPs \citep{SchurSurvey}. In this line of work, \citep{RMTPP, NeuralHawkes} propose to encode past event occurrences using recurrent architectures, while \citep{TransformerHawkes, SelfAttentiveHawkesProcesses, EHR, yang2022transformer} rely instead on the success of self-attention mechanisms. Regarding the TPP function being parametrized, most work traditionally focus on the MCIF, which usually requires expensive numerical integration techniques to evaluate the likelihood. To palliate this, \citep{FullyNN, EHR} instead propose to directly parametrize the cumulative MCIF, from which the MCIF can be easily retrieved through differentiation. Alternatively, \citep{IntensityFree} leverages the flexibility of a mixture of log-normals to approximate the distribution of inter-arrival times. Their work is further extended in \citep{Waghmare, bosser2023revisiting} to account for the inter-dependencies between arrival-times and marks. To learn the parameters of the model, alternatives to the NLL objective have been explored, such as reinforcement learning \citep{li2020learning, TPPReinf}, noise contrastive estimation \citep{Guo, MeiNoiseCons}, adversarial learning \citep{WassersteinTPP, WassersteinTPP2}, VAE objectives \citep{UserDep} and CRPS \citep{souhaib}. For an overview of neural TPP models, we refer the reader to the works of \citep{SchurSurvey, bosser2023predictive}. \\

\textbf{Conformal Prediction.} Our work builds upon Conformal Prediction (CP), first introduced by \cite{Vovk1999-vy}. CP is a powerful tool in machine learning for providing reliable uncertainty estimates. \cite{Angelopoulos2021-rc} offer a modern introduction, while \cite{Shafer2008-tg} present a more classical perspective. Our research specifically focuses on the split-conformal prediction method \citep{papadopoulos2002inductive}.

In the context of temporal data, CP has seen significant recent development. \cite{Gibbs2021-pj,Zaffran2022-se} proposed methods to adapt CP for sequential data shifts, continuously adjusting an internal coverage target. \cite{Stankeviciute2021-ay} extended CP to time series, and considered multi-step predictions, assuming exchangeability in individual time series. Conversely, a branch of research led by \cite{Tibshirani2019-fb,Foygel_Barber2021-ig,Xu2023-yo} challenges the exchangeability assumption by applying weighted samples. This approach, while offering stronger uncertainty estimates by leveraging similar past instances, results in a weaker conformal guarantee.

Although conformal prediction has been explored in the closely related field of survival analysis \citep{candès2023conformalized, gui2023conformalized}, these studies have primarily focused on univariate survival times.

For continuous variables, our work builds on \cite{Romano2019-kp}, which adjusts quantile regression estimates, and \cite{Izbicki2022-ru}, which outputs regions in the form of highest density regions (HDR) for univariate responses.
For discrete variables, we consider \cite{sadinle2019least}, which minimizes the average prediction set length, and \cite{Romano2020-ed,angelopoulos2022uncertainty}, which demonstrate good conditional coverage.


\textbf{Multi-response Conformal Prediction.} Our study also intersects with the field of multi-output CP. \cite{Sun2022-jb} introduced CopulaCPTS, applying CP to time series with multivariate targets and adapting the calibration set in each step based on a copula of the target variables. \cite{Feldman2023-cc} used a deep generative model to learn a unimodal representation of the response, allowing for the application of multiple-output quantile regression on this learned lower dimensional representation. This method generates flexible and informative regions in the response space, a capability not present in earlier methods. 

\section{Background on neural TPPs}
\label{sec:problem_setup}


Temporal point processes (TPPs) \citep{Daley} are stochastic processes which define a probability distribution over event sequences, serving as a valuable tool for modeling and predicting the evolution of events in continuous time. A realization of a marked TPP is a sequence $\mathcal{S} = \Set{\bm{e}_j}_{j=1}^m$ of $m$ events $\bm{e}_j=(t_j,k_j)$ where $t_j \in \R_+$ corresponds to the \emph{arrival time} and $k_j \in \K = \Set{1,...,K}$ is the associated discrete label, or \textit{mark}. The arrival times form a sequence of strictly increasing random values observed within a specified time interval $[0, T]$, i.e. $0 \leq t_1 < t_2 < \ldots < t_m \leq T$. Each arrival time is additionally associated with a random mark. Moreover, the total number of events, $m$, is also random. Alternatively, we can write $\bm{e}_j=(\tau_j,k_j)$, where $\tau_j = t_j - t_{j-1}$ corresponds to the inter-arrival time\footnote{We will use both notations interchangeably throughout the paper.}. 

For a given time $t$, we denote the counting process of mark $k \in \K$ as $N_k(t) = \sum_{j = 1}^{m} \mathbbm{1}(t_j \leq t ~\cap~ k_j = k)$. If we denote $e_{j-1}=(t_{j-1}, k_{j-1})$ the last observed event before time $t$,A marked TPP (MTPP) can be characterized by its $|\mathbb{K}|$ \textit{marked conditional intensity functions} (MCIFs) defined for $t \geq t_{j-1}$ as
\begin{equation}
    \lambda_k^\ast(t) = \lambda_k(t|\Ht_t) = \lim_{\Delta t\downarrow 0} \frac{\mathbb{E}[N_k(t + \Delta t) - N_k(t)|\Ht_t]}{\Delta t}, 
\label{eq:mcif}
\end{equation}
where $\Ht_{t} = \{(t_i,k_i) \in \mathcal{S}~|~t_i < t \}$ is the event \textit{history} until time $t$. The intensity $\lambda_k^\ast(t)$ can be interpreted as the expected occurence rate of mark-$k$ events per unit of time, conditional on $\Ht_{t}$\footnote{In ($\ref{eq:mcif}$), we employed the notation '$\ast$' of \citep{Daley} to remind us of the dependence on $\Ht_t$.}. 

\begin{figure}[h!]
    \centering
    \includegraphics[width=0.9\textwidth]{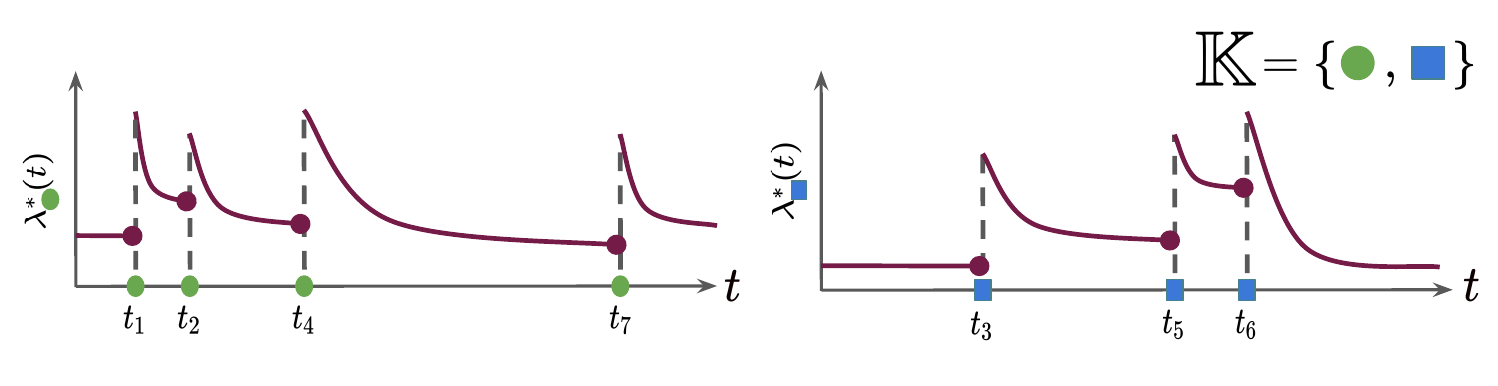}
    \caption{Illustration of a MCIF for a Hawkes process with two marks. This example highlights how events can influence the occurrence of future events. For the sake of simplicity, this illustration assumes independence between events of different marks.}
    \label{fig:hawkes_drawing}
\end{figure}

An important example of MTPP is the multivariate Hawkes process \citep{Hawkes}, whose MCIF accounts for past event influences on future ones in a positive, additive, and exponentially decaying manner:
\begin{equation}
    \lambda^\ast_k(t) = \mu_k + \sum_{k^{'}=1}^K \sum_{ (t_j,k_j) \in \Ht_{t}^{k'} } \boldsymbol{\alpha}_{k^{'}k} \boldsymbol{\beta}_{k^{'}k}e^{-\boldsymbol{\beta}_{k^{'}k}(t-t_j)}, 
\label{eq:hawkes}
\end{equation}
where $\Ht_{t}^k = \{(t_i,k_i) \in \mathcal{S}~|~t_i < t, k_j = k \}$, and in vector notation $\boldsymbol{\mu} = [\mu_1, \mu_2, \dots, \mu_K]^{T} \in \R_+^K$, $\boldsymbol{\alpha} = [\alpha_{k, k'}]$, $\boldsymbol{\beta} = [\beta_{k, k'}] \in \R_+^{K \times K}$. Figure \ref{fig:hawkes_drawing} presents an illustration of the MCIF for a Hawkes process with two marks. This figure effectively demonstrates how events can trigger the occurrence of subsequent events.

Equivalently, we can fully characterize an MTPP by the joint density of inter-arrival times and marks, denoted as $f^\ast(\tau, k)= f(\tau, k|\mathcal{H}_t)$, which can be dervied from the MCIF as follows:
\begin{align}
    f^\ast(\tau,k) &= \lambda_k^\ast(t_{j-1} + \tau)(1-F^\ast(\tau)) \label{eq:joint_CDF} \\
    &= \lambda_k^\ast(t_{j-1} + \tau) \text{exp}\left(-\sum_{k=1}^K \Lambda_k^\ast(t)\right), \label{eq:joint_Lambda}
\end{align}
where $F^\ast(\tau) = \int_{0}^\tau \sum_{k=1}^K f^\ast(s,k)ds$ is the CDF of inter-arrival times and $\Lambda_k^\ast(t) = \int_{t_{j-1}}^t \lambda_k^\ast(s)ds$ is the cumulative MCIF. 

Defining a valid MTPP model involves specifying a parametric form of $f^\ast(\tau, k; \btheta)$, $\lambda^\ast_k(t;\btheta)$ or $\Lambda^\ast_k(t;\btheta)$ with learnable parameters $\btheta$,  as long as the chosen parametrization defines a valid probability distribution over event sequences \citep{rasmussen2018lecture}. In the framework of neural TPPs \citep{SchurSurvey}, such parametrizations are computed by performing three major steps, each typically involving different neural network components. Firstly, each event $\bm{e}_j \in \mathcal{S}$ is embedded into a representation $\boldsymbol{l}_j \in \R^{d_l}$. Then, a history encoder $\text{ENC}$ generates a history embedding $\boldsymbol{h}_j $ for each $\bm{e}_j$ using its past event representations, namely $\boldsymbol{h}_j = \text{ENC}(\Set{\boldsymbol{l}_{j-1},  \boldsymbol{l}_{j-2}, ,...,\boldsymbol{l}_{\max\{1, j-p\}}}) \in \R^{d_h}$, where $p$ controls how far we wish to go back in the history. In essence, $\boldsymbol{h}_j$ acts as a neural representation of the history $\mathcal{H}_{t_j}$ of an event $\bm{e}_j$. Finally, given $\boldsymbol{h}_j$ and a time $t \geq t_{j-1}$, a decoder outputs the parameters of a function that uniquely characterizes the process, e.g. $f^\ast(\tau,k; \btheta) = f(\tau,k|\boldsymbol{h}_j; \btheta)$.

Let us consider density-based neural TPP models which factorize the joint density of the inter-arrival times and mark $f(\tau,k|\boldsymbol{h};\btheta)$ as $f(\tau,k|\boldsymbol{h};\btheta)= f(\tau|\boldsymbol{h};\btheta)  p(k|\tau,\boldsymbol{h};\btheta)$, where  $f(\tau|\boldsymbol{h};\btheta)$ is inter-arrival time PDF and $p(k|\tau,\boldsymbol{h};\btheta)$ is the mark PMF given the inter-arrival time.

These density-based models are typically trained using maximum likelihood estimation (MLE). Additionally, in the context of neural TPP research, it is common to assume that the sequences in a dataset are \textit{exchangeably} drawn from the underlying ground-truth TPP process. 
Given a dataset $\mathcal{D}^\ast = \{\mathcal{S}_1,..., \mathcal{S}_n\}$, where each sequence $\mathcal{S}_i = \Set{\bm{e}_{i,j} =(t_{i,j}, k_{i,j)})_{j=1}^{m_i}}$ comprises \(m_i\) events with arrival times observed within the interval \([0,T]\) and \(i=1,...,n\), the negative log-likelihood (NLL) writes:
\begin{equation}
    \mathcal{L}(\btheta; \mathcal{D}^\ast) = -\frac{1}{n}\sum_{i=1}^n \left[\sum_{j=1}^{m_i} \left[ \text{log }f(\tau_{i,j}|\boldsymbol{h}_{i,j};\btheta) + \text{log }p(k_{i,j}|\tau_{i,j},\boldsymbol{h}_{i,j}; \btheta)\right] + \text{log }(1 - F(T-t_{i,m_i}|\boldsymbol{h}_{i,m_i}; \btheta))\right].
\label{eq:nll}
\end{equation}

While the NLL has been largely adopted as the default scoring rule for learning (neural) MTPP models, \cite{brehmer2021comparative} showed that we can define alternative (strictly) consistent loss function for $f^\ast(\tau,k)$ by replacing the log score in (\ref{eq:nll}) with (strictly) proper scoring rules for PDFs and PMFs.

After training the neural MTTP model, the estimate $\hat{f}(\tau,k|\boldsymbol{h})$\footnote{For clarity purposes, we omit the dependency of $\hat{f}(\tau,k|\boldsymbol{h})$ on $\btheta$ for the remainder of the paper.}  can be used for prediction tasks for a new test sequence, addressing queries such as \enquote{When is the next event likely to occur?}, \enquote{What will be the type of the next event, given that it occurs at a certain time t?} or \enquote{How long until an event of type $k$ occurs?} \cite{bosser2023predictive}.

\section{Problem formulation and goals}
\label{sec:problem_formulation}

Using any trained probabilistic MTPP model, we can derive a prediction region for the next arrival time, mark, or both. This region typically represents a subset of potential values that have a high probability of occurrence. However, due to model misspecification or lack of training data, the model may provide a poor approximation of the true unknown underlying process. Consequently, prediction regions derived solely from the model's estimates may be unreliable, failing to accurately reflect the true underlying uncertainty. Building on the framework of conformal prediction (CP) \citep{Vovk2005-ib}, this paper develops more reliable methods for uncertainty quantification in neural TPP models. CP enables the construction of distribution-free prediction regions, offering a finite-sample coverage guarantee even when the base model is unreliable.

\begin{figure}[t!]
    \centering
    \includegraphics[width=0.83\textwidth,height=4.55cm]{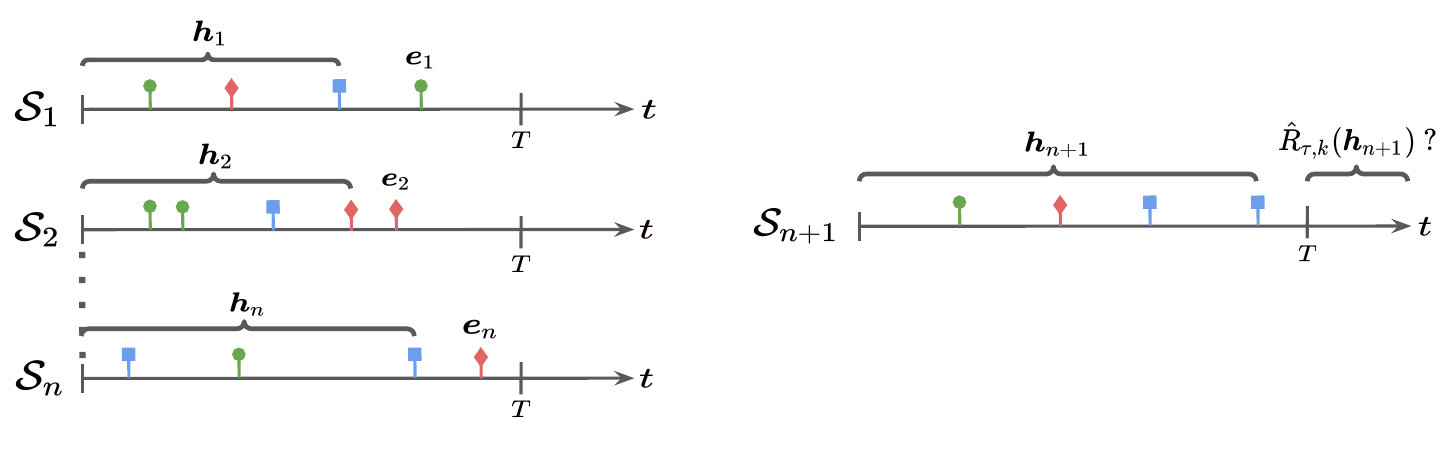}
    \caption{Illustration of the input-output pairs $\mathcal{D} = \Set{(\boldsymbol{h}_{i}, \bm{e}_{i})}_{i=1}^n$ and the joint prediction region we aim to construct for $\boldsymbol{e}_{n+1}$ given a new input $\boldsymbol{h}_{n+1}$.} 
    \label{fig:setting}
\end{figure}

Let us consider the dataset $\mathcal{D}^\ast$, which is composed of $n$ sequences assumed to be drawn exchangeably from the ground-truth MTPP process. For each sequence $ \mathcal{S}_i \in \mathcal{D}^\ast$, we define the input-output pair $(\boldsymbol{h}_{i,m_i}, \bm{e}_{i,m_i})$, where $\bm{e}_{i,m_i} = (\tau_{i,m_i}, k_{i, m_i})$ is a bivariate response corresponding to the last event in $\mathcal{S}_i$, and $\boldsymbol{h}_{i,m_i}$ is the history embedding associated to it. A similar scenario has been explored by \cite{ctsf}, but for conformal time series forecasting. From these input-output pairs, we construct the dataset $\mathcal{D} = \Set{(\boldsymbol{h}_{i,m_i}, \bm{e}_{i,m_i})}_{i=1}^n$. To simplify notation, we will henceforth denote $(\boldsymbol{h}_{i}, \bm{e}_{i}) = (\boldsymbol{h}_{i,m_i}, \bm{e}_{i,m_i})$, implying that these quantities are consistently defined for the last event of a sequence $\mathcal{S}_i$.

Given $\mathcal{D}$ and a new test input \(\boldsymbol{h}_{n+1}\), our primary goal is to construct an informative distribution-free joint prediction region $\hat{R}_{\tau,k}(\boldsymbol{h}_{n+1}) \subseteq \R_+ \times \K$ for the bivariate pair \(\bm{e}_{n+1} = (\tau_{n+1}, k_{n+1})\) of \(\boldsymbol{h}_{n+1}\). This prediction region must achieve finite-sample marginal coverage at level \(1-\alpha\), that is 
\begin{equation}
    \mathbb{P}((\tau_{n+1},k_{n+1}) \in \hat{R}_{\tau, k}(\boldsymbol{h}_{n+1})) \geq 1 - \alpha. 
\label{eq:marginal_cov_joint}
\end{equation}

Here, the probability is taken over all $n+1$ observations $\mathcal{D} \cup \{(\boldsymbol{h}_{n+1}, \boldsymbol{e}_{n+1})\}$, and the condition must hold true for any chosen values of $\alpha$ and $n$. Essentially, this entails developing a joint prediction region for a bivariate response, accommodating both a continuous and a categorical response, without relying on strong distributional assumptions. Figure \ref{fig:setting} illustrates our primary objective given a dataset $\mathcal{D} = \Set{\mathcal{S}_i}_{i=1}^n$ and a new test sequence $\mathcal{S}_{n+1}$.

We will also examine scenarios where we generate individual prediction regions for both the arrival time and the mark. Given that the arrival time is a continuous variable, we will use conformal regression techniques, while for the mark, a categorical variable, conformal classification methods will be used.

For the inter-arrival times, given $\mathcal{D}= \Set{(\boldsymbol{h}_i, \tau_i)}_{i=1}^n$ and a new test input \(\boldsymbol{h}_{n+1}\), we seek to construct a prediction region \(\hat{R}_{\tau}(\boldsymbol{h}_{n+1})  \subseteq \R^+\) for  \(\tau_{n+1}\) which achieve finite-sample marginal coverage at level \(1-\alpha\), that is 
\begin{equation}
    \mathbb{P}(\tau_{n+1} \in \hat{R}_{\tau}(\boldsymbol{h}_{n+1})) \geq 1 - \alpha.
\label{eq:marginal_cov_tau}
\end{equation} 

Similarly for the marks, given $\mathcal{D}= \Set{(\boldsymbol{h}_i, k_i)}_{i=1}^n$ and a new test input \(\boldsymbol{h}_{n+1}\), we want to generate a prediction set $\hat{R}_{k}(\boldsymbol{h}_{n+1}) \subseteq \K$ for \(k_{n+1}\) which achieve finite-sample marginal coverage at level \(1-\alpha\), that is 
\begin{equation}
    \mathbb{P}(k_{n+1} \in \hat{R}_{k}(\boldsymbol{h}_{n+1})) \geq 1 - \alpha.
\label{eq:marginal_cov_k}
\end{equation} 

Finally, beyond ensuring a finite-sample coverage guarantee, it is essential that the prediction regions are informative, which implies striving for the smallest possible region.

To accomplish this, we will adopt the split conformal prediction framework \cite{papadopoulos2002inductive}, a widely used variant of CP known for its reduced computational demands. This method, which involves partitioning the data, is relatively simple but effective in transforming any heuristic notion of uncertainty into a rigorous one \citep{Angelopoulos2021-rc}. It enables the construction of distribution-free prediction regions that achieve finite-sample coverage guarantees. We elaborate on this methodology in the following.

Consider $\mathcal{D} = \Set{(\boldsymbol{h}_i, \bm{y}_i)}_{i=1}^n$, a dataset consisting of $n$ exchangeable pairs. In the context of our problem setup, the response $\bm{y}_i \in \mathcal{Y}$ varies according to the scenario: it can be bivariate as $\bm{y}_i = \bm{e}_i$ with $\mathcal{Y} = \R_+ \times \K$, or univariate as either $\bm{y}_i = \tau_i$ or $\bm{y}_i = k_i$, with $\mathcal{Y} = \R_+$ or $\mathcal{Y}=\K$, respectively. Additionally, we have access to an MTPP model that provides a heuristic measure of uncertainty $\hat{g}$ for $\bm{y}$ given $\boldsymbol{h}$. The split conformal procedure to generate a prediction region for a new observation $\boldsymbol{y}_{n+1}$ at coverage level $1-\alpha$ can be summarized in the following steps:
\begin{enumerate}
    \item Split $\mathcal{D}$ into two non-overlapping sets, $\mathcal{D}_{\text{train}}$ and $\mathcal{D}_{\text{cal}}$ with $\mathcal{D}_{\text{train}} \cup \mathcal{D}_{\text{cal}} =\mathcal{D}$.
    \item Fit the MTPP model to the observations in $\mathcal{D}_{\text{train}}$, yielding a heuristic measure of uncertainty $\hat{g}$ for $\bm{y}$ given $\boldsymbol{h}$. 
    \item Use $\hat g$ to define a non-conformity score function $s\left(\boldsymbol{h}, \bm{y}\right) \in \R$ that assigns larger value to worse agreement between $\boldsymbol{h}$ and $\bm{y}$. 
    \item Compute the calibration scores using the observations in $\mathcal{D}_{\text{cal}}$, i.e. $\Set{s_i}_{i=1}^{|\mathcal{D}_{\text{cal}}|} := \Set{s\left(\boldsymbol{h}, \bm{y}\right):(\boldsymbol{h}, \bm{y}) \in \mathcal{D}_{\text{cal}}}$ 
    \item Compute the $1-\alpha$ empirical quantile of these calibration scores:
    \begin{equation}
    \hat{q} = \text{Quantile}\left(s_{1},...,s_{|\mathcal{D}_{\text{cal}}|} \cup \Set{\infty}; \frac{\lceil (|\mathcal{D}_{\text{cal}}| + 1)(1-\alpha)\rceil}{|\mathcal{D}_{\text{cal}}|}\right).
    \label{eq:qhat}
    \end{equation}
    \item For a new test input $\boldsymbol{h}_{n+1}$, use $\hat{q}$ to construct a prediction region for $\bm{y}_{n+1}$  with a $1 - \alpha$ coverage level as follows:
    \begin{equation}    
        \hat{R}_{\bm{y}}(\boldsymbol{h}_{n+1}) = \Set{\bm{y} \in \mathcal{Y}: s(\boldsymbol{h}_{n+1}, \bm{y}) \leq \hat{q}}. 
    \label{eq:pred_region}
    \end{equation}

\end{enumerate}

By the quantile lemma, we can write
\begin{equation}
    \mathbb{P}(\bm{y}_{n+1} \in \hat{R}_{\bm{y}}(\boldsymbol{h}_{n+1})) = \mathbb{P}(s(\boldsymbol{h}_{n+1}, \tau_{n+1}) \leq \hat{q}) \geq 1 - \alpha,
\end{equation}
In other words, the marginal coverage guarantees in \eqref{eq:marginal_cov_joint}, \eqref{eq:marginal_cov_tau} and \eqref{eq:marginal_cov_k} are satisfied. Moreover, if no ties between the scores occur with probability one, we can further show that this marginal coverage is upper bounded, i.e. 
\begin{equation}
    1 - \alpha \leq \mathbb{P}(\bm{y}_{n+1} \in \hat{R}_{\bm{y}}(\boldsymbol{h}_{n+1})) \leq 1 -\alpha + \frac{1}{|\mathcal{D}_{\text{cal}}|+1}. 
\end{equation}

Finally, while marginal coverage is a desirable and practically achievable property, we are also interested in the stronger notion of conditional coverage:
\begin{equation}
    \mathbb{P}(\bm{y}_{n+1} \in \hat{R}_{\bm{y}}(\boldsymbol{h}_{n+1}) \mid \boldsymbol{h}_{n+1}) \geq 1 - \alpha \quad \forall~\boldsymbol{h}_{n+1},
\label{eq:cond_cov}
\end{equation}
which requires the desired coverage level $1-\alpha$ to be met for all $\boldsymbol{h}_{n+1}$. Despite (\ref{eq:cond_cov}) not being achievable without strong distributional assumptions \citep{vovk2012conditional, Foygel_Barber2021-ig}, we still aspire for the prediction regions to achieve approximate notions of conditional coverage. To meet such desiderata, \cref{sec:individual_time} and \cref{sec:individual_mark} explore conformal scores to achieve the guarantees in \eqref{eq:marginal_cov_tau} and \eqref{eq:marginal_cov_k}, respectively. Similarly, \cref{sec:joint} seeks conformal scores to attain the joint guarantee in \eqref{eq:marginal_cov_joint}.


\section{Individual prediction regions for arrival times and marks}
\label{sec:individual}

In this section, we outline the methods for generating individual prediction regions $\hat{R}_\tau({\boldsymbol{h}_{n+1}})$ and $\hat{R}_k({\boldsymbol{h}_{n+1}})$ for $\tau_{n+1}$ and $k_{n+1}$, respectively. As $\tau_{n+1}$ is a continuous variable, we rely on conformal regression techniques to construct $\hat{R}_\tau({\boldsymbol{h}_{n+1}})$. Conversely, $k_{n+1}$ being a categorical variable, we leverage conformal classification approaches to build $\hat{R}_k({\boldsymbol{h}_{n+1}})$.

\subsection{Constructing a prediction region for the arrival time}
\label{sec:individual_time}

Using a dataset $\mathcal{D}= \Set{(\boldsymbol{h}_i, \tau_i)}_{i=1}^n$, our objective is to construct a prediction region \(\hat{R}_{\tau}(\boldsymbol{h}_{n+1})  \subseteq \R^+\) for the arrival time \(\tau_{n+1}\) of a new test input \(\boldsymbol{h}_{n+1}\). This prediction region must achieve finite-sample marginal coverage at level \(1-\alpha\), as given in \eqref{eq:marginal_cov_tau}. An intuitive approach is to create an equal-tailed prediction interval using conditional quantiles at levels $\nicefrac{\alpha}{2}$ and $1 - \nicefrac{\alpha}{2}$. Let $\hat{Q}_\tau(\cdot | \boldsymbol{h})$ be the predictive quantile function of $\tau$ given $\boldsymbol{h}$ trained using $\mathcal{D}$. We can define a symmetric prediction interval for $\tau_{n+1}$ as:
\begin{equation}
    \hat{R}_\tau(\boldsymbol{h}_{n+1}) = [\hat{Q}_\tau(\nicefrac{\alpha}{2}| \boldsymbol{h}_{n+1} ),\hat{Q}_\tau(1-\nicefrac{\alpha}{2} | \boldsymbol{h}_{n+1} )],
\label{eq:oracle_CQR}
\end{equation}
However, as previously mentioned, there is no guarantee that the estimate $\hat{Q}_{\tau}(\cdot|\boldsymbol{h}_{n+1})$ is a good approximation of the true $Q^*_{\tau}(\cdot|\boldsymbol{h}_{n+1})$, resulting in no finite-sample coverage guarantee. By adjusting \eqref{eq:oracle_CQR}, Conformalized Quantile Regression (C-QR) can provide a symmetric prediction interval with a finite-sample coverage guarantee (see Theorem 2 in \citep{Romano2019-kp}). For a symmetric prediction interval given by \eqref{eq:oracle_CQR}, the CQR nonconformity score can be defined as
\begin{equation}
    s_{\text{CQR}}(\boldsymbol{h}, \tau) = \text{max}\left(\hat{Q}_\tau(\nicefrac{\alpha}{2} | \boldsymbol{h} ) - \tau, \tau - \hat{Q}_\tau(1-\nicefrac{\alpha}{2}| \boldsymbol{h} )\right), 
\label{eq:score_cqr}
\end{equation}
and accounts for both potential undercoverage and overcoverage from the model. Indeed, the further $\tau$ falls \textit{outside} of the interval $\hat{R}_\tau(\boldsymbol{h}_{n+1})$ in (\ref{eq:oracle_CQR}), the greater is the positive value of $s_{\text{CQR}}$. Conversely, $s_{\text{CQR}}$ decreases the further $\tau$ correctly falls \textit{within} the interval $\hat{R}_\tau(\boldsymbol{h}_{n+1})$. After evaluating $s_{\text{CQR}}$ on hold-out calibration samples and computing $\hat{q}$ using (\ref{eq:qhat}), we construct a valid prediction interval for $\tau_{n+1}$ as:
\begin{equation}
    \hat{R}_{\tau,\text{CQR}}(\boldsymbol{h}_{n+1}) = [\hat{Q}_\tau(\nicefrac{\alpha}{2}  |\boldsymbol{h}_{n+1} )-\hat{q},\hat{Q}_\tau(1-\nicefrac{\alpha}{2} |\boldsymbol{h}_{n+1} )+ \hat{q}], 
\label{eq:R_CQR}
\end{equation}
which satisfies marginal coverage at level $1-\alpha$ since
\begin{equation}
    \mathbb{P}(\tau_{n+1} \in \hat{R}_\tau(\boldsymbol{h}_{n+1})) = \mathbb{P}(s_{\text{CQR}}(\boldsymbol{h}_{n+1}, \tau_{n+1}) \leq \hat{q}) \geq 1 - \alpha.
\end{equation}
However, since the strictly positive arrival times often show a skewed distribution with a significant concentration of probability mass close to 0, this method would not encompass the high-density region between levels 0 and $\nicefrac{\alpha}{2}$, and thus would lead to unnecessarily large intervals. Therefore, a more effective strategy involves generating an asymmetric interval extending from level 0 to $\alpha$, where the lower bound of the interval remains fixed at 0, and the upper bound, or the right tail, is independently adjusted. This translates into the following asymmetric prediction interval for $\tau_{n+1}$:
\begin{equation}
    \hat{R}_\tau(\boldsymbol{h}_{n+1}) = [0,\hat{Q}_\tau(1-\alpha | \boldsymbol{h}_{n+1} )],
\label{eq:oracle_CQRL}
\end{equation}
for which we define a Conformalized Quantile Regression Left (C-QRL) nonconformity score, expressed as: 
\begin{equation}
    s_{\text{CQRL}}(\boldsymbol{h}, \tau) = \tau - \hat{Q}_\tau(1-\alpha | \boldsymbol{h} ). 
\label{eq:score_cqrl}
\end{equation}
Naturally, this score inherits the same interpretation as the one of C-QR, and leads to the following asymmetric prediction region after estimating $\hat{q}$ on the calibration samples: 
\begin{equation}
    \hat{R}_{\tau, \text{CQRL}}(\boldsymbol{h}_{n+1}) = [0,\hat{Q}_\tau(1-\alpha |\boldsymbol{h}_{n+1} )+ \hat{q}], 
\label{eq:R_CQRL}
\end{equation}
Additionally, it is worth noting that both $s_{\text{CQR}}$ and $s_{\text{CQRL}}$ can be directly computed from the cumulative MCIF:                  
\begin{equation}
    \hat{Q}_\tau(\alpha | \boldsymbol{h} ) = \hat{\Lambda}^{-1}\left(-\text{log }(1-\alpha)|\boldsymbol{h}\right) - t_{j-1}, 
\end{equation}
where we reused the notations of Section \ref{sec:problem_setup} to indicate that $t_{j-1}$ is the last observed event arrival time in $\mathcal{H}_t$. Moreover, 
if the estimators of the quantiles are consistent (that is, they converge to the true conditional quantiles as the sample size increases), C-QR and C-QRL have asymptotic conditional coverage, and therefore (\ref{eq:cond_cov}) will hold approximately if $n$ is large \citep[Corolary 1]{sesia2020comparison}. As alternatives to C-QR and C-QRL for constructing prediction intervals for $\tau_{n+1}$, one could instead leverage the approaches of Conformal Histogram Regression (CHR) \citep{Sesia2021-tn} or HPD-Split \citep{Izbicki2022-ru}. While we expand further on HPD-split in a later section, we leave CHR as inquiry for future work.

\subsection{Constructing a prediction set for the mark} 
\label{sec:individual_mark}



Using a dataset $\mathcal{D}= \Set{(\boldsymbol{h}_i, k_i)}_{i=1}^n$, our objective is to construct a prediction set \(\hat{R}_{k}(\boldsymbol{h}_{n+1}) \subseteq \K\) for the mark \(k_{n+1}\) of a new test input \(\boldsymbol{h}_{n+1}\). This prediction set must achieve finite-sample marginal coverage at level \(1-\alpha\), as given in \eqref{eq:marginal_cov_k}.

Let $\hat{p}(\cdot|\boldsymbol{h})$ denote the predictive PMF of $k$ given $\boldsymbol{h}$, trained using $\mathcal{D}$. To generate a prediction set for $k_{n+1}$, a simple method involves ranking the marks in descending order by their associated conditional probabilities and retaining those marks where the cumulative sum of these probabilities is less than or equal to the pre-specified probability coverage. However, as previously mentioned, there is no guarantee that we have a good approximation of the true conditional probabilities $p(\cdot|\boldsymbol{h})$.

Furthermore, MTPPs often involve a large number of marks, for example, up to $|\mathbb{K}| = 50$ in our experiments, yet in practice, only a few of these marks hold significant probability. Identifying and focusing on these high-probability marks is essential as it leads to more informative prediction sets. This is the rationale behind the method of conformal Adaptive Prediction Sets (APS) \citep{Romano2020-ed}. We focus on the more recent method of Regularized Adaptive Prediction Sets (RAPS) \citep{angelopoulos2022uncertainty}, which consistently generates prediction sets of smaller size than APS by introducing regularization.

Specifically, given $\hat{p}(\cdot|\boldsymbol{h})$, RAPS defines the following nonconformity score:
\begin{equation}
    s_{\text{RAPS}}(\boldsymbol{h}, k) = \sum_{{k':\hat{p}(k'|\boldsymbol{h}) \geq \hat{p}(k|\boldsymbol{h})}} \hat{p}(k'|\boldsymbol{h}) + u \cdot \hat{p}(k|\boldsymbol{h}) + \gamma \left(o(k)-k_{\text{reg}}\right)^+,
\label{eq:raps}
\end{equation}
where $u$ is a uniform random variable handling discrete jumps in the cumulative sum of $\hat{p}(k|\boldsymbol{h})$, and $o(k) = |\Set{k' \in \mathbb{K}: \hat{p}(k'|\boldsymbol{h}) \geq \hat{p}(k|\boldsymbol{h})}|$ is the ranking of the observed mark $k$ among the probabilities in $\hat{p}(\cdot|\boldsymbol{h})$. In (\ref{eq:raps}), $(x)^+$ further denotes the positive part of $x$, and $\gamma, k_\text{reg} \geq 0$ are regularization parameters that help promote smaller set sizes compared to the ones generated by APS. The nonconformity score of APS can be easily recovered by setting $\gamma=0$ in (\ref{eq:raps}). Minus the randomization and regularization terms, the RAPS score essentially computes the cumulative sum of mark probabilities that are greater or equal to the probability of the observed ground-truth mark $k$.  If we had knowledge of the true PMF, we could construct a prediction set for $k_{n+1}$ as  
\begin{equation}
    \hat{R}_k(\boldsymbol{h}_{n+1})  = \Set{k' \in \mathbb{K}: s_{\text{RAPS}}(\boldsymbol{h}_{n+1}, k') \leq 1- \alpha},
\label{eq:oracle_RAPS1}
\end{equation}
that would meet the required marginal coverage of $1-\alpha$. Instead, the split conformal procedure introduced in Section \ref{sec:problem_setup} first computes the RAPS scores on a hold-out calibration set $\mathcal{D}_{cal}$. Then, having computed the adjusted $1-\alpha$ quantile $\hat{q}$ for these scores from (\ref{eq:qhat}), we construct the following prediction set for $k_{n+1}$:
\begin{equation}
    \hat{R}_k(\boldsymbol{h}_{n+1}) = \Set{k' \in \mathbb{K}: s_{\text{RAPS}}(\boldsymbol{h}_{n+1}, k') \leq \hat{q}},
\label{eq:r_k}
\end{equation}
which satisfies the desired marginal coverage guarantee at level $1-\alpha$ since
\begin{equation}
    \mathbb{P}(k_{n+1} \in \hat{R}_k(\boldsymbol{h}_{n+1})) = \mathbb{P}(s_{\text{RAPS}}(\boldsymbol{h}_{n+1}, k_{n+1}) \leq \hat{q}) \geq 1 - \alpha. 
\end{equation}
Finally, it is worth noting that the APS/RAPS scores can be derived from the MCIF by recovering the marginal conditional probabilities using the following definition:
\[
\hat{p}(k|\boldsymbol{h}) = \mathbb{E}_\tau\left[\hat{p}(k|\tau,\boldsymbol{h})\right] = \mathbb{E}_\tau\left[\frac{\hat{\lambda}_{k}(t_{j-1} + \tau|\bm{h})}{\hat{\lambda}(t_{j-1} + \tau|\bm{h})}\right]. 
\]

\section{Joint prediction regions for the arrival times and marks}
\label{sec:joint}

Working with a dataset $\mathcal{D}= \Set{(\boldsymbol{h}_i, \bm{e}_i)}_{i=1}^n$ where $\bm{e}_i = (\tau_i, k_i)$, our aim is to construct an informative, distribution-free bivariate joint prediction region $\hat{R}_{\tau,k}(\boldsymbol{h}_{n+1}) \in \R^+ \times \K$ for the pair $(\tau_{n+1}, k_{n+1})$ associated with a new test input $\boldsymbol{h}_{n+1}$. This prediction region should satisfy finite-sample marginal coverage at level \(1-\alpha\), as given by \eqref{eq:marginal_cov_joint}. Essentially, this involves generating a joint prediction region for a bivariate response, which integrates a continuous and a categorical variable, without relying on distributional assumptions.

In the following section, we will first explore a naive yet statistically sound method that combines individual prediction regions for the event arrival time and the mark, as outlined in Section \ref{sec:individual}. However, by neglecting potential dependencies between these variables, this method can be overly conservative, resulting in large prediction regions which would not reflect the true underlying uncertainty. A more effective strategy involves jointly predicting the event arrival time and the mark, which will better reflect the true distribution of these two variables. The associated joint prediction region can then exclude unlikely combinations of the two, while still attaining the pre-specified coverage level.

As discussed in Section \ref{sec:relatedwork}, the body of literature on conformal prediction for multi-response scenarios is limited, with notable contributions including \cite{Feldman2023-cc} and \cite{Lei2013-ol}. However, \cite{Feldman2023-cc} propose a method centered on multi-output quantile regression for continuous random vectors, which is not easily applicable in our context. Additionally, while \cite{Lei2013-ol} does present a density-based conformal method, it is not suited for estimation problems involving covariates. Instead, we explore an adaptation of the univariate HPD-split method \citep{Izbicki2022-ru} for bivariate responses. This method enables us to construct highest density regions using the joint predictive density of event arrival time and mark.

\subsection{Combining individual conformal prediction regions}
\label{sec:naive}

Let $\hat{R}_{\tau}(\boldsymbol{h}_{n+1}) \subseteq \R^+$ and $\hat{R}_k(\boldsymbol{h}_{n+1}) \subseteq \K$ represent the prediction regions for the arrival time $\tau_{n+1}$ and the mark $k_{n+1}$, respectively, for a new test input $\boldsymbol{h}_{n+1}$, as described in Section \ref{sec:individual}. Based on Bonferroni correction, the nominal coverage level for these two regions, specifically the right-hand side of equations \eqref{eq:marginal_cov_tau} and \eqref{eq:marginal_cov_k}, is set to $1-\nicefrac{\alpha}{2}$. Then, the joint prediction region $\hat{R}_{\tau,k}(\boldsymbol{h}_{n+1}) = \hat{R}_\tau(\boldsymbol{h}_{n+1}) \times \hat R_k(\boldsymbol{h}_{n+1}) \subseteq \R^+ \times \K$ for $(\tau_{n+1}, k_{n+1})$ obtained by combining these two regions has coverage at least $1 - \alpha$. In fact, by the union bound, we have:
\begin{align}
    \mathbb{P}((\tau_{n+1},k_{n+1}) \in \hat{R}_\tau(\boldsymbol{h}_{n+1}) \times \hat R_k(\boldsymbol{h}_{n+1}))
    &= \mathbb{P}(\tau_{n+1} \in \hat{R}_\tau(\boldsymbol{h}_{n+1}) \cap k_{n+1} \in \hat R_k(\boldsymbol{h}_{n+1}) ) \\
    &= 1 - \underbrace{\mathbb{P}(\tau_{n+1} \not\in \hat{R}_\tau(\boldsymbol{h}_{n+1}) \cup k_{n+1} \not\in \hat R_k(\boldsymbol{h}_{n+1}))}_{\leq \nicefrac{\alpha}{2} + \nicefrac{\alpha}{2}} \\ 
    &\geq 1 - \alpha. \label{eq:bound}
\end{align}

\begin{figure}[ht]
  \centering
  \begin{subfigure}[t]{0.56\linewidth}
    \begin{adjustbox}{valign=t}
      \includegraphics[width=\linewidth]{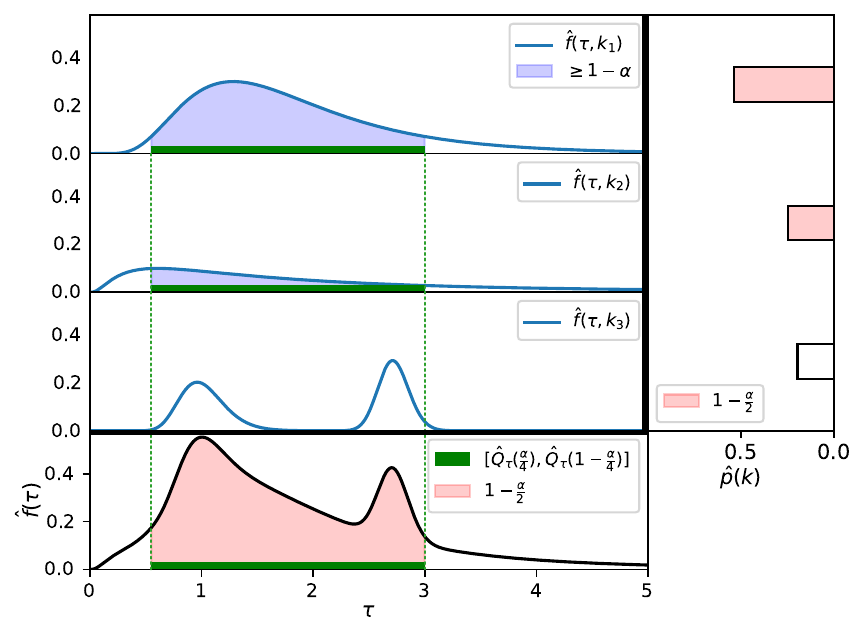}
    \end{adjustbox}
    \caption{The joint region obtained by combining individual regions, each with coverage $1 - \frac{\alpha}{2}$, has a coverage of at least $1 - \alpha$.}
    \label{fig:joint_cqr_aps}
  \end{subfigure}
  \hfill
  \begin{subfigure}[t]{0.412\linewidth}
    \begin{adjustbox}{valign=t}
      \includegraphics[width=\linewidth]{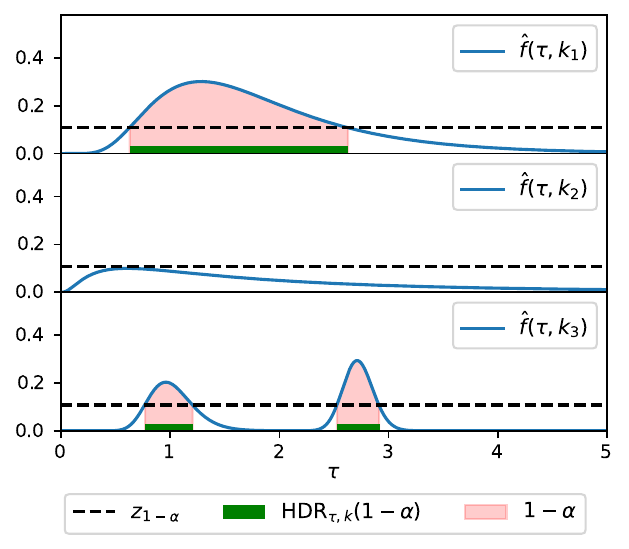}
    \end{adjustbox}
    \caption{The joint prediction region produced by HDR is composed of different regions $\hat{R}_{\tau}^{(k_1)}(\boldsymbol{h}_{n+1})$ and $\hat{R}_{\tau}^{(k_3)}(\boldsymbol{h}_{n+1})$ depending on the mark. It also demonstrates the ability to exclude marks by producing an empty region $\hat{R}_{\tau}^{(k_2)}(\boldsymbol{h}_{n+1})$.}
    \label{fig:joint_hdr}
  \end{subfigure}
  \caption{Example of joint bivariate prediction regions with $\alpha = 0.4$ on a synthetic example with $\tau \in \R^+$ and marks $\mathbb{K} = \{k_1, k_2, k_3\}$.}
  \label{fig:joint_visualization}
\end{figure}

However, this method, which treats the arrival time $\tau$ and the mark $k$ separately, can be overly conservative, resulting in large and inflexible prediction regions. Indeed, the joint prediction region generated by this approach yields equal length prediction intervals for the arrival times across all selected marks, i.e.:
\begin{equation}
\hat{R}_{\tau,k}(\boldsymbol{h}_{n+1}) =  \{(\tau', k') | \tau' \in \hat R_\tau(\boldsymbol{h}_{n+1}), k' \in \hat R_k(\boldsymbol{h}_{n+1}) \} = \{\tau' | \tau' \in \hat R_\tau(\boldsymbol{h}_{n+1}) \} \times \{k' | k' \in \hat R_k(\boldsymbol{h}_{n+1})\}.
\label{eq:indepsets}
\end{equation}
In other words, for each of the selected marks in $\hat{R}_k(\boldsymbol{h}_{n+1})$, the same prediction interval is constructed for $\tau_{n+1}$. In the following, we refer to this approach as Conformal Independent (C-IND). \cref{fig:joint_cqr_aps} shows an example of such prediction region, using CQR for the time and APS for the mark. First, the region $\hat{R}_\tau(\boldsymbol{h}_{n+1})$ is constructed for $\tau_{n+1}$ at level $1 - \frac{\alpha}{2}$ (bottom of \cref{fig:joint_cqr_aps}) and the region $\hat{R}_k(\boldsymbol{h}_{n+1})$ is constructed for $k_{n+1}$, also at level $1 - \frac{\alpha}{2}$ (right side of \cref{fig:joint_cqr_aps}). Finally, $\hat R_{\tau,k}(\boldsymbol{h}_{n+1})$ is obtained by taking the cartesian product between the two regions (middle of \cref{fig:joint_cqr_aps}).

\subsection{Conformal highest joint density regions}
\label{sec:conformal_hdr}

A better strategy for generating a joint prediction region for the arrival time and the mark involves leveraging their joint distribution. By doing so, this approach excludes unlikely combinations of the two variables, while achieving the pre-specified coverage level. To accomplish this, we propose to compute the highest density region (HDR, \cite{Hyndman1996-wx}) with a nominal coverage level of $1 - \alpha$, based on the joint density of the arrival time and mark. Let $\hat{f}(\tau,k|\boldsymbol{h}_{n+1})$ denote this predictive joint density for a new test point $\boldsymbol{h}_{n+1}$. The HDR of $\hat{f}$ with nominal coverage level $1-\alpha$ is defined as:
\begin{equation}
    \text{HDR}(1 - \alpha | \boldsymbol{h}_{n+1}) = \Set{(\tau, k)| \hat{f}(\tau,k|\boldsymbol{h}_{n+1}) \geq z_{1-\alpha}},
\label{eq:oracle_HPD}
\end{equation}
where
\begin{equation}
z_{1-\alpha} = \text{sup}\Set{z' | \mathbb{P}(\hat{f}(\tau,k|\boldsymbol{h}_{n+1}) \geq z') \geq 1 - \alpha}.
\end{equation}

It is important to highlight that in cases where the underlying distribution exhibits multimodality, an HDR approach will result in a union of intervals that, collectively, are shorter in length than a single interval with the same coverage level. Specifically, the oracle HDR has the useful property of generating the smallest possible region that guarantees conditional coverage \citep{Hyndman1996-wx}.

Moreover, in contrast to the C-IND method outlined in the previous section, an HDR approach can produce prediction regions for the arrival time that vary in length across different selected marks. Specifically, the joint HDR can be expressed as
\begin{equation}
\hat{R}_{\tau,k}(\boldsymbol{h}_{n+1}) = \text{HDR}(1 - \alpha | \boldsymbol{h}_{n+1})  = \bigcup_{k^{'} \in \hat{R}_k(\boldsymbol{h}_{n+1})} \{(\tau', k')| \tau' \in \hat R_\tau^{(k')}(\boldsymbol{h}_{n+1})\}, \label{eq:hdrsets}
\end{equation}
where
\begin{equation}
\hat{R}_k(\boldsymbol{h}_{n+1}) = \{k' | \exists~\tau \in \R^+: \hat{f}(\tau,k' | \boldsymbol{h}_{n+1}) \geq z_{1-\alpha} \}, \quad\text{and}\quad
\hat R_\tau^{(k)}(\boldsymbol{h}_{n+1}) = \{\tau' | \hat{f}(\tau',k|\boldsymbol{h}_{n+1}) \geq z_{1-\alpha} \}.\label{eq:hdr}
\end{equation}

In simpler terms, \(\hat{R}_k(\boldsymbol{h}_{n+1})\) encompasses all marks \(k \in \mathbb{K}\) for which \(\hat{f}(\tau,k|\boldsymbol{h}_{n+1})\) exceeds \(z_{1-\alpha}\) over any non-zero interval in \(\R_+\). Subsequently, for each \(k' \in \hat{R}_k(\boldsymbol{h}_{n+1})\), \(\hat{R}_{\tau}^{(k)}(\boldsymbol{h}_{n+1})\) contains the range of \(\tau\) values where the joint distribution \(\hat{f}(\tau,k|\boldsymbol{h}_{n+1})\) surpasses the threshold \(z_{1-\alpha}\). This shows that the HDR is capable of adapting to the joint distribution of the arrival time and mark, leading to more tailored and potentially more efficient prediction regions than \eqref{eq:indepsets}.

Figure \ref{fig:joint_hdr} illustrates the HDR for a simplified example with $\mathbb{K} = \{k_1, k_2, k_3\}$. The various prediction sets constituting the HDR are as follows: $\hat{R}_k(\boldsymbol{h}_{n+1}) = \{k_1, k_3\}$, referring to the selected marks; $\hat{R}_{\tau}^{(k_1)}(\boldsymbol{h}_{n+1}) = [0.8, 2.6]$ and $\hat{R}_{\tau}^{(k_3)}(\boldsymbol{h}_{n+1}) = [0.8, 1.2] \cup [2.5, 2.9]$, referring to the arrival time intervals associated to marks $k_1$ and $k_3$, respectively. Finally, $\hat{R}_{\tau}^{(k_2)}(\boldsymbol{h}_{n+1}) = \emptyset$, indicating that mark $k_2$ does not belong to the prediction region in this scenario.


Unfortunately, the heuristic joint prediction region presented in \eqref{eq:hdrsets} does not come with a finite-sample coverage guarantee. To address this, we modify the nominal coverage level $1 - \alpha$ of the HDR by using a generalization of the univariate HPD-split conformal procedure \citep{Izbicki2022-ru}. We refer to this approach as Conformal HDR (C-HDR).

Let $\hat q \in [0, 1]$. For a new test input $\boldsymbol{h}_{n+1}$, and as per the definition of HDR in \eqref{eq:oracle_HPD}, it holds that
\[
(\tau_{n+1}, k_{n+1}) \in \text{HDR}(\hat q | \boldsymbol{h}_{n+1}) \iff \text{HPD}(\tau_{n+1}, k_{n+1}|\boldsymbol{h}_{n+1}) \leq \hat q,
\]
where
\[
\text{HPD}(\tau, k|\boldsymbol{h}) = \sum_{k' \in \K} \int_{\Set{\tau'|\hat{f}(\tau',k' |\boldsymbol{h}) \geq \hat{f}(\tau,k|\boldsymbol{h})}} \hat{f}(\tau',k'|\boldsymbol{h})d\tau, 
\]
effectively calculates the probability coverage of pairs $(\tau', k')$ having a higher density than  $(\tau, k)$. The C-HDR method defines nonconformity scores based on HPD values, 
\begin{equation}
    s_{\text{HPD}}(\boldsymbol{h}, (\tau, k)) = \text{HPD}(\tau, k|\boldsymbol{h}), 
\end{equation}
and returns a joint HDR with an adjusted nominal coverage level $\hat{q}$, computed as the $1-\alpha$ empirical quantile of the $s_{\text{HPD}}$ scores evaluated on $\mathcal{D}_{\text{cal}}$, i.e.
\begin{equation}
    \hat{R}_{\tau,k}(\boldsymbol{h}_{n+1}) = \text{HDR}(\hat{q} | \boldsymbol{h}_{n+1}).
\label{eq:r_HPD}
\end{equation}

Furthermore, the quantile lemma ensures that this prediction region verifies the conformal guarantee at nominal level $1-\alpha$, i.e.
\begin{equation}
    \mathbb{P}((\tau_{n+1},k_{n+1}) \in \hat{R}_{\tau,k}(\boldsymbol{h}_{n+1})) = \mathbb{P}(s_{\text{HPD}}(\boldsymbol{h}_{n+1}, (\tau_{n+1},k_{n+1})) \leq \hat{q}) \geq 1 - \alpha.
\end{equation}

Moreover, if the estimator of $\hat f$ is consistent (that is, it converges to the true $f$ as the sample size increases), the C-HDR 
 nonconformity scores are approximately independent of the input features, which leads to
 asymptotic conditional coverage. That is,  (\ref{eq:cond_cov}) will hold approximately if $n$ is large \citep[Theorem 27]{Izbicki2022-ru}. Moreover, under these assumptions,  C-HDR will also converge to the smallest prediction region that achieves conditional coverage of $1-\alpha$ \citep[Theorem 27]{Izbicki2022-ru}.

\section{Experiments}

We assess the validity and statistical efficiency of the prediction regions produced by various CP methods, as detailed in Sections \ref{sec:individual} and \ref{sec:joint}. Our evaluation is based on five marked event sequence datasets from real-world scenarios, which have been previously considered in neural TPP research. Below, we provide a description of these datasets:

\begin{itemize}
    \item \textbf{LastFM} \citep{kumar2019predicting}: This dataset comprises records of individuals’ song-listening events over time, with each song’s artist serving as the mark.

    \item \textbf{MOOC} \citep{kumar2019predicting}: It captures the activities of students on an online course platform, where the mark denotes the specific type of activity, such as watching a video.

    \item \textbf{Reddit} \citep{kumar2019predicting}: This consists of sequences of posts made to various subreddits. Each sequence is associated with a user, and the mark is the post that the user responds to.
    
    \item \textbf{Retweets} \citep{FullyNN}: It includes sequences of retweets occurring after an initial tweet over time. Each sequence is linked to a specific tweet, with the mark being a category assigned to the retweeter (small, medium, or large retweeter).

    \item \textbf{Stack Overflow} \citep{RMTPP}: This dataset records the badges awarded to users on Stack Overflow. Each user is assigned a sequence, and the type of badge received is used as the mark.

   \item \textbf{Github} \citep{dyrep}: Records of developers' actions on the open-source platform Github. A sequence refers to a developer, and the marks correspond to the action being performed.

   \item \textbf{MIMIC2} \citep{RMTPP} Electronic health records (HER) of patients in an intensive care units for seven years. A sequence corresponds to a patient, and the marks are the types of diseases.

   \item \textbf{Wikipedia} \citep{kumar2019predicting} Records of edits made to Wikipedia pages. Each sequence is a page, and the marks refer the users that made the edits.

\end{itemize}

Following \cite{bosser2023predictive}, we preprocess each dataset to include only its 50 most frequently occurring marks. Additionally, to prevent numerical instabilities, we scaled the arrival times of events to fall within the range of $[0,10]$.  In addition to the real-world data, we also created a synthetic dataset using a multi-dimensional Hawkes process with exponential kernels \citep{Hawkes}. We set the MCIF parameters in \eqref{eq:hawkes} to align with the values used in \citep{bosser2023predictive}, as follows:
\begin{equation*}
\boldsymbol{\mu} = \begin{pmatrix}
    0.2\\
    0.6\\
    0.1\\
    0.7\\
    0.9
\end{pmatrix} \qquad 
\boldsymbol{\alpha} = \begin{pmatrix}
                    0.25 &0.13& 0.13& 0.13& 0.13\\ 0.13& 0.35& 0.13& 0.13& 0.13\\ 0.13& 0.13& 0.2& 0.13& 0.13\\
                    0.13& 0.13& 0.13& 0.3& 0.13\\ 
                    0.13& 0.13& 0.13& 0.13& 0.25
                    \end{pmatrix} \qquad 
\boldsymbol{\beta} = \begin{pmatrix}
                    4.1& 0.5& 0.5& 0.5& 0.5\\
                    0.5& 2.5& 0.5& 0.5& 0.5\\
                    0.5& 0.5& 6.2& 0.5& 0.5\\
                    0.5& 0.5& 0.5& 4.9& 0.5\\
                    0.5& 0.5& 0.5& 0.5& 4.1        
                \end{pmatrix}.
\end{equation*}
This configuration results in a marked process with $K=5$ marks. We generated five distinct Hawkes datasets, each containing 14,408 sequences, with the \textit{tick} library \citep{tick}. For all datasets, we randomly divided the sequences into training ($\mathcal{D}_{\text{train}}$), validation ($\mathcal{D}_{\text{val}}$), calibration ($\mathcal{D}_{\text{cal}}$), and test ($\mathcal{D}_{\text{test}}$) splits, in proportions of 65\%, 10\%, 15\%, and 10\%, respectively. Detailed summary statistics of the pre-processed real-world and Hawkes datasets  are presented in Table \ref{tab:datasets}. Note that the Github, MIMIC2, and Wikipedia datasets contain a short number of sequences, which will amount to a limited number of observations in the calibration and test splits.

\begin{table}[h]
\centering
\small
\setlength\tabcolsep{4pt}
\begin{tabular}{ccccccc}
& \#Seq. & \#Events & Mean Length & Max Length & Min Length & \#Marks \\
\midrule
LastFM &     856 &    193441 & 226.0 &      6396 &         2 &       50 \\
            MOOC &    7047 &    351160 &  49.8 &       416 &         2 &       50 \\
    Reddit &    4278 &    238734 &  55.8 &       941 &         2 &       50 \\
  Retweets &   12000 &   1309332 & 109.1 &       264 &        50 &        3 \\
  Stack Overflow &    7959 &    569688 &  71.6 &       735 &        40 &       22 \\
          Github &     173 &     20656 & 119.4 &      4698 &         3 &        8 \\
          MIMIC2 &     599 &      1812 &   3.0 &        32 &         2 &       43 \\
       Wikipedia &     590 &     30472 &  51.6 &      1163 &         2 &       50 \\
Hawkes &   14408 &   1056172 &  73.3 &       205 &        17 &        5 \\
\bottomrule
\end{tabular}
\caption{Real-world Datasets statistics}
\label{tab:datasets}
\end{table}


\subsection{Heuristic and conformal prediction methods}
\label{sec:predmethods}


We first focus on creating distinct univariate prediction regions for the event arrival time and the event mark of new test inputs. This is achieved through the application of conformal regression and classification techniques, as detailed in Section \ref{sec:individual}. Additionally, we explore CP methods for constructing bivariate prediction regions for both the event arrival time and its associated mark, as described in Section \ref{sec:joint}. Finally, we also consider heuristic versions, which correspond to non-conformal versions of these methods, by simply replacing the model estimate in the corresponding oracle prediction region. We provide a summary of these methods below.

\textbf{Prediction regions for the event arrival time}. We explore various methods to generate a prediction region for $\tau_{n+1}$ of a test input $\boldsymbol{h}_{n+1}$, targeting marginal coverage $1 - \alpha$.  For the heuristic methods, the first baseline is \textit{Heuristic QR (H-QR)}, which constructs a symmetric interval centered at the median:
\begin{equation}
    \hat{R}_{\tau,\text{H-QR}}(\boldsymbol{h}_{n+1}) =[\hat{Q}_\tau(\alpha|\boldsymbol{h}_{n+1}),\hat{Q}_\tau(1-\alpha|\boldsymbol{h}_{n+1})].
\end{equation}

The second baseline is \textit{Heuristic QRL (H-QRL)}, which generates an asymmetrical interval with the left bound at zero:
\begin{equation}
    \hat{R}_{\tau, \text{H-QRL}}(\boldsymbol{h}_{n+1}) = [0,\hat{Q}_\tau(1-\alpha|\boldsymbol{h}_{n+1})].
\end{equation}

The third baseline is \textit{Heuristic HDR (H-HDR)} which forms a HDR, i.e.
\begin{equation}
    \hat{R}_{\tau,\text{H-HDR}}(\boldsymbol{h}_{n+1}) = \{\tau| \hat{f}(\tau|\boldsymbol{h}_{n+1}) \geq z_{1-\alpha}\},
\end{equation}
where $z_{1-\alpha} = \text{sup}\Set{z' | \mathbb{P}(\hat{f}(\tau|\boldsymbol{h}_{n+1}) \geq z') \geq 1 - \alpha}$. Unlike the HDR defined in \eqref{eq:oracle_HPD} for a joint prediction on the time and mark, this method represents a univariate HDR, specifically focusing on the arrival time.

We also consider conformal versions of these approaches, denoted as C-QR, C-QRL, and C-HDR, with prediction regions defined in equations (\ref{eq:R_CQR}), (\ref{eq:R_CQRL}), and (\ref{eq:r_HPD}), respectively. Additionally, we analyze \textit{C-CONST}, a simple conformal baseline. Its nonconformity score is defined as $s_\text{C}(\boldsymbol{h}, \tau) = \tau$ and it generates predictions regions of the form $\hat{R}_\tau(\boldsymbol{h}_{n+1}) = [0, \hat{q}]$, independent of the model and history $\boldsymbol{h}$, where $\hat{q}$ is defined by the split-conformal prediction algorithm in \eqref{eq:qhat}.

\textbf{Prediction sets for the event mark}. We explore various methods to generate a prediction set for $k_{n+1}$ given a test input $\boldsymbol{h}_{n+1}$.
The first baseline methods, called Heuristic APS (H-APS) and Heuristic RAPS (H-RAPS), generate the following sets:
\begin{equation}
    \hat{R}_{k,\text{H-APS}}(\boldsymbol{h}_{n+1})  = \Set{k' \in \mathbb{K}: s_{\text{APS}}(\boldsymbol{h}_{n+1}, k') \leq 1- \alpha},
\label{eq:oracle_APS}
\end{equation}
and
\begin{equation}
    \hat{R}_{k,\text{H-RAPS}}(\boldsymbol{h}_{n+1})  = \Set{k' \in \mathbb{K}: s_{\text{RAPS}}(\boldsymbol{h}_{n+1}, k') \leq 1- \alpha}. 
\label{eq:oracle_RAPS}
\end{equation}
For their conformal counterparts, we derive prediction regions as detailed in (\ref{eq:r_k}) and the unregularized C-APS algorithm described in \citep{Romano2020-ed}. Recall that $s_\text{APS}$ is recovered from $s_{\text{RAPS}}$ by setting $\gamma=0$ in (\ref{eq:raps}). Additionally, we explore the C-PROB conformal baseline, introduced in \cite{sadinle2019least}. This baseline defines its nonconformity score in terms of the estimated probability mass function over the mark:
\begin{equation}
s_{\text{C-PROB}}(\boldsymbol{h}, k) = 1 - \hat{p}(k|\boldsymbol{h}),
\end{equation}
which yields the following prediction region after computing $\hat{q}$ with the split-conformal prediction algorithm (\ref{eq:qhat}):
\begin{equation}
    \hat{R}_{k,\text{C-PROB}}(\boldsymbol{h}_{n+1}) = \Set{k' \in \K: \hat{p}(k'|\boldsymbol{h}) \geq 1 - \hat{q}}. 
\label{eq:rprob}
\end{equation}
Moreover, to avoid generating empty prediction sets, the mark associated to the highest estimated probability is systematically included for all methods that we consider, namely H-APS, H-RAPS, C-PROB, C-APS and C-RAPS.

\textbf{Bivariate prediction regions for the arrival time and the associated mark}. We explore two methodologies to construct a bivariate prediction region, $\hat{R}_{\tau,k}(\boldsymbol{h}_{n+1})$, for the pair $(\tau_{n+1}, k_{n+1})$. The first method combines individual univariate prediction regions, as detailed in Section \ref{sec:naive}. For this method, we develop two variants, each based on the specific construction of $\hat{R}_\tau(\boldsymbol{h}_{n+1})$ and $\hat{R}_k(\boldsymbol{h}_{n+1})$. The first variant, C-QRL-RAPS, combines C-QRL for $\hat{R}_\tau(\boldsymbol{h}_{n+1})$ and C-RAPS for $\hat{R}_k(\boldsymbol{h}_{n+1})$. The second variant, C-HDR-RAPS, uses C-HDR for $\hat{R}_\tau(\boldsymbol{h}_{n+1})$, while still employing C-RAPS for $\hat{R}_k(\boldsymbol{h}_{n+1})$. The second method generates joint HDR regions, as described in Section \ref{sec:conformal_hdr}. In parallel, we also examine their heuristic counterparts, referred to as H-QRL-RAPS, H-HDR-RAPS, and H-HDR, respectively.

\subsection{Neural TPP models} 
\label{sec:baselines}

We present several SOTA neural TPP models designed to estimate the joint density of event arrival time and mark, represented as $\hat{f}(\tau,k|\boldsymbol{h})$. From these models, we will derive a heuristic-based measure of uncertainty. Recall that given a sequence of events $\mathcal{S}=\Set{\boldsymbol{e}_1,..., \boldsymbol{e}_m}$, these models can be essentially decomposed into three main components: an event encoder, a history 
encoder, and a decoder. To obtain an event representation $\boldsymbol{l}_j \in \R^{d_e}$ for $\boldsymbol{e}_j=(t_j,k_j)$, we proceed in two steps. First, we follow \citep{EHR} by mapping $t_j$ to a vector of sinusoidals functions \citep{EHR}: 
\begin{equation}
\boldsymbol{l}^t_{j} = \underset{s=0}{\overset{d_t/2-1}{\bigoplus}} \text{sin}~(\alpha_s t_j) \oplus \text{cos}~(\alpha_s t_j) \in \mathbb{R}^{d_t},
\label{eq:temporalencoding}
\end{equation}
where $\alpha_s \propto 1000^{\frac{-2s}{d_t}}$ and $\oplus$ is the concatenation operator. Then, a mark embedding $\boldsymbol{l}_j^k \in \mathbb{R}^{d_k}$ for $k_j$ is computed as $\boldsymbol{l}_j^k = \mathbf{E}^k \boldsymbol{k}_j$, where $\mathbf{E}^k \in \mathbb{R}^{d_k \times K}$ is a learnable embedding matrix, and $\boldsymbol{k}_j \in \{0,1\}^K$ is the one-hot encoding of $k_j$. The event representation $\boldsymbol{l}_j$ is finally obtained through concatenation, i.e. $\boldsymbol{e}_j = [\boldsymbol{e}_j^t \oplus \boldsymbol{e}_j^k]$. Subsequently, we generate the history embedding $\boldsymbol{h}_j \in \R^{d_h}$ of an event $\boldsymbol{e}_j$ by sequentially processing its set of past events representations through a GRU encoder:
\begin{equation}
    \boldsymbol{h}_j = \text{GRU}\left(\Set{\boldsymbol{l}_1,...,\boldsymbol{l}_{j-1}}\right). 
\end{equation}
Finally, given $\boldsymbol{h_j}$ and a query time $t > t_{j-1}$, we consider several decoders which computes either $\hat{f}(\tau,k|\boldsymbol{h}_j)$, $\hat{\lambda}_k(t|\boldsymbol{h}_j)$, or $\hat{\Lambda}_k(t|\boldsymbol{h}_j)$. Recall that each of these functions can be retrieved from the others through (\ref{eq:joint_Lambda}). We describe four different decoders below.

\begin{itemize}
    \item \textbf{Conditional LogNormMix} (CLNM) \citep{IntensityFree, Bosser2023-yn} models $\hat{f}(\tau, k|\boldsymbol{h}_j) = \hat{f}(\tau|\boldsymbol{h_j})\hat{p}(k|\tau,\boldsymbol{h}_j)$ with $\hat{f}(\tau|\boldsymbol{h}_j)$ being a mixture of log-normal distributions:
    \begin{equation}
    \hat{f}(\tau|\boldsymbol{h}_j) = \sum_{c=1}^C p_c \frac{1}{\tau \sigma_c \sqrt{2\pi}}\text{exp}\Big(-\frac{(\text{log}~\tau -\mu_c)^2}{2\sigma_c^2}\Big),
    \label{eq:lognormmixtime}
    \end{equation}
    where $p_c= \text{Softmax}\big(\mathbf{W}_p \boldsymbol{h}_j + \boldsymbol{b}_p\big)_c$, $\mu_c = (\mathbf{W}_\mu \boldsymbol{h}_j + \boldsymbol{b}_\mu)_c$, and $\sigma_c = \text{exp}(\mathbf{W}_\sigma \boldsymbol{h}_j + \boldsymbol{b}_\sigma)_c$ are the weight, mean and standard deviation of the $c^{th}$ mixture component, respectively. In (\ref{eq:lognormmixtime}), $\mathbf{W}_p, \mathbf{W}_\mu, \mathbf{W}_\sigma \in \mathbb{R}^{C \times d_h}$ and $\boldsymbol{b}_p, \boldsymbol{b}_\mu, \boldsymbol{b}_\sigma \in \mathbb{R}^C$ with $C$ being the number of mixture components. The mark PMF is given by 
    \begin{equation}
    \hat{p}(k|\tau, \boldsymbol{h}_j) = \sigma_S\left(\mathbf{W}_2 \sigma_R\left(\mathbf{W}_1\left[\boldsymbol{h}_j \oplus \text{log }\tau\right] + \boldsymbol{b}_1)\right)   + \boldsymbol{b}_2\right), 
    \label{eq:pmf_cond_dis}
    \end{equation}
    where $\mathbf{W}_1 \in \mathbb{R}^{d_1 \times (d_h + 1)}$, $\boldsymbol{b}_1 \in \mathbb{R}^{d_1}$, $\mathbf{W}_2 \in \mathbb{R}^{K \times d_1}$,  and $\boldsymbol{b}_2 \in \mathbb{R}^{K}$. \\
    \item \textbf{FullyNN} (FNN) \citep{FullyNN, EHR} directly parametrizes the cumulative MCIF as follows: 
   \begin{equation}
        \hat{\Lambda}_k^\ast(t) = G_k^\ast(t) - G_k^\ast(t_{j-1}),
    \end{equation}
    \begin{equation}
    G_k^\ast(t) = \sigma_{S,k}\big(\boldsymbol{w}_k^\mathsf{T}(\sigma_{GS,k}\big(\boldsymbol{w}^t \text{log }\tau + \mathbf{W}^h \boldsymbol{h}_j + \boldsymbol{b}\big) + b_k \big),
    \label{eq:fullyNN}
    \end{equation}
    where $\sigma_{S,k}$ and $\sigma_{GS,k}$ are the mark-wise Softplus and Gumbel-Softplus activation functions, respectively. In (\ref{eq:fullyNN}), $\boldsymbol{w}_k^\mathsf{T} \in \R_{+}^{K \times d_1}$, $\boldsymbol{w}^t \in \R_{+}^{d_1}$, $\mathbf{W}^h \in \R_{+}^{d_1 \times d_h}$, $\boldsymbol{b} \in \R_{+}^{d_1}$, and $b_k \in \R_{+}$.\\
    \item \textbf{Self-attentive Hawkes Process} (SAHP) \citep{SelfAttentiveHawkesProcesses} also proposes a parametrization of the MCIF given by
    \begin{equation}
        \hat{\lambda}_k^\ast(t) = \sigma_{S,k} \left(\mu_k -(\eta_k- \mu_k)\text{exp}(-\gamma_k(t-t_{j-1}))\right).  
    \label{eq:sahp}
    \end{equation}
    In the above, $\mu_k = \sigma_G(\boldsymbol{w}_{\mu,k}^\mathsf{T} \boldsymbol{h}_j)$, $\eta_k = \sigma_S(\boldsymbol{w}_{\eta,k}^\mathsf{T} \boldsymbol{h}_j)$ and $\boldsymbol{\gamma} = \sigma_G(\boldsymbol{w}_{\gamma,k}^\mathsf{T} \boldsymbol{h}_j)$ where $\sigma_G$ is the GeLU activation function \citep{hendrycks2023gelu} and $\boldsymbol{w}_{\mu,k}, \boldsymbol{w}_{\eta,k}, \boldsymbol{w}_{\gamma,k} \in \R^{d_h}$. 
\end{itemize}



Each model mentioned is trained by minimizing the average negative log-likelihood (NLL), given in (\ref{eq:nll}), across training sequences contained in $\mathcal{D}_{\text{train}}$. For optimization, we use mini-batch gradient descent with the Adam optimizer \citep{adam} and a learning rate of $\alpha=10^{-3}$. The models are trained for at most 500 epochs, and training is interrupted through an early-stopping procedure if there is no improvement in NLL on the validation dataset $\mathcal{D}_{\text{val}}$ for 100 consecutive epochs. In such instances, the model's parameters revert to the state where the validation loss was lowest.

With a trained TPP model, we are able to calculate the necessary prediction functions for computing prediction regions, as described in Section \ref{sec:predmethods}, for all test inputs within $\mathcal{D}_{\text{test}}$. For the CP methods, the non-conformity scores are computed on $\mathcal{D}_{\text{cal}}$. To compute individual prediction regions for the arrival time and the mark, it's essential to compute the predictive marginals, $\hat{f}(\tau|\boldsymbol{h})$ and $\hat{p}(k|\boldsymbol{h})$, respectively.

To derive $\hat{f}(\tau|\boldsymbol{h})$, we sum over the joint density for each mark, as follows: $\hat{f}(\tau|\boldsymbol{h}) = \sum_{k=1}^K \hat{f}(\tau,k|\boldsymbol{h})$. Meanwhile, $\hat{p}(k|\boldsymbol{h})$ is approximated through integration over the positive real line:
\begin{equation}
    \hat{p}(k|\boldsymbol{h}) = \int_{\mathbb{R^+}} \hat{f}(s,k|\boldsymbol{h}_{n+1})ds = \mathbb{E}_\tau[\hat{p}(k|\tau, \boldsymbol{h})] \simeq \frac{1}{N} \sum_{s=1}^N \hat{p}(k|\tau_s, \boldsymbol{h}), 
\end{equation}
where $N=100$ samples $\tau_s$ are generated from $\hat{f}(\tau|\boldsymbol{h})$. This sampling is achieved with the inverse transform sampling method using a binary search algorithm.

\subsection{Evaluation metrics}
\label{sec:metrics}

We assess the prediction regions $\hat{R}_{\bm{y}}(\boldsymbol{h}_i)$, generated for every input $\boldsymbol{h}_i \in \mathcal{D}_{\text{test}}$, using metrics that quantify probability coverage and the length of each region.

The empirical \textit{marginal coverage} is calculated as
\begin{equation}
    \text{MC} = \hat{\mathbb{P}}_{\mathcal{D}_{\text{test}}}\left(\bm{y}_i \in \hat{R}_{\bm{y}}(\boldsymbol{h}_i)\right) = \frac{1}{|\mathcal{D}_{\text{test}}|} \sum_{i=1}^{|\mathcal{D}_{\text{test}}|} \mathbbm{1}\left[ \bm{y}_i\in \hat{R}_{\bm{y}}(\boldsymbol{h}_{i}) \right].
\end{equation}
Essentially, this metric calculates the proportion of instances where $\hat{R}_{\bm{y}}(\boldsymbol{h}_i)$ contains $\bm{y}_i$ across all observations in $\mathcal{D}_{\text{test}}$.

The \textit{average length} of the prediction regions is computed as
\begin{equation}
    \text{Length} = \frac{1}{|\mathcal{D}_{\text{test}}|} \sum_{i=1}^{|\mathcal{D}_{\text{test}}|} |\hat{R}_{\bm{y}}(\boldsymbol{h}_i) |,
\end{equation}
where $|\cdot|$ denotes the length of a region. Specifically, for univariate prediction regions, if $\bm{y} = \tau$, $|\cdot|$ represents the length of prediction intervals or the cumulative length in the case of union of intervals. When $\bm{y} = k$, it is the cardinality of the discrete prediction set. For bivariate prediction regions, where $\bm{y} = (\tau, k)$, the calculation differs based on the method. For naive prediction regions as defined in \eqref{eq:indepsets}, the length is given by:
\[
|\hat{R}_{\tau,k}(\boldsymbol{h}_i)| = |\hat{R}_{\tau}(\boldsymbol{h}_i)|  *   |\hat{R}_{k}(\boldsymbol{h}_i)|.
\]

For the bivariate HDRs as defined in \eqref{eq:hdrsets}, the length is calculated as:
\[
|\hat{R}_{\tau,k}(\boldsymbol{h}_i)| =  \sum_{k' \in \hat{R}_{k}(\boldsymbol{h}_i))} |\hat{R}_{\tau}^{(k')}(\boldsymbol{h}_i))|.
\]

We also consider the \textit{geometric mean of the lengths} computed on $\D_{\text{test}}$ as:
\begin{equation}
    \text{G. Length} = \frac{1}{|\mathcal{D}_{\text{test}}|} \sum_{i=1}^{|\mathcal{D}_{\text{test}}|} \log (|\hat{R}_{\bm{y}}(\boldsymbol{h}_i) | + \epsilon),
\end{equation}
where $\epsilon$ is an offset that we fix at $\epsilon = 0.01$ to avoid values of $-\infty$ when $|\hat{R}_{\bm{y}}(\boldsymbol{h}_i)| = 0$.

For a better comparison, when comparing a set of $M$ conformal methods with average lengths $L_1, \dots, L_M$, we report the \textit{relative length} of the $i$th method as:
\[
\text{R. Length} = \frac{L_i}{\min_{j \in \Set{1, \dots, M}} L_j}.
\]

We also consider an approximate measure of conditional coverage using the \textit{Worst Slab Coverage} (WSC) metric, as introduced in \citep{Cauchois2021-mj}. The WSC metric evaluates the lowest coverage across all slabs $\vv \in \mathbb{R}^{d_h}$, each containing at least a proportion $\delta$ of the total mass, where $0 < \delta \leq 1$. Given $\vv \in \R^{d_h}$, $\text{WSC}_\vv$ is defined as follows:
\begin{align}
\text{WSC}_\vv &= \underset{a<b}{\inf} \left\{
 \hat{\mathbb{P}}_{\mathcal{D}_{\text{test}}}\left(\vy_i \in \hat{R}_{\bm{y}}(\boldsymbol{h}_i) |~a \leq \boldsymbol{v}^\intercal \boldsymbol{h}_i \leq b\right)  ~\text{s.t.}~~ \hat{\mathbb{P}}_{\mathcal{D}_{\text{test}}}(a \leq \boldsymbol{v}^\intercal \boldsymbol{h}_i \leq b) \geq \delta \right \},
 \label{eq:wsc} 
\end{align}
where $a, b \in \R$.
This quantity assesses the conditional coverage by conditioning on the history encodings $\vh_i$ which have a certain level of similarity with the slab $\vv$ where the similarity is measured by the dot product $\boldsymbol{v}^\intercal \boldsymbol{h}_i$.
To estimate the worst slab, we follow \cite{Cauchois2021-mj} and draw 1000 samples $\vv_j \in \R^d$ uniformly in the simplex $\mb{S}^{d-1}$ and compute the slab with minimum conditional coverage as:
\begin{equation}
    \text{WSC} = \min_{\vv_j \in \mb{S}^{d-1}} \text{WSC}_{\vv_j}.
\end{equation}
In practice, to avoid biases due to overfitting on the test dataset, we follow \cite{Romano2020-ed,Sesia2021-tn} and first divide the test set in two parts $\D_\text{test} = \D^{(1)}_\text{test} \cup \D^{(2)}_\text{test}$. Then, we compute the worst combination of $a$, $b$ and $\vv$ on $\D^{(1)}_\text{test}$ according to the minimum $\text{WSC}(\boldsymbol{h}_i)$ metric with $\delta = 0.2$, and evaluate conditional coverage on $\D^{(2)}_\text{test}$.

We further assess (approximate) conditional calibration using the input space partitioning approach from the CD-split$^+$ method detailed in \cite{Izbicki2022-ru}, which we call \textit{conditional coverage error}.
Instead of the Cramér–von Mises distance, we consider the 2-Wasserstein distance, which we estimate via samples.
Let $Z$ represent the random variable corresponding to the HPD values. The 2-Wasserstein distance, comparing two random variables with quantile function $F^{-1}_{Z}(\cdot \mid \bm{h}_a)$ and $F^{-1}_{Z}(\cdot \mid \bm{h}_b)$, is expressed as:
\begin{align}
    d_Z(\bm{h}_a, \bm{h}_b) = \left( \int_{0}^{1} \left|F^{-1}_{Z}(u \mid \bm{h}_a) - F^{-1}_{Z}(u \mid \bm{h}_b)\right|^2 du \right)^{\frac{1}{2}}.
\end{align}

In practice, we approximate this distance by generating two samples $Z^{(a)}_{1}, \dots, Z^{(a)}_m$ and $Z^{(b)}_1, \dots, Z^{(b)}_m$, each with $m$ observations conditional on $\bm{h}_a$ and $\bm{h}_b$, respectively.
Based on the observation that the order statistic $Z_{(i)}$ of a sample $Z_1, \dots, Z_m$ approximates the quantile function $F^{-1}_Z\left(\frac{i}{m+1}\right)$, we can approximate $d_Z(\bm{h}_a, \bm{h}_b)$ using $\left( \sum_{i=1}^m \left| Z^{(a)}_{(i)} - Z^{(b)}_{(i)}) \right|^2 \right)^{\frac{1}{2}}$.

Using this distance function, we calculate centroids $C_1, \dots, C_J \in \R^{d_h}$ by applying the $k$-means++ clustering algorithm on $\mc{D}_\text{cal}$.
Then, we consider a partition $\mc{A}$ of $\R^{d_h}$ defined as $\vh \in A_j$ if and only if $d_Z(\vh, C_j) < d_Z(\vh, C_k)$ for every $k \neq j$.

In \cref{sec:cond_coverage}, we show that we have to further adapt the distance function from $d_Z$ to $d_{\log Z}$ since distributions with the longest tails exhibit extreme distances from other distributions, often resulting in their isolation into small clusters.
By focusing on the random variable $\log Z$ instead of $Z$, we achieve more balanced cluster sizes, which is crucial for enhancing the accuracy of conditional coverage estimation.
In our experiments, we additionally use $J = 4$ centroids to ensure an accurate estimation of conditional coverage per cluster.

Finally, the conditional coverage error is defined as:
\begin{align}
    \text{CCE} = \frac{1}{|\D_{\text{test}}|} \sum_{i=1}^{|\D_{\text{test}}|} \sum_{j=1}^J \left(\hat{\mathbb{P}}_{\mathcal{D}_{\text{test}}}\left( \bm{y}_i \in \hat{R}_{\bm{y}}(\boldsymbol{h}_i) ~\middle|~ \boldsymbol{h}_i \in A_j\right) - (1 - \alpha)\right)^2.
\end{align}

\subsection{Results and Discussion}
\label{sec:results}

We first detail the results for individual prediction regions for the arrival time and the mark in Sections \ref{sec:results_time} and \ref{sec:results_mark}, respectively. Subsequently, the results for the joint prediction regions are presented in Section \ref{sec:results_joint}. Our primary focus is on a probability miscoverage level of $\alpha = 0.2$. Following this, we show the results at various other coverage levels in Section \ref{sec:coverage_per_level}.
In this section, we focus on the neural TPP model CLNM, and on the real-world datasets LastFM, MOOC, Retweets, Reddit, and Stack Overflow.
Additional results for other neural TPP models, as well as results on the smaller and synthetic Hawkes datasets, are provided in \cref{sec:results_other_models}
with similar conclusions.

\subsubsection{Prediction regions for the arrival time}
\label{sec:results_time}

\begin{figure}[h!]


    \centering
    \includegraphics[width=0.92\linewidth]{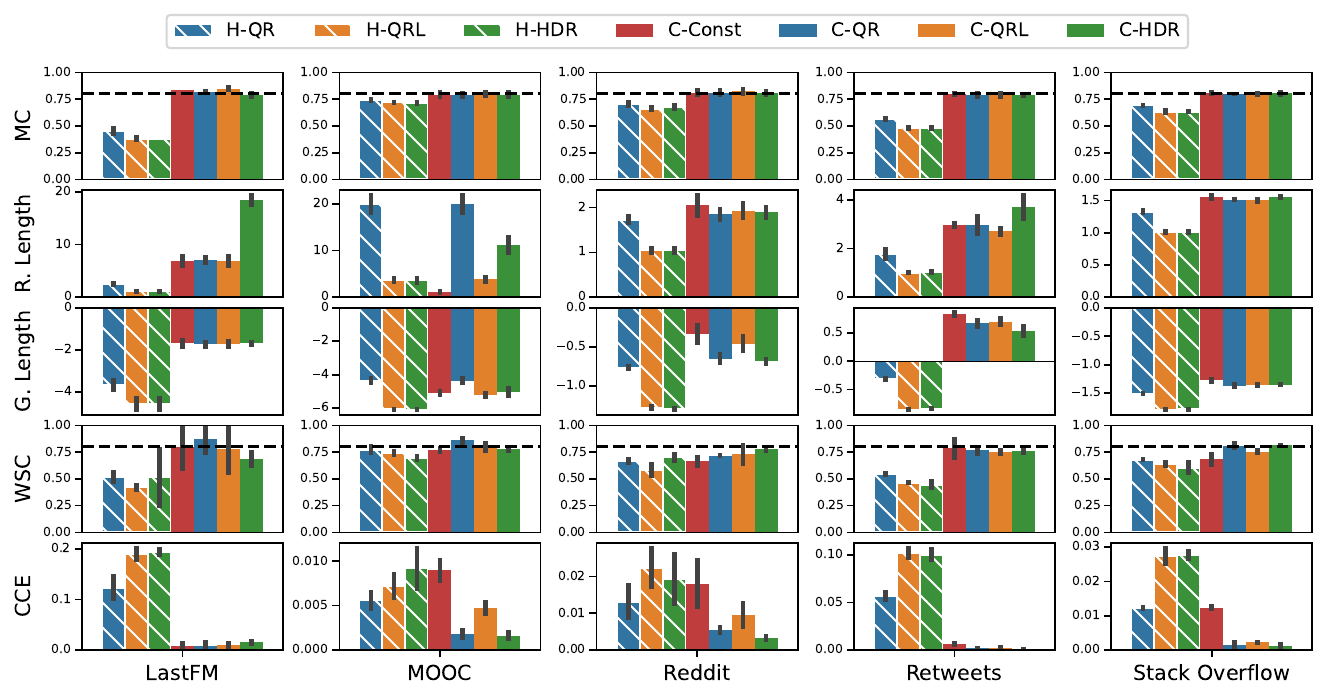}
    \caption{Performance of different methods producing a region for the time on real world datasets using the CLNM model. Heuristic methods are hatched.}
    \label{fig:real_world_metrics/per_model/CLNM_time}
\end{figure}

In Figure \ref{fig:real_world_metrics/per_model/CLNM_time}, the results are systematically organized in a table where each row represents a specific metric, as detailed in Section \ref{sec:metrics}, and each column corresponds to one of the datasets. This figure gives the results for various methods, described in Section \ref{sec:predmethods}, that are used to generate prediction regions solely for the arrival time. Each heuristic method and its corresponding conformal counterpart are represented in matching colors. To differentiate them, the heuristic methods are marked with hatching patterns.

The first row of the figure demonstrates that all CP methods attain the desired marginal coverage. In contrast, heuristic methods generally undercover, which aligns with expectations. The second row focuses on the average length of the prediction regions. Here, it is evident that heuristic methods generate smaller regions compared to their conformal counterparts. While this might seem beneficial, it is important to note that these smaller regions result from undercoverage, which diminishes their practical utility.

Among the heuristic methods, H-HDR consistently produces regions of smaller or equal lengths compared to H-QR and H-QRL for each prediction instance. Consequently, H-HDR emerges as the method with the smallest average region length.
H-QR, not adjusting adequately to the right-skewed nature of the distributions, tends to yield larger regions.

\begin{figure}[h]
    \centering
    \includegraphics[width=\linewidth]{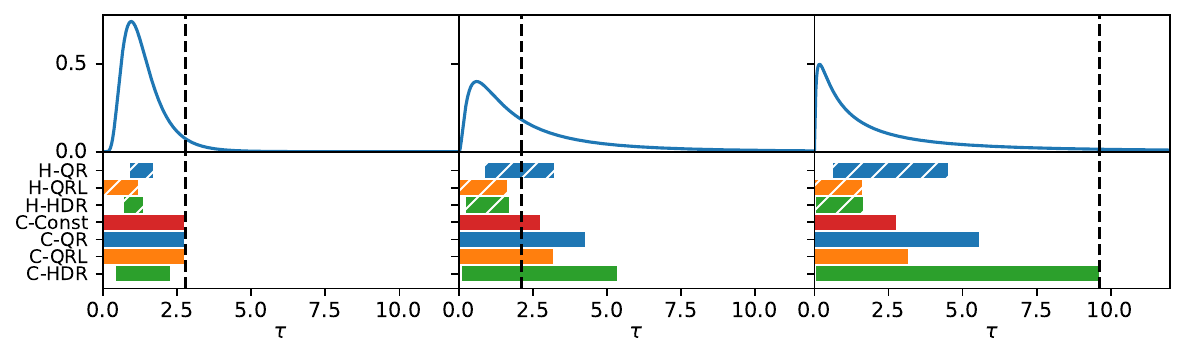}
    \caption{The figure showcases predictive distributions (blue) and realizations (dashed lines) in the first row, based on a calibration dataset. The second row illustrates prediction regions for various methods with $\alpha = 0.5$. It highlights the undercoverage of heuristic methods, the adaptive adjustments of conformal methods, and the notable differences between C-HDR and other methods in terms of region size. We provide an additional example with $\alpha = 0.2$ in \cref{sec:additional_example}.}
    \label{fig:conformal_intervals}
\end{figure}

Focusing now on the conformal methods, we exclude heuristic methods from this analysis due to their inability to achieve marginal coverage, which can lead to arbitrarily small regions. In the second row, the variations in average region length among CP methods differ across datasets. Notably, C-HDR, unlike its heuristic counterpart H-HDR, often yields larger average region lengths, especially in the LastFM, MOOC, and Retweets datasets. This difference arises because C-HDR adjusts the initial H-HDR prediction regions adaptively based on the individual predictive distributions. In contrast, C-QR and C-QRL modify their respective heuristic initial regions by a constant amount. While C-Const generates identical regions regardless of the history $\vh$, it occasionally has the smallest average region length while still maintaining marginal coverage. This occurs because C-Const does not tailor its regions to account for extreme right-skewed distributions, leading to regions that are either slightly larger or significantly smaller compared to other conformal methods. These two phenomena are exemplified in a toy example shown in \cref{fig:conformal_intervals}.

This figure demonstrates a scenario where the average region length of C-HDR is larger than that of other conformal methods in inter-arrival time prediction. The first row shows predictive distributions in blue and their corresponding realizations as dashed lines, based on three observations from a calibration dataset. In the second row, the prediction regions for seven methods are depicted with $\alpha = 0.5$. All heuristic methods underperform, achieving a maximum coverage of only $1/3$, which is less than the desired coverage of 0.5. Conformal prediction methods, in response, adjust their prediction regions to achieve coverage in at least two out of three cases. Despite H-HDR always producing shorter or equivalent lengths compared to H-QR and H-QRL, C-HDR   generates larger regions on average than other conformal methods. Again, C-Const, which does not adapt to individual predictive distributions, presents the smallest average regions among the conformal methods in this particular example. C-Const however does not achieve conditional coverage even asymptotically.

Returning to Figure \ref{fig:real_world_metrics/per_model/CLNM_time}, the third row introduces an alternative aggregation method for region lengths – the geometric mean. This method assigns less weight to larger regions and more to smaller ones. Here, C-HDR's performance is more in line with other conformal methods, indicating that average region length might not be a reliable metric, particularly in cases of high variability in conditional distributions.

The fourth and fifth rows of the figure assess conditional coverage.
WSC denotes coverage over the worst slab, with methods closer to $1 - \alpha$ being preferable, whereas CCE represents a conditional coverage error, which should be minimized.
Conformal methods, already proficient in achieving marginal coverage, exhibit a conditional coverage that is usually better than heuristic methods
based on the evaluated metrics.
Methods capable of tailoring prediction regions to specific instances are expected to exhibit enhanced conditional coverage. Although the WSC metric reveals no marked distinction among conformal methods, the CCE metric shows that C-HDR frequently attains one of the highest levels of conditional coverage. Moreover, C-QR often outperforms C-QRL in conditional coverage. As anticipated, the CCE metric reveals that C-Const generally exhibits the poorest conditional coverage, attributable to its lack of adaptability.


\subsubsection{Prediction regions for the mark}
\label{sec:results_mark}



\begin{figure}[h!]
    \centering
    \includegraphics[width=0.8\linewidth]{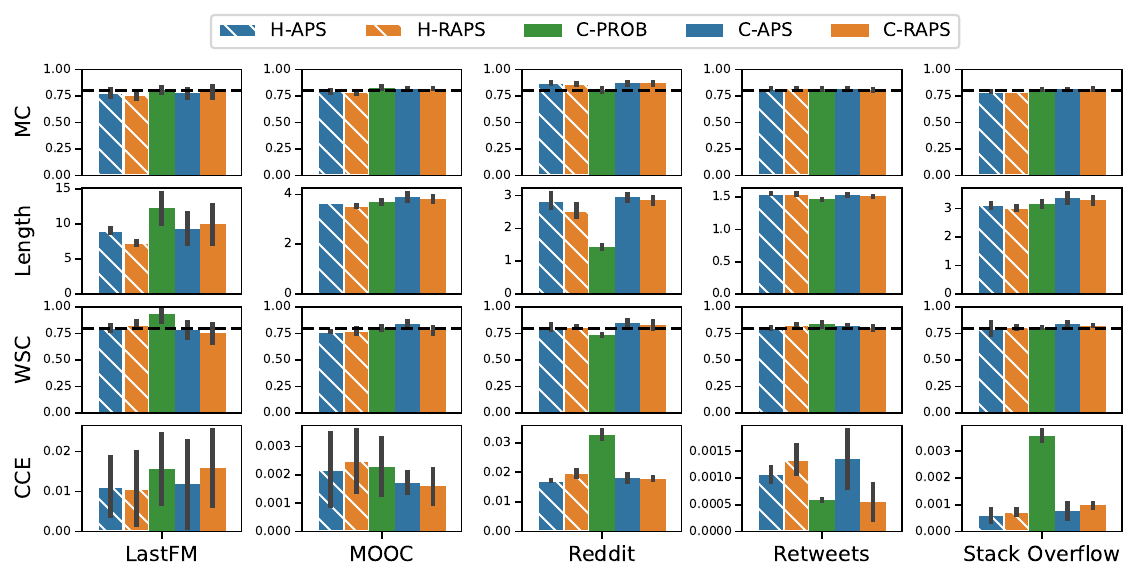}
    \caption{Performance of different methods producing a region for the mark on real world datasets using the CLNM model. Heuristic methods are hatched.}
    \label{fig:real_world_metrics/per_model/CLNM_mark}
\end{figure}

Figure \ref{fig:real_world_metrics/per_model/CLNM_mark} presents similar metrics than in Figure \ref{fig:real_world_metrics/per_model/CLNM_time}, but focuses on methods that generate prediction sets exclusively for the mark. Here, the heuristic methods H-APS and H-RAPS already meet the marginal coverage criteria, meaning that conformal prediction primarily offers theoretical backing rather than significant changes in predictions.

Turning our attention to the conformal methods, these methods show similar region lengths across all datasets, with the exception of Reddit, where C-PROB exhibits smaller region lengths.
However, on this same dataset and on Stack Overflow, C-PROB has a poor conditional coverage compared to both other conformal methods and heuristic methods.
This reflects similar findings discussed in Section \cref{sec:results_time}, where the method C-Const manages to attain short prediction regions, albeit with weak conditional coverage.
This is explained due to the fact that, in contrast to C-APS, C-PROB does not achieve conditional coverage asymptotically.


\subsubsection{Joint prediction regions for the arrival time and the mark}
\label{sec:results_joint}

\begin{figure}[h!]
    \centering
    \includegraphics[width=0.92\linewidth]{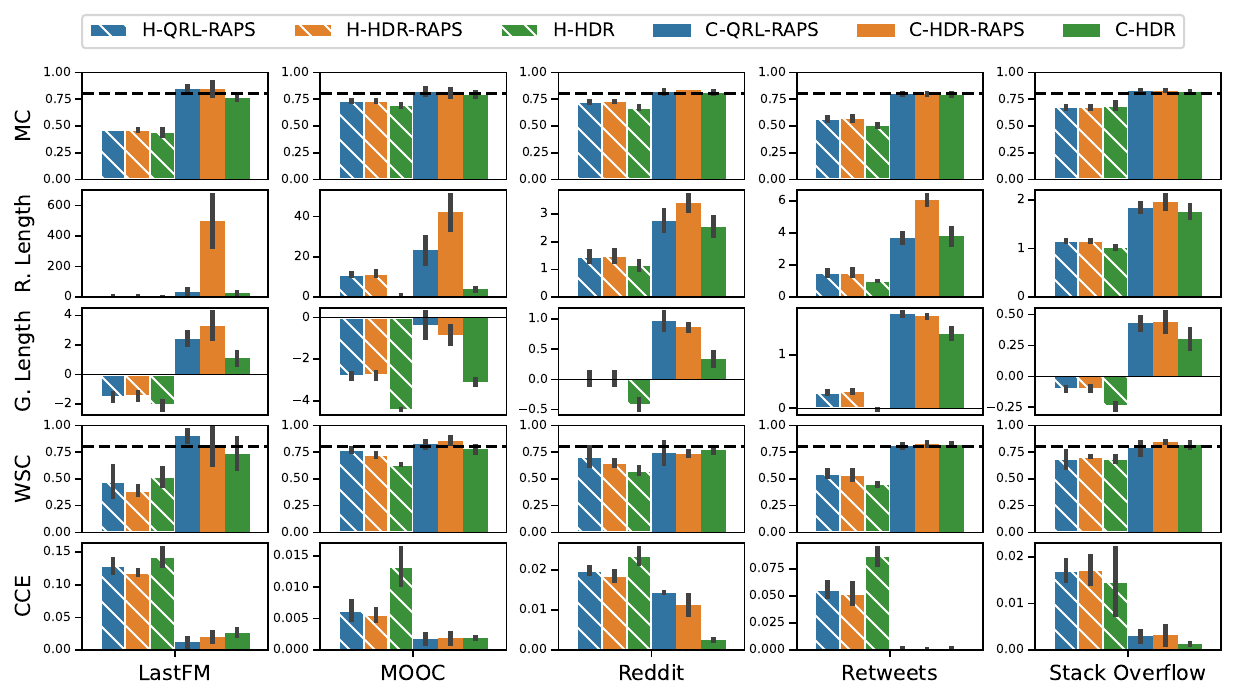}
    \caption{Performance of different methods producing a joint region for the time and mark on real world datasets using the CLNM model. Heuristic methods are hatched.}
    \label{fig:real_world_metrics/per_model/CLNM_joint}
\end{figure}

Figure \ref{fig:real_world_metrics/per_model/CLNM_joint} displays the same metrics as Figures \ref{fig:real_world_metrics/per_model/CLNM_time} and \ref{fig:real_world_metrics/per_model/CLNM_mark}, but it specifically focuses on methods that generate bivariate prediction sets for both the arrival time and the mark. Recall that we consider two main approaches. The first combines individual prediction regions, as detailed in \cref{sec:naive}. Under this approach, we examine two variants: QRL-RAPS, which merges QRL for time with RAPS for the mark, and HDR-RAPS, which combines HDR for time with RAPS for the mark. The second approach, outlined in \cref{sec:conformal_hdr}, directly creates a bivariate prediction region. Conformal versions of these methods are also considered in our analysis.

In the first row, we see that conformal methods successfully achieve marginal coverage, whereas heuristic methods tend to undercover. This observation mirrors the findings in the context of inter-arrival time prediction, as detailed in \cref{sec:results_time}. The second row focuses on the average region length. Here, C-HDR-RAPS tends to produce larger regions on average compared to C-QRL-RAPS, consistent with the explanations provided in \cref{sec:results_time} and \cref{fig:conformal_intervals}. Notably, C-HDR shows competitive average region lengths, outperforming C-HDR-RAPS. This highlights the benefit of creating joint regions that account for the interdependence between time and mark. In the third row, C-HDR stands out as the most effective on each dataset when using the geometric mean to average region lengths.

The last two rows illustrate that conformal methods attain better conditional coverage than their heuristic counterparts, echoing the results observed in \cref{sec:results_time}.
C-HDR obtains a competitive conditional coverage, especially on the dataset Reddit.

\begin{figure}
    \centering
    \includegraphics[width=\textwidth]{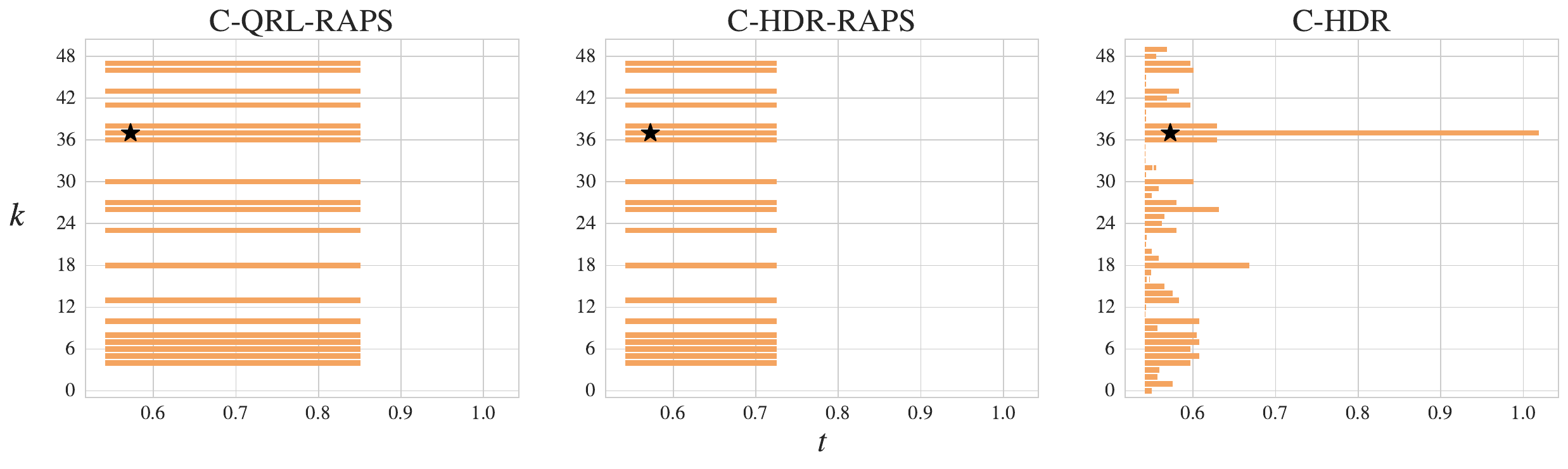}
    \caption{Examples of prediction regions generated by CLNM using the C-QRL-RAPS, C-HDR-RAPS, and C-HDR methods for the last event of a test sequence of the LastFM dataset. The black star corresponds to the actual event that materializes.}
    \label{fig:interval}
\end{figure}

For illustration, Figure \ref{fig:interval} gives examples of naive bivariate prediction regions generated by C-QRL-RAPS and C-HDR-RAPS, as well as a bivariate highest density region using C-HDR. We can see that the naive approaches produce constant size intervals for each of the marks selected by the C-RAPS approach. Conversely, C-HDR is able to generate variable-length prediction intervals for each mark by taking the inter-dependencies between the two variables into account.

\subsubsection{Empirical coverage for different coverage levels}
\label{sec:coverage_per_level}

\begin{figure}[h]
    \centering
    \includegraphics[width=\linewidth]{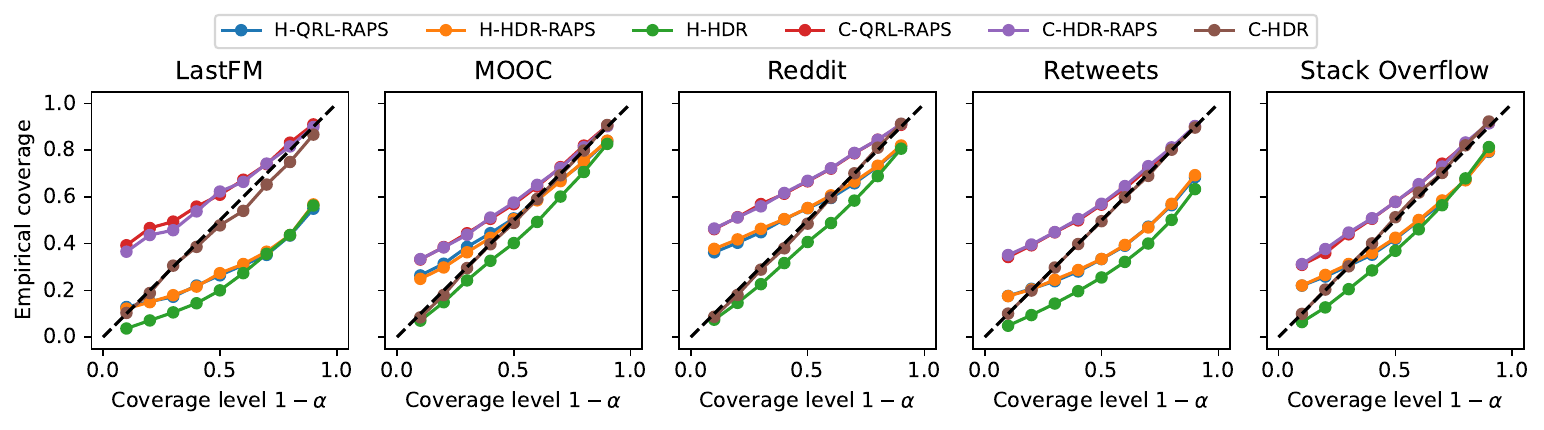}
    \caption{Empirical marginal coverage for different coverage levels with the CLNM model. All conformal methods achieve marginal coverage but the naive method tends to overcover, especially for small coverage levels. The heuristic methods do not achieve coverage in most cases.}
    \label{fig:coverage_per_level/CLNM/joint}
\end{figure}

\cref{fig:coverage_per_level/CLNM/joint} shows the marginal coverage achieved by various methods generating joint prediction regions for both the arrival time and mark. This figure extends the analysis beyond the specific miscoverage level of $\alpha = 0.2$, as shown in the first row of  \cref{fig:real_world_metrics/per_model/CLNM_joint}, by including a range of coverage levels.

It is evident that heuristic methods generally underperform at all coverage levels, with this tendency becoming more pronounced at higher coverage levels. Conversely, conformal methods that construct individual predictions (outlined in Section \cref{sec:naive}) often overcover, particularly at lower coverage levels, due to their inherent conservativeness. Notably, C-HDR strikes an appropriate balance, maintaining the correct level of conservativeness across the various coverage levels.

In \cref{sec:additional_coverage_per_level}, we provide additional results for the coverage per level in the context of prediction regions for the time or for the mark.

\section{Conclusion and future work}

By integrating the methodologies of conformal prediction and neural TPPs, we have established a more robust approach to uncertainty quantification in TPPs. This is achieved by creating distribution-free joint prediction regions for the arrival time and its associated mark. The main challenge is to handle both a strictly positive, continuous response and a categorical response without distributional assumptions. We have also explored independently generating univariate prediction regions for the arrival time and the mark.

Our experiments highlight the importance of using conformal inference to ensure finite-sample marginal coverage. Indeed, heuristic methods tend to undercover in cases involving the prediction of arrival time or the simultaneous prediction of both arrival time and marks, with occasional success in predicting marks alone.

We also emphasize the significance of choosing appropriate conformal scores. C-HDR and C-QR show good conditional coverage, unlike C-Const, which lacks adaptability. The non-adaptive nature of C-Const leads to shorter average region lengths, which may appear advantageous at first glance. The same holds for C-PROB.

Our analysis underscores the importance of considering interdependence. Indeed, C-HDR, our extension of HPD-split \citep{Izbicki2022-ru} to bivariate outputs, outperforms C-HDR-RAPS. The superiority of C-HDR stems from its incorporation of joint regions that effectively consider and account for interdependence, whereas C-HDR-RAPS, in contrast, simplistically combines individual prediction regions through Bonferroni adjustments.

While this paper focuses on marked TPPs, the techniques presented here, especially those involving C-HDR, have the potential to extend to other prediction problems where the target variable is a vector comprising a combination of continuous and categorical variables. To the best of our knowledge, this is the first time such prediction regions are explored in the context of conformal prediction.

There are multiple possible directions for future work. Firstly, we plan to explore conformal methods that can adapt to temporal dependencies by either iteratively modifying the targeted coverage level \citep{Gibbs2021-pj} or the quantile of the scores \citep{Angelopoulos2023-bz}. Secondly, we intend to examine sequence splitting techniques that leverage the exchangeability of non-conformity scores within a sequence when splits progressively lose temporal dependence. A potential approach would be to detect significant deviations from exchangeability using statistical tests such as \citep{Saha2023-wc}. Thirdly, we aim to address the challenge of managing non-exchangeable groups of exchangeable sequences based on ideas in \cite{Dunn2023-ba}. For instance, given groups of sequences, each associated to a specific subject, the task could be to generate predictions for a new subject. Extensions to conformal prediction are needed because sequences associated to different subjects are not exchangeable. Lastly, the framework of conformal risk control offers an extension to conformal prediction by not only controlling marginal coverage but also others risks such as the conditional coverage error \citep{Angelopoulos2021-ud,Angelopoulos2024-ld}.


\section*{Declarations}

\textbf{Funding} The research leading to these results received funding from Fonds De La Recherche Scientifique under Grant Agreement No 40020829 and 40003156 for authors Victor Dheur, Tanguy Bosser and Souhaib Ben Taieb. It additionally received funding from Fundação de Amparo à Pesquisa do Estado de São Paulo under Grant Agreement No 2023/07068-1 and from Conselho Nacional de Desenvolvimento Cientifico e Tecnológico under Grant Agreement No 306943/2017-4 for the author Rafael Izbicki.

\textbf{Conflicts of interest/Competing interests} I declare that the authors have no competing interests as defined by Springer, or other interests that might be perceived to influence the results and/or discussion reported in this paper.

\textbf{Ethics approval} Not Applicable.

\textbf{Consent to participate} Not applicable.

\textbf{Consent for publication} Not applicable.

\textbf{Availability of data and material} Data is provided in the associated repository at \url{https://github.com/tanguybosser/conf_tpp}.

\textbf{Code availability} Code to reproduce the experiments is provided in the associated repository at \url{https://github.com/tanguybosser/conf_tpp}.

\textbf{Authors' contributions} Tanguy Bosser, Victor Dheur, and Souhaib Ben Taieb developed the main ideas for this paper. 
Bosser and Dheur took the lead in implementing experiments and creating figures. 
Souhaib Ben Taieb managed the paper, offering key directions, feedback, and organizational oversight.
Bosser, Dheur, and Ben Taieb, collaborated on writing the initial draft. Rafael Izbicki provided valuable feedback and made significant contributions to critical revisions of the manuscript. All authors thoroughly reviewed and approved the final version of the manuscript.

\clearpage

\bibliography{sn-bibliography,paperpile}

\clearpage
\appendix

\section{Results on other models}
\label{sec:results_other_models}

In \cref{sec:results}, we provided results for the CLNM neural TPP model presented in \cref{sec:baselines}.
In \cref{sec:results_FNN,sec:results_RMTPP,sec:results_sahp}, we present additional findings for the FNN, RMTPP, and SAHP models, respectively, on the datasets discussed in the main text. Across all these models, our conclusions align with those outlined in Section \ref{sec:results}, applicable to all scenarios considered, namely, predictions for the time, the mark, or joint predictions on the time and mark.

\subsection{FNN}
\label{sec:results_FNN}

\begin{figure}[H]
    \centering
    \includegraphics[width=0.92\linewidth]{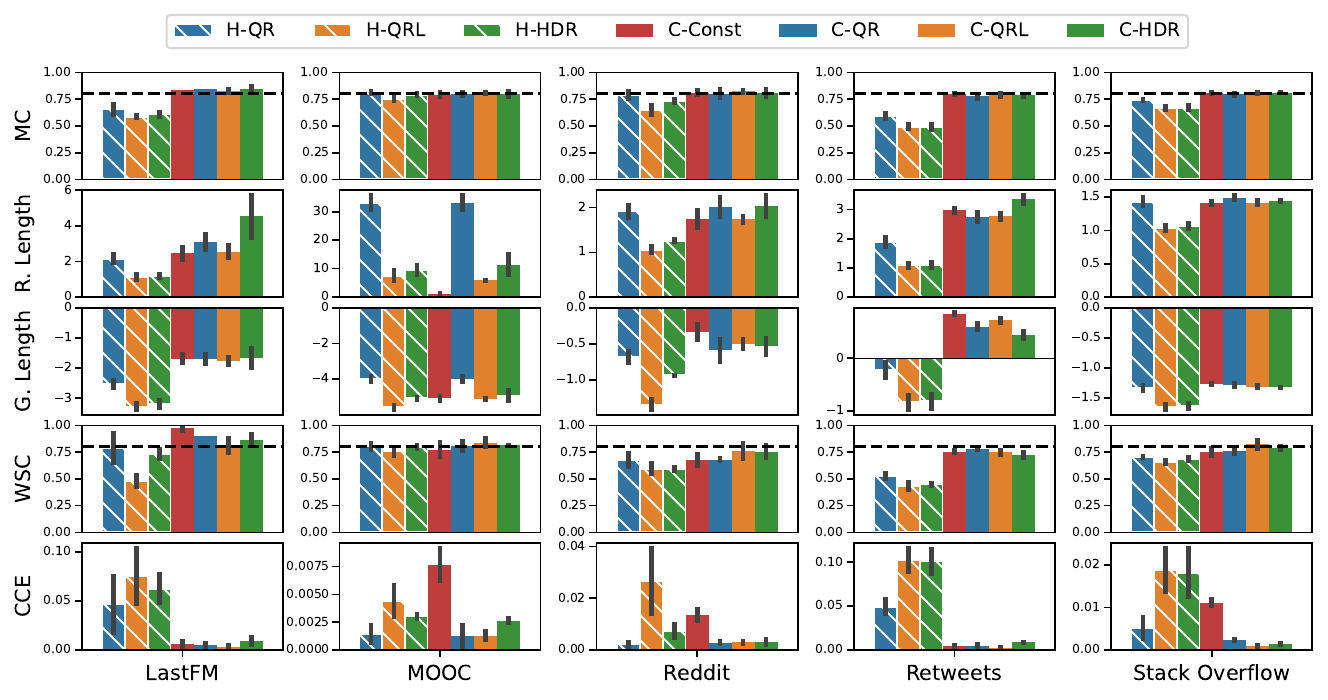}
    \caption{Performance of different methods producing a region for the time on real world datasets using the FNN model.}
\end{figure}

\begin{figure}[H]
    \centering
    \includegraphics[width=0.8\linewidth]{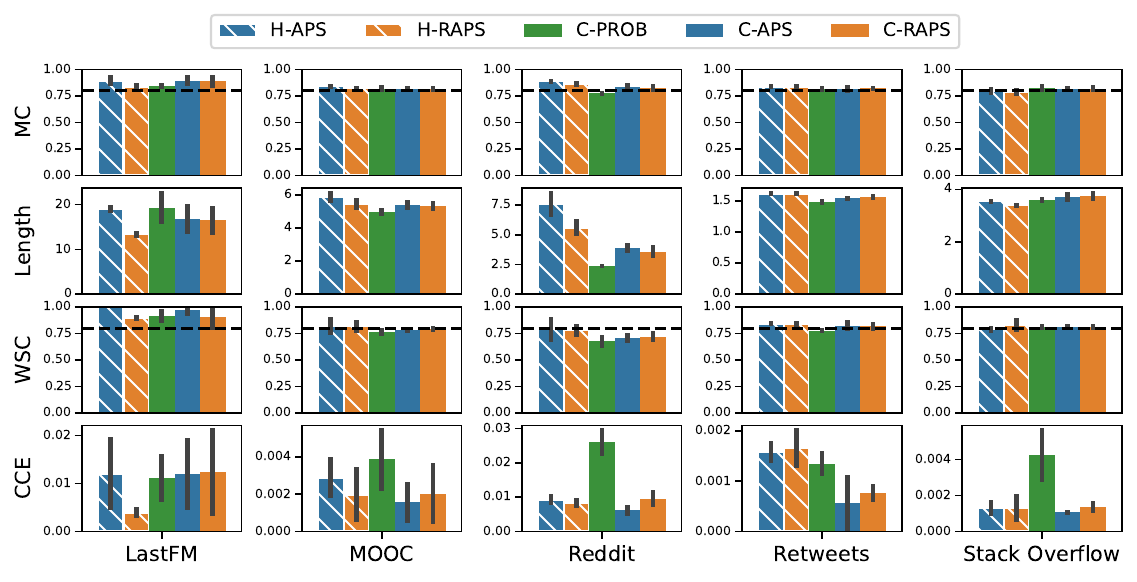}
    \caption{Performance of different methods producing a region for the mark on real world datasets using the FNN model.}
\end{figure}

\begin{figure}[H]
    \centering
    \includegraphics[width=0.92\linewidth]{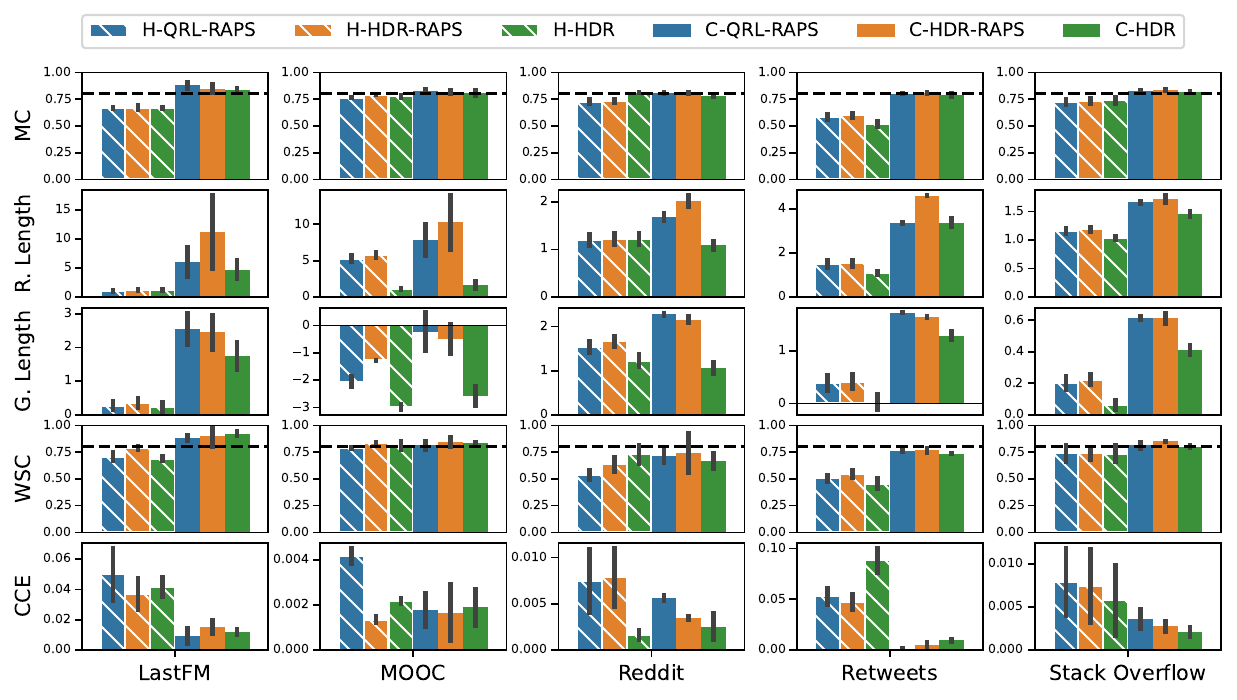}
    \caption{Performance of different methods producing a joint region for the time and mark on real world datasets using the FNN model.}
\end{figure}

\subsection{RMTPP}
\label{sec:results_RMTPP}

\begin{figure}[H]
    \centering
    \includegraphics[width=0.92\linewidth]{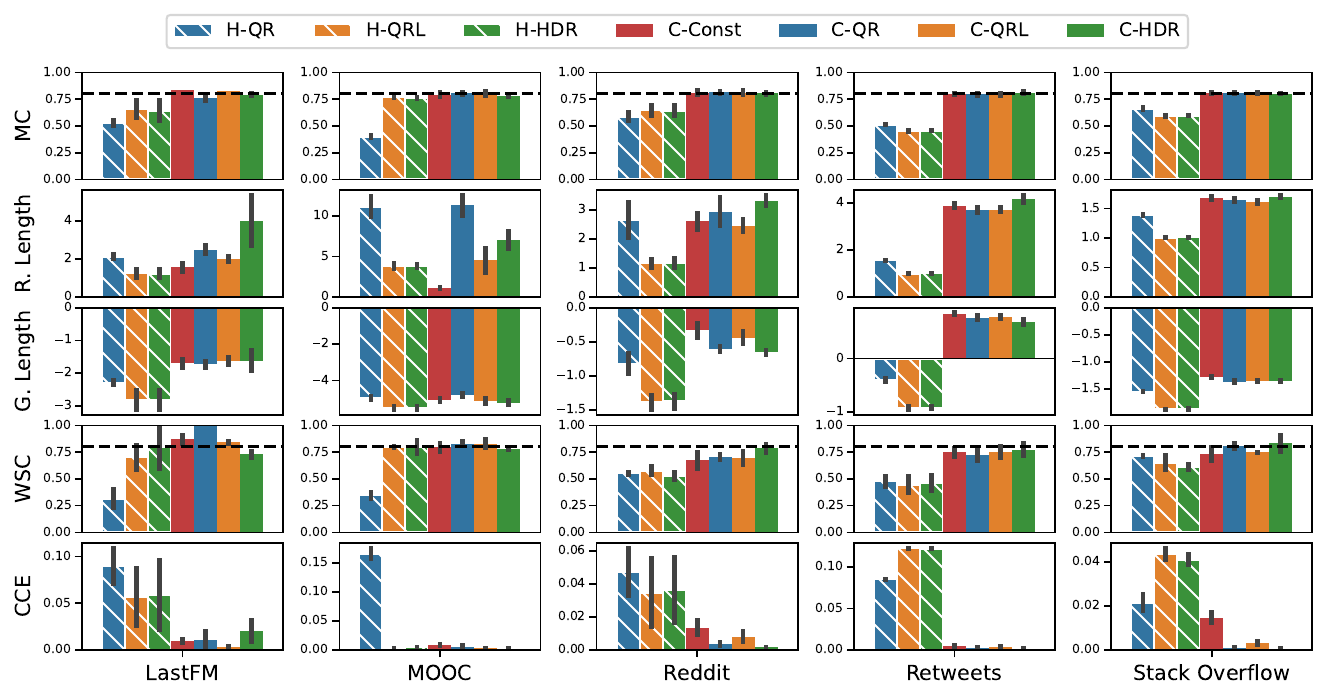}
    \caption{Performance of different methods producing a region for the time on real world datasets using the RMTPP model.}
\end{figure}

\begin{figure}[H]
    \centering
    \includegraphics[width=0.8\linewidth]{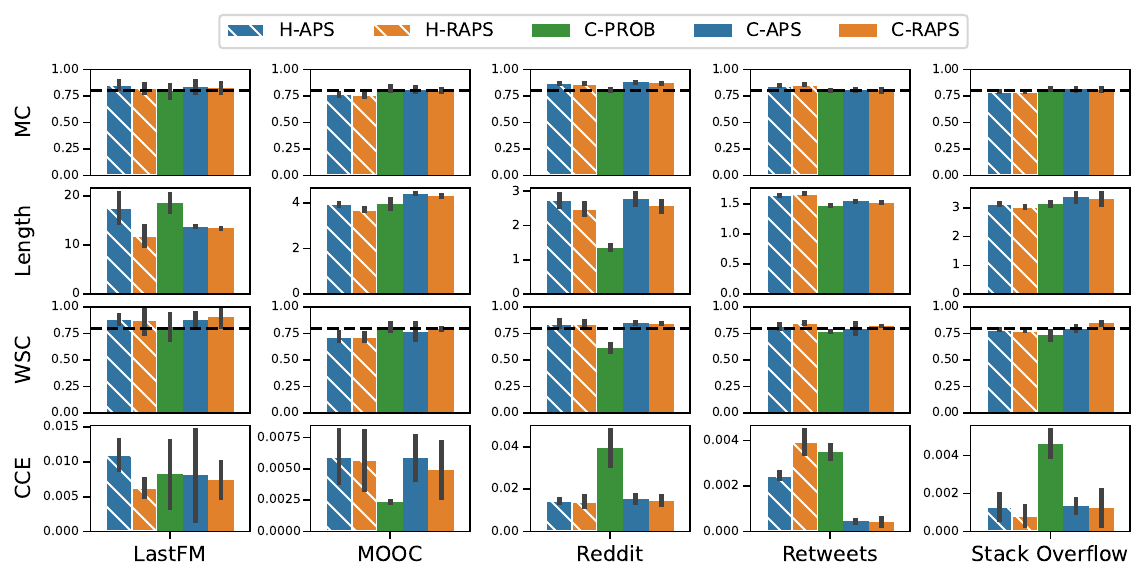}
    \caption{Performance of different methods producing a region for the mark on real world datasets using the RMTPP model.}
\end{figure}

\begin{figure}[H]
    \centering
    \includegraphics[width=0.92\linewidth]{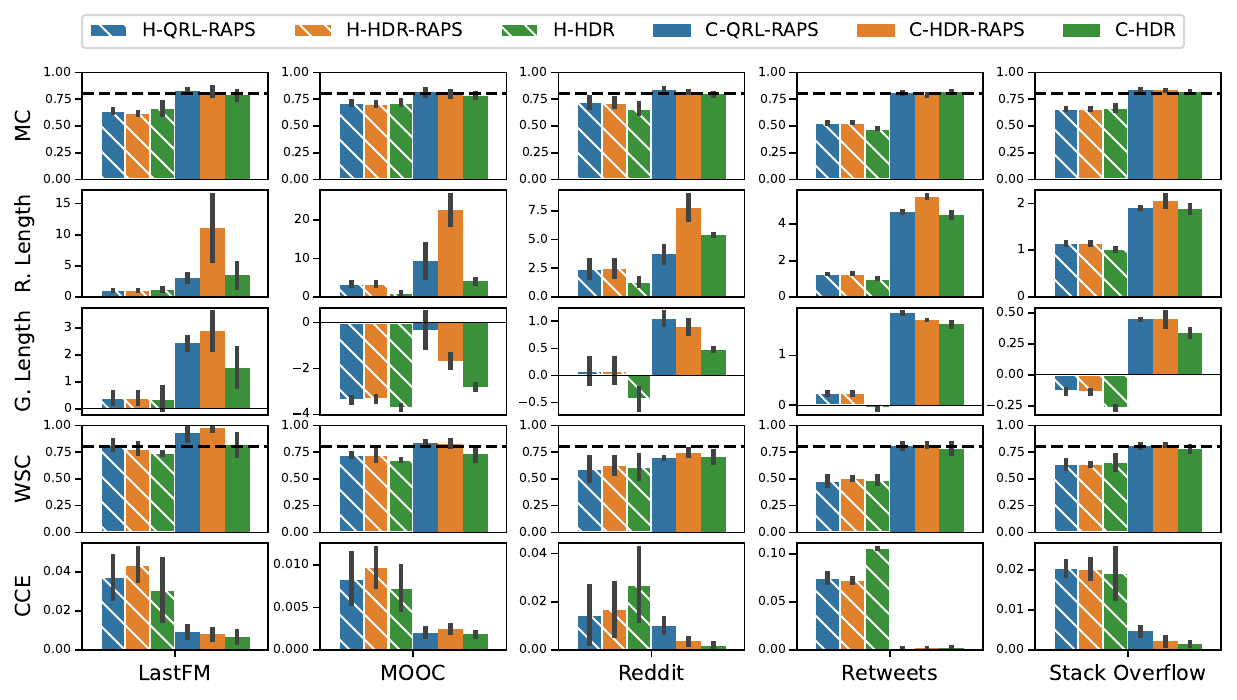}
    \caption{Performance of different methods producing a joint region for the time and mark on real world datasets using the RMTPP model.}
    \caption{}
\end{figure}

\subsection{SAHP}
\label{sec:results_sahp}

\begin{figure}[H]
    \centering
    \includegraphics[width=0.92\linewidth]{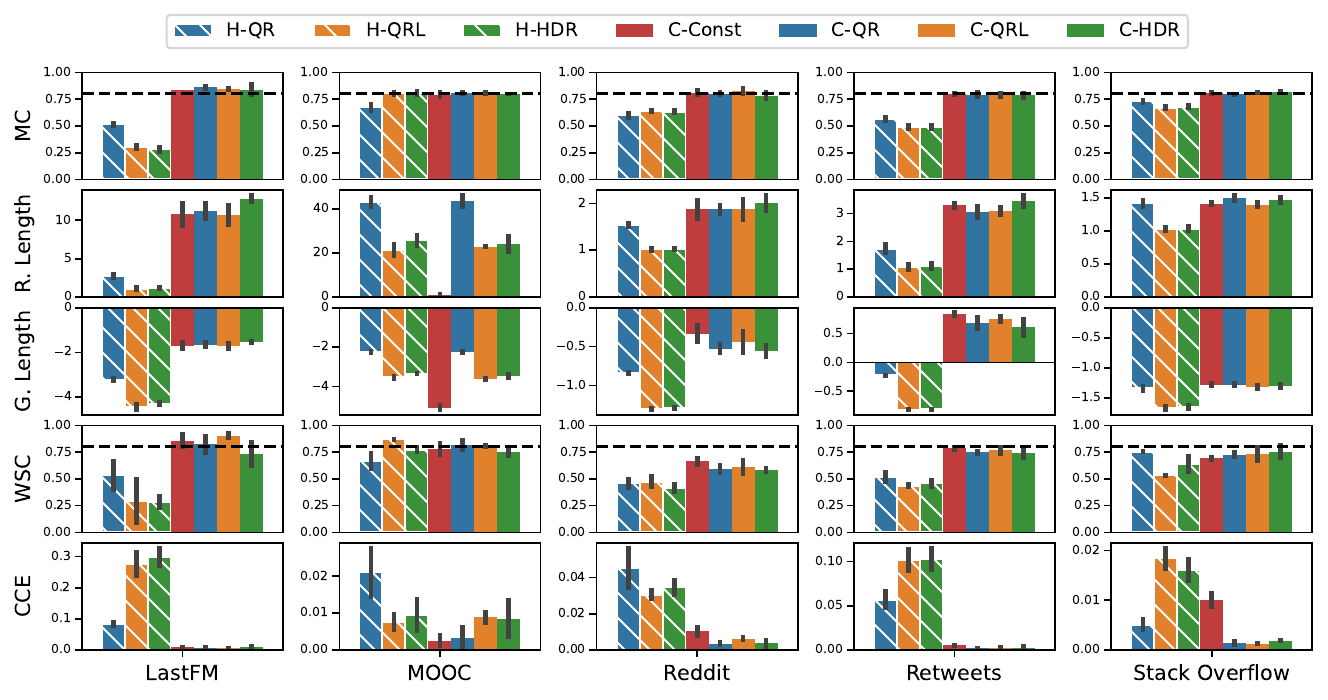}
    \caption{Performance of different methods producing a region for the time on real world datasets using the SAHP model.}
\end{figure}

\begin{figure}[H]
    \centering
    \includegraphics[width=0.8\linewidth]{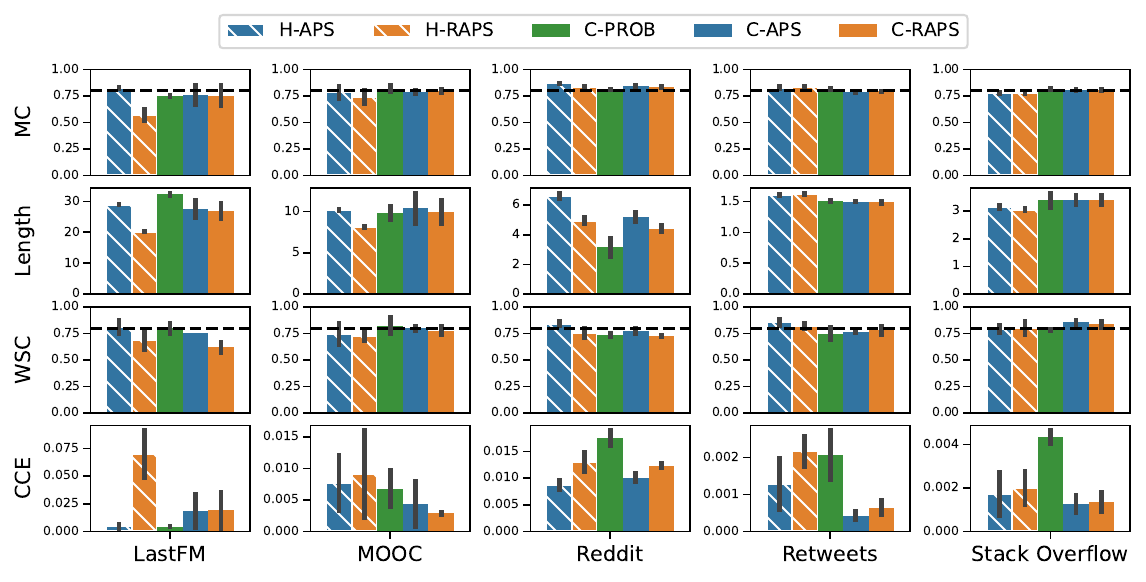}
    \caption{Performance of different methods producing a region for the mark on real world datasets using the SAHP model.}
\end{figure}

\begin{figure}[H]
    \centering
    \includegraphics[width=0.92\linewidth]{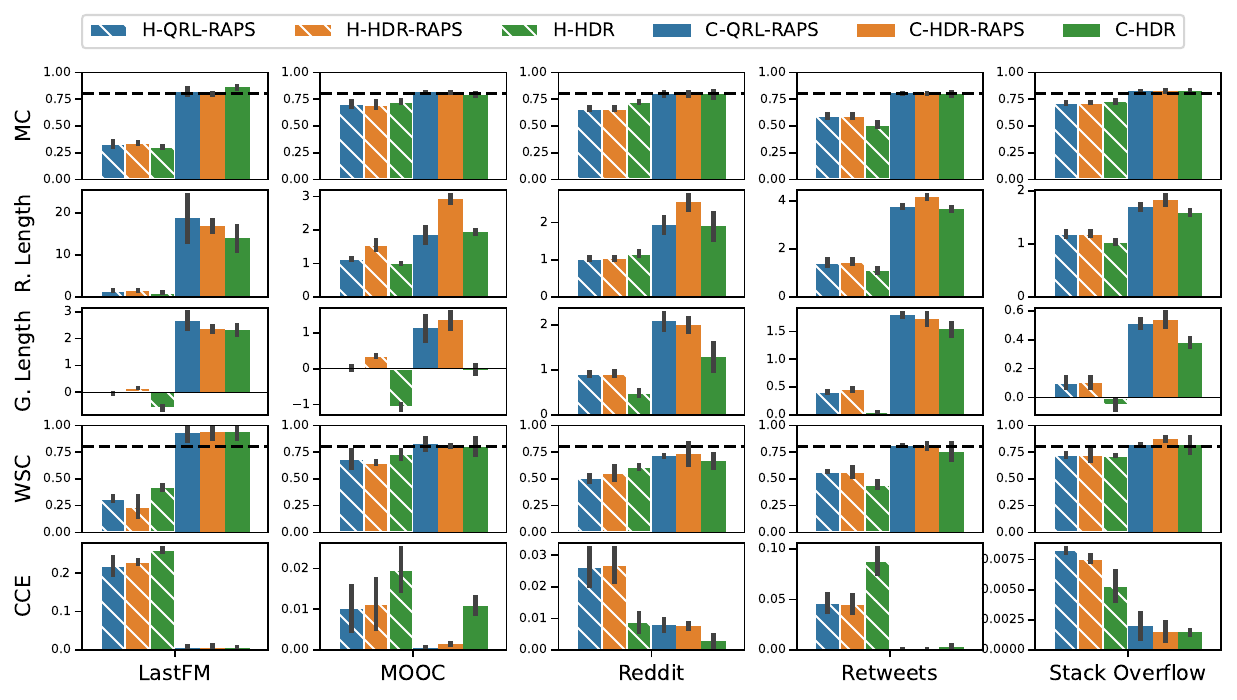}
    \caption{Performance of different methods producing a joint region for the time and mark on real world datasets using the SAHP model.}
\end{figure}

\section{Results on other real-world and synthetic datasets}

In this section, we report the results on the Github, MIMIC2, Wikipedia, and Hawkes datasets for all models and all scenarios. We note that the findings on these datasets are also generally consistent with our conclusions from \cref{sec:results}. Nonetheless, we usually observe a large variability in the results for Github, MIMIC2, and Wikipedia, explained by the few number of observations in the calibration and test sequences. We therefore invite the reader to exercise caution when interpreting the findings on these real-world datasets.

Finally, for the Hawkes dataset, we observe that heuristic methods tend to already attain the desired coverage level. This finding may be explained by a too simplistic underlying generative Hawkes process, which is already well fitted by the MTPP models. We plan to investigate more complex simulated point processes as part of our future work.

\subsection{CLNM}

\begin{figure}[H]
    \centering
    \includegraphics[width=0.7\linewidth]{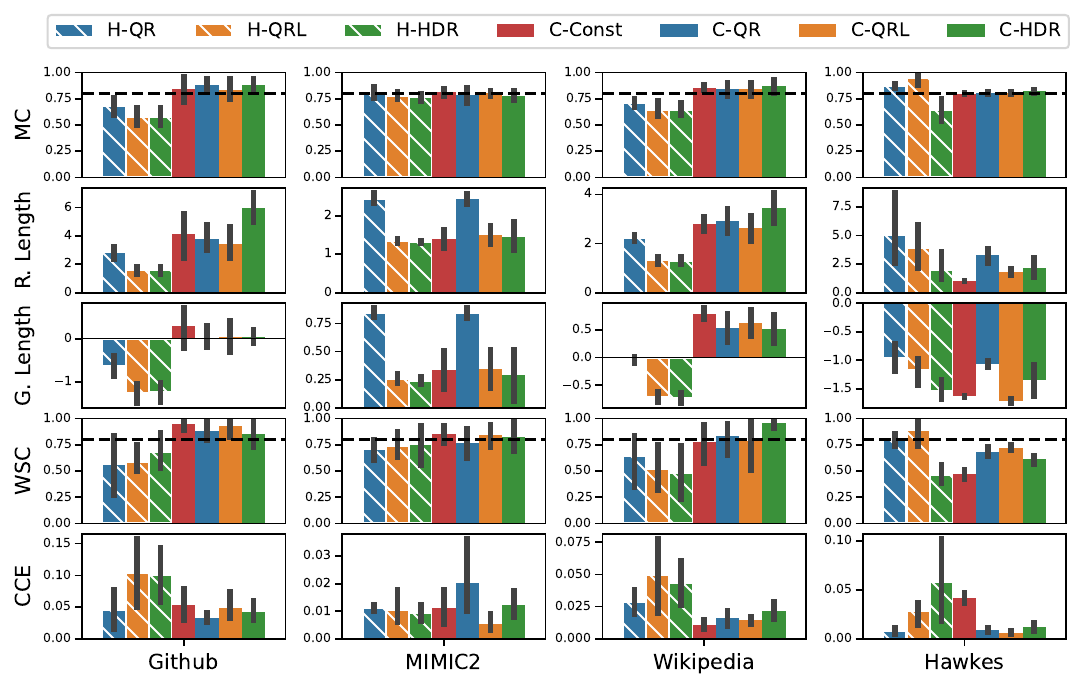}
    \caption{Performance of different methods producing a region for the time on the datasets not discussed in the main text using the CLNM model.}
\end{figure}

\begin{figure}[H]
    \centering
    \includegraphics[width=0.63\linewidth]{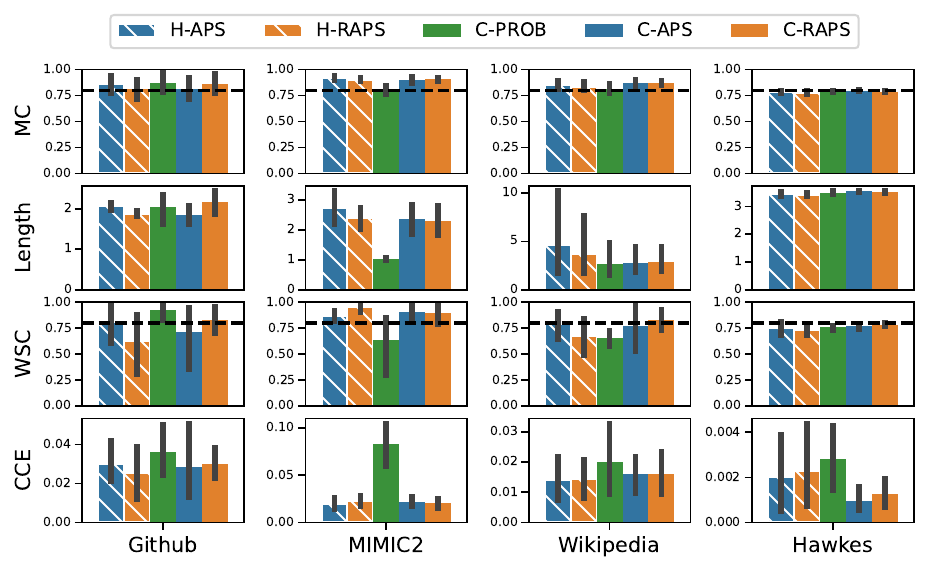}
    \caption{Performance of different methods producing a region for the mark on the datasets not discussed in the main text using the CLNM model.}
\end{figure}

\begin{figure}[H]
    \centering
    \includegraphics[width=0.84\linewidth]{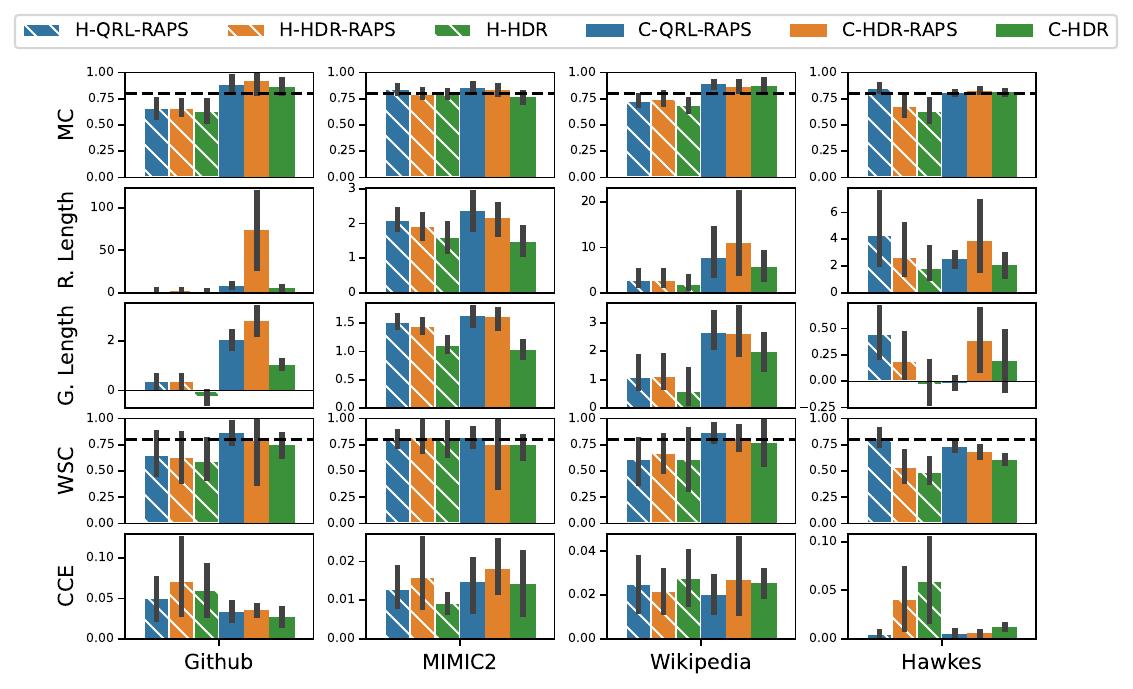}
    \caption{Performance of different methods producing a joint region for the time and mark on the datasets not discussed in the main text using the CLNM model.}
\end{figure}

\subsection{FNN}

\begin{figure}[H]
    \centering
    \includegraphics[width=0.7\linewidth]{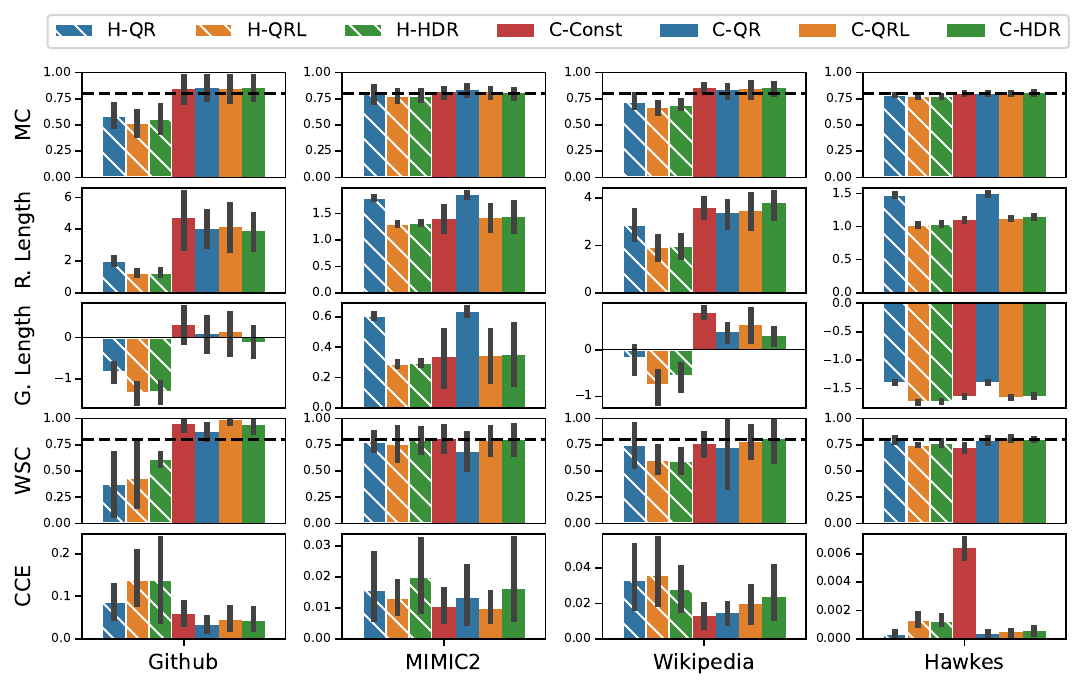}
    \caption{Performance of different methods producing a region for the time on the datasets not discussed in the main text using the FNN model.}
\end{figure}

\begin{figure}[H]
    \centering
    \includegraphics[width=0.63\linewidth]{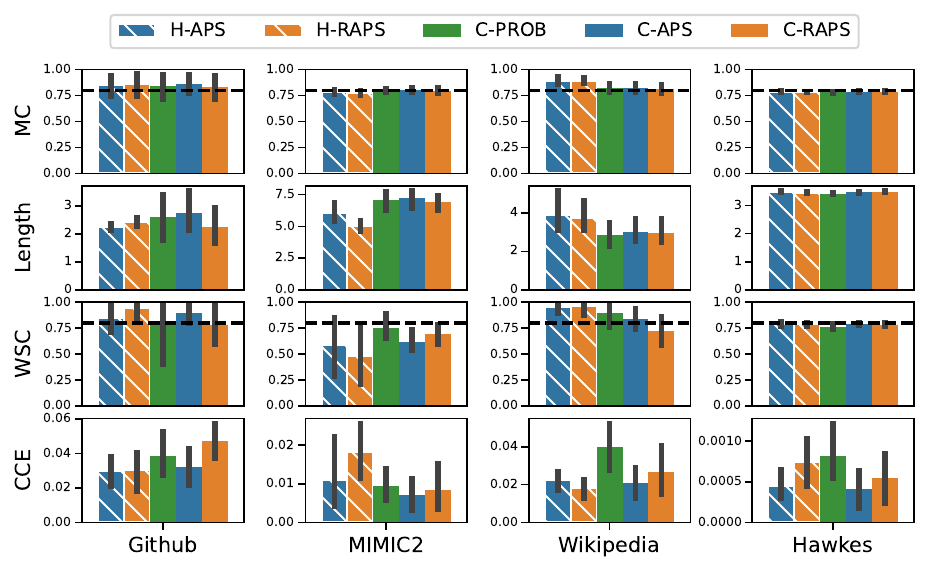}
    \caption{Performance of different methods producing a region for the mark on the datasets not discussed in the main text using the FNN model.}
\end{figure}

\begin{figure}[H]
    \centering
    \includegraphics[width=0.84\linewidth]{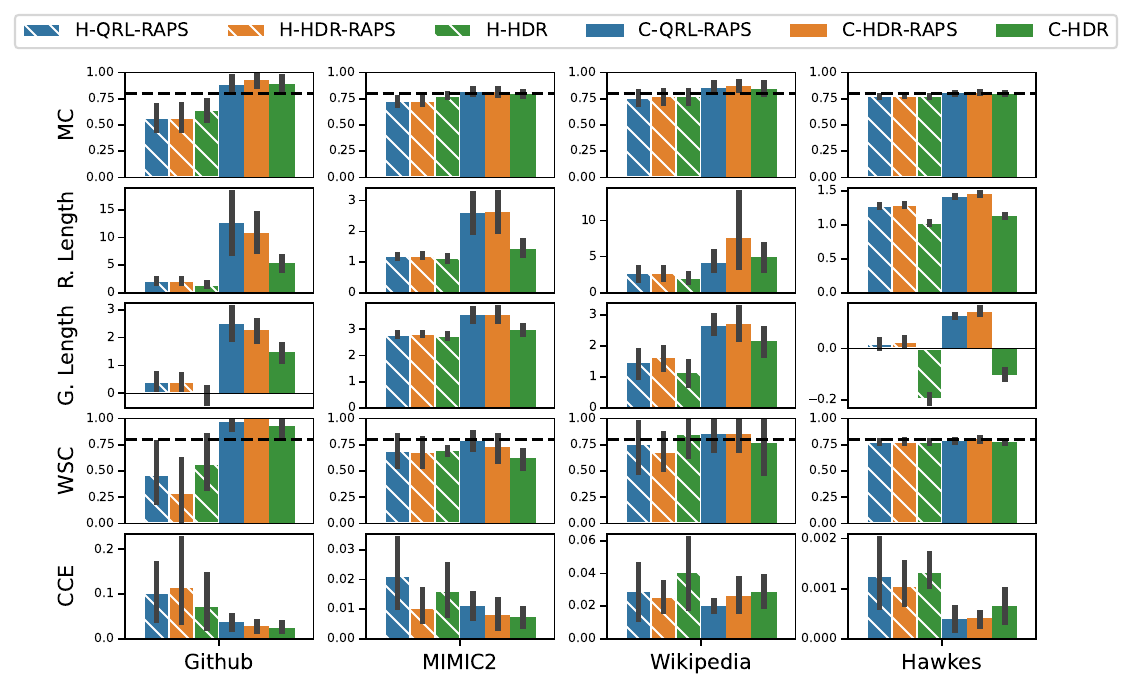}
    \caption{Performance of different methods producing a joint region for the time and mark on the datasets not discussed in the main text using the FNN model.}
\end{figure}

\subsection{RMTPP}

\begin{figure}[H]
    \centering
    \includegraphics[width=0.7\linewidth]{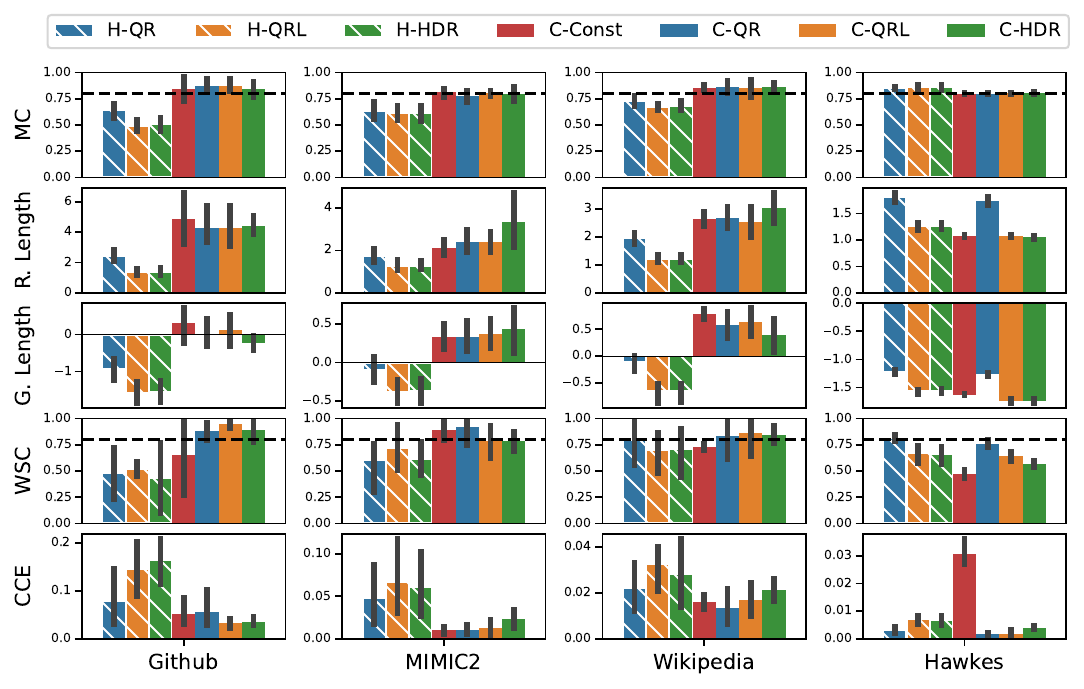}
    \caption{Performance of different methods producing a region for the time on the datasets not discussed in the main text using the RMTPP model.}
\end{figure}

\begin{figure}[H]
    \centering
    \includegraphics[width=0.63\linewidth]{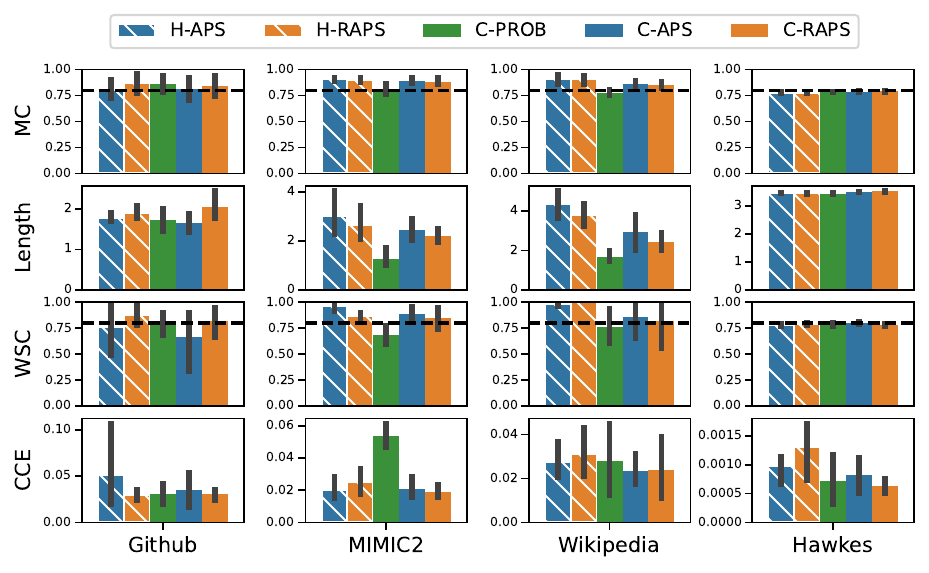}
    \caption{Performance of different methods producing a region for the mark on the datasets not discussed in the main text using the RMTPP model.}
\end{figure}

\begin{figure}[H]
    \centering
    \includegraphics[width=0.84\linewidth]{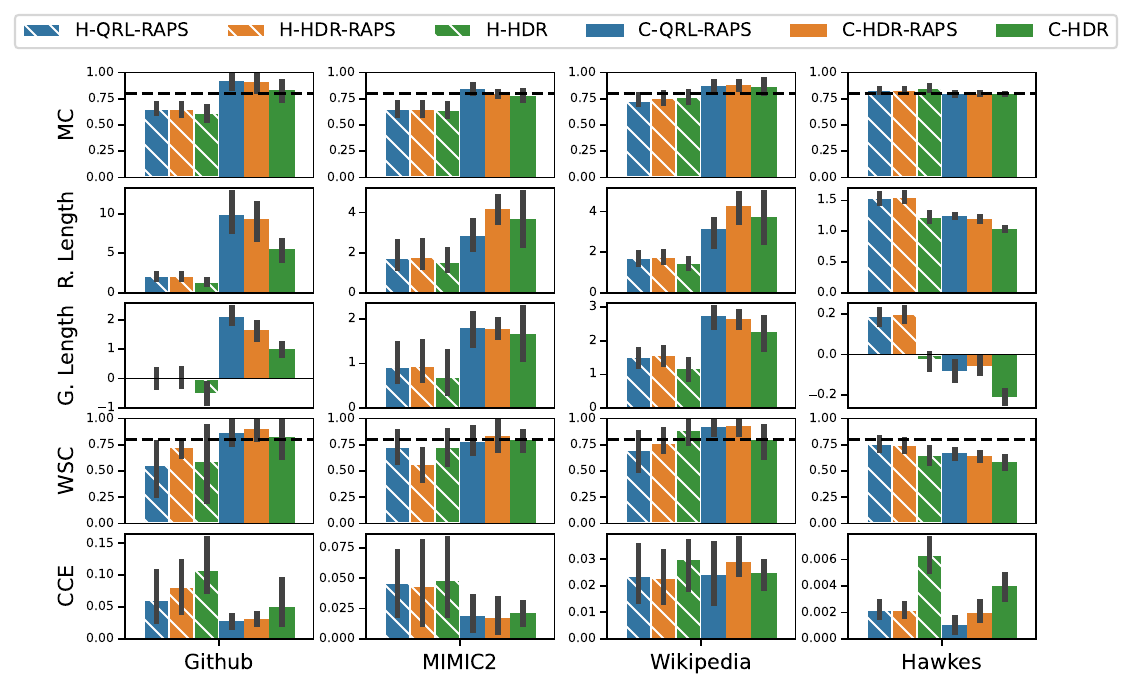}
    \caption{Performance of different methods producing a joint region for the time and mark on the datasets not discussed in the main text using the RMTPP model.}
\end{figure}

\subsection{SAHP}

\begin{figure}[H]
    \centering
    \includegraphics[width=0.7\linewidth]{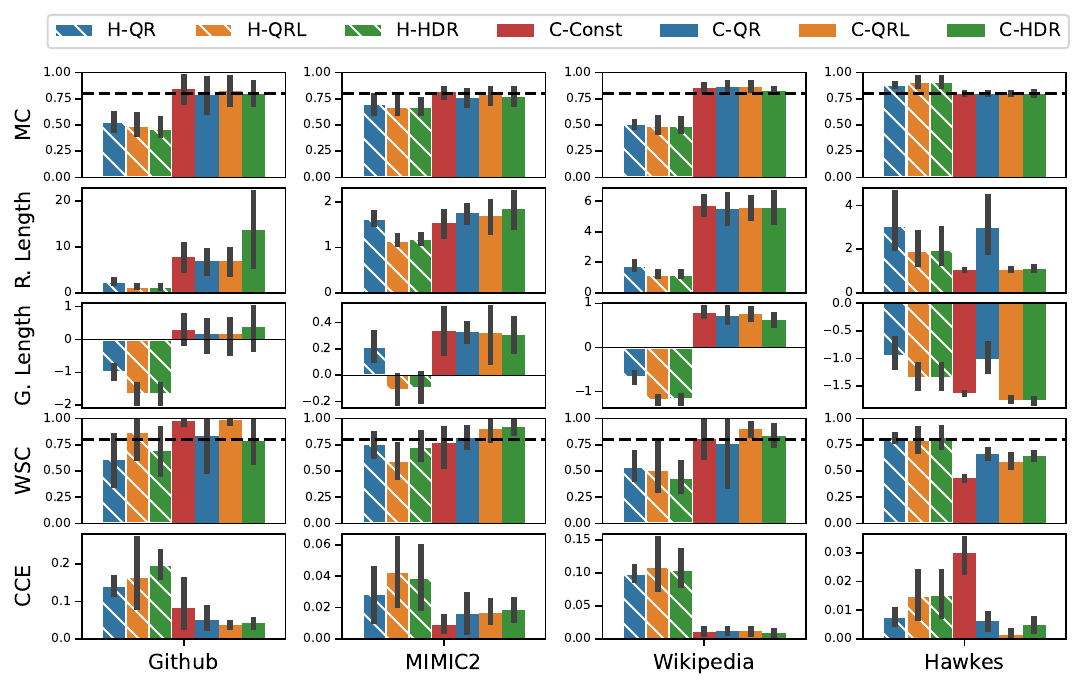}
    \caption{Performance of different methods producing a region for the time on the datasets not discussed in the main text using the SAHP model.}
\end{figure}

\begin{figure}[H]
    \centering
    \includegraphics[width=0.63\linewidth]{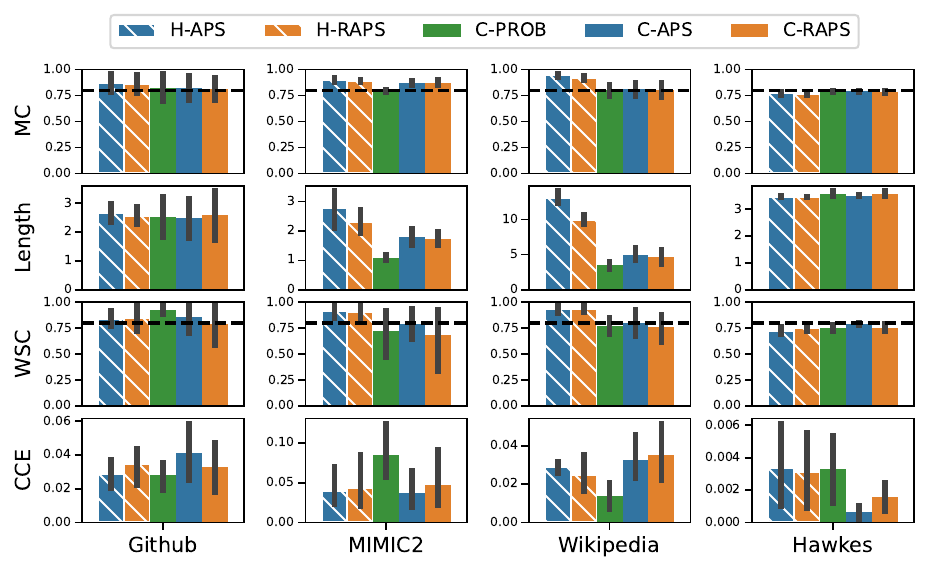}
    \caption{Performance of different methods producing a region for the mark on the datasets not discussed in the main text using the SAHP model.}
\end{figure}

\begin{figure}[H]
    \centering
    \includegraphics[width=0.84\linewidth]{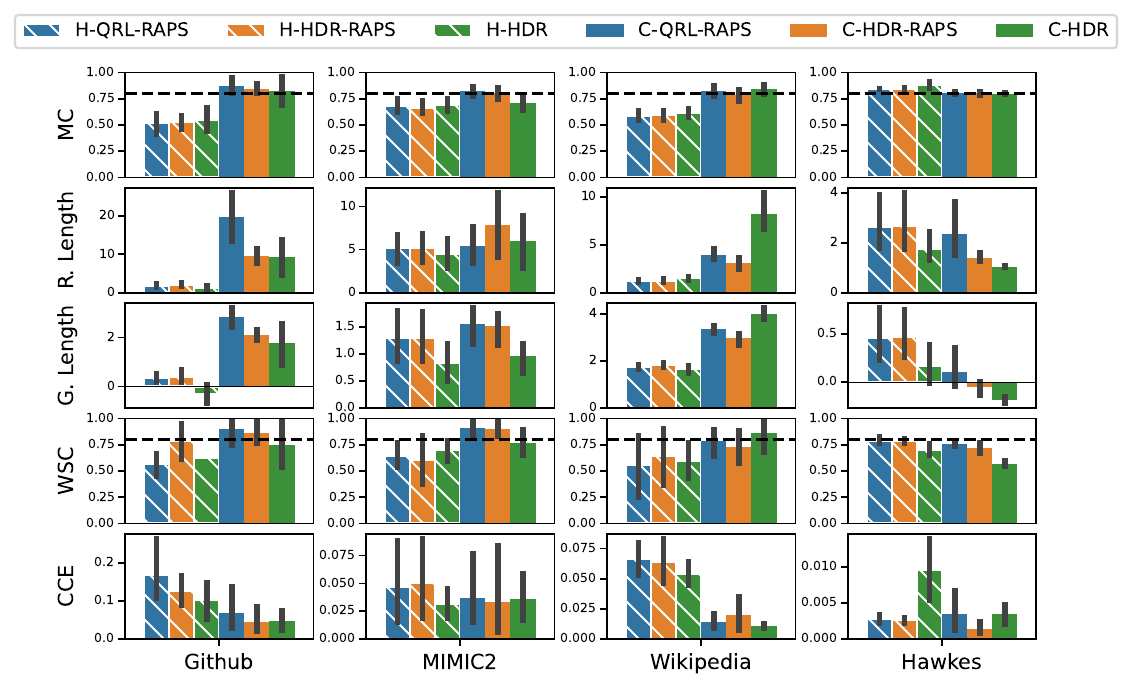}
    \caption{Performance of different methods producing a joint region for the time and mark on the datasets not discussed in the main text using the SAHP model.}
\end{figure}

\section{Coverage per level}
\label{sec:additional_coverage_per_level}

\cref{sec:coverage_per_level} discussed the empirical marginal coverage obtained at different coverage levels for methods that generate a joint prediction region on the arrival time and mark.
In this section, we present additional results for methods that generate a prediction region individually for either the time or mark.

\begin{figure}[H]
    \centering
    \includegraphics[width=\linewidth]{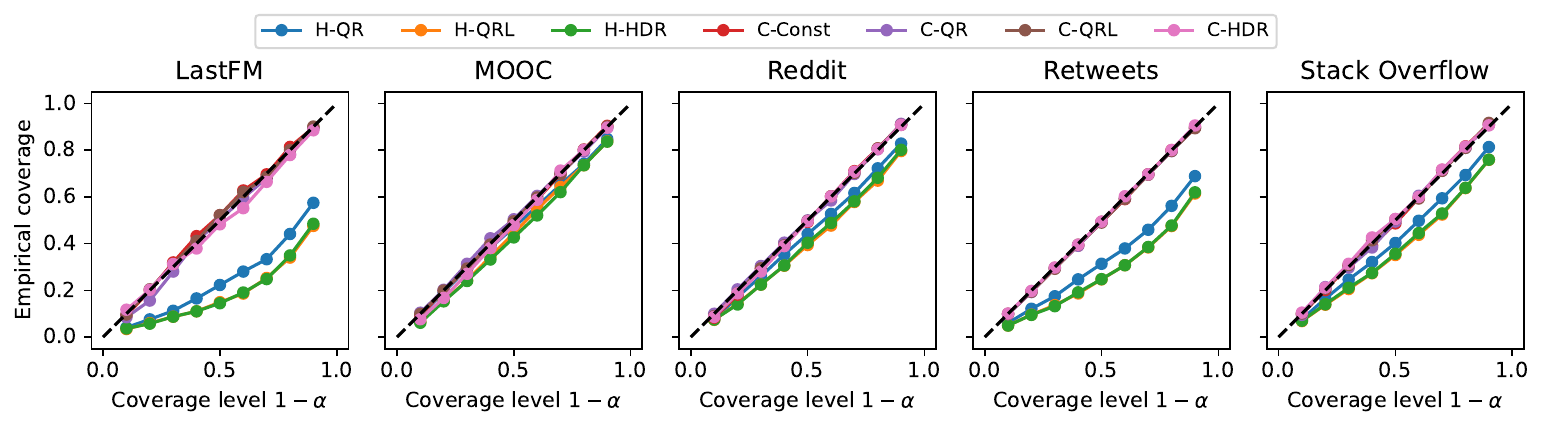}
    \caption{Empirical marginal coverage for different coverage levels for methods that produce a prediction region for the time with the CLNM model.}
    \label{fig:coverage_per_level/CLNM/time}
\end{figure}

\cref{fig:coverage_per_level/CLNM/time} shows that conformal methods for the time attain the desired coverage at all levels while heuristic methods generally undercover.
This is expected and mirrors the observations in \cref{sec:coverage_per_level}.

\begin{figure}[H]
    \centering
    \includegraphics[width=\linewidth]{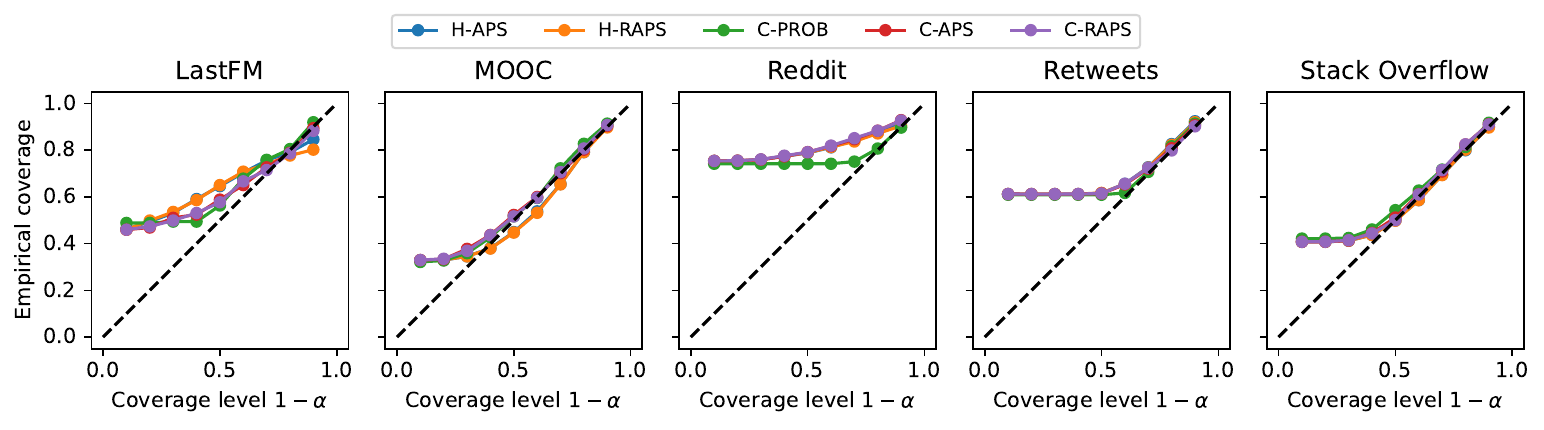}
    \caption{Empirical marginal coverage for different coverage levels for methods that produce a prediction region for the mark with the CLNM model.}
    \label{fig:coverage_per_level/CLNM/mark}
\end{figure}

\cref{fig:coverage_per_level/CLNM/mark} shows that all methods, either heuristic or conformal, overcover for small coverage levels, while coverage is attained for high coverage levels.
The reason is that all methods that generate a prediction set for the mark guarantee that prediction sets are not empty by always adding the class with the highest probability, as presented in \cref{sec:predmethods}.
We do not observe overcoverage for high coverage levels because the class with highest probability will almost always be included.
However, for low coverage levels, prediction sets that would normally be empty now include the mark with the highest probability, which leads to increased coverage.

\section{Additional example}
\label{sec:additional_example}

\begin{figure}[H]
    \centering
    \includegraphics[width=\linewidth]{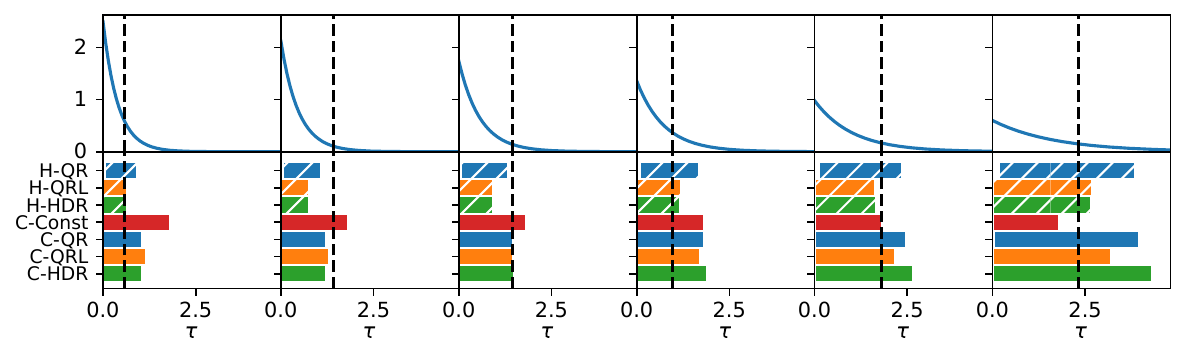}
    \caption{Toy example with $\alpha = 0.2$ and a calibration dataset of 6 data points.}
    \label{fig:conformal_intervals2}
\end{figure}

In \cref{fig:conformal_intervals} in \cref{sec:results_time}, we presented an example illustrating prediction regions for the time for seven methods with $\alpha = 0.5$.
For completeness, we provide an additional toy example with $\alpha = 0.2$ and a calibration dataset of 6 data points in \cref{fig:conformal_intervals2}.
As in \cref{fig:conformal_intervals}, the heuristic methods undercover, achieving a maximum coverage of $4/6$, which is less than the desired coverage of $0.8$.
Notably, H-QRL and H-HDR produce exactly the same prediction regions because the densities are decreasing in this case.
Conformal methods adjust the predictions regions to achieve coverage in at least five out of six cases.
Similarly to \cref{fig:conformal_intervals}, C-HDR generates larger regions on average than other conformal methods despite H-HDR always producing shorter or equivalent lengths compared to H-QR and H-QRL.

\section{Partitioning for conditional coverage}
\label{sec:cond_coverage}

\begin{figure}[H]
  \centering
  \begin{subfigure}[b]{0.48\linewidth}
    \includegraphics[width=\linewidth]{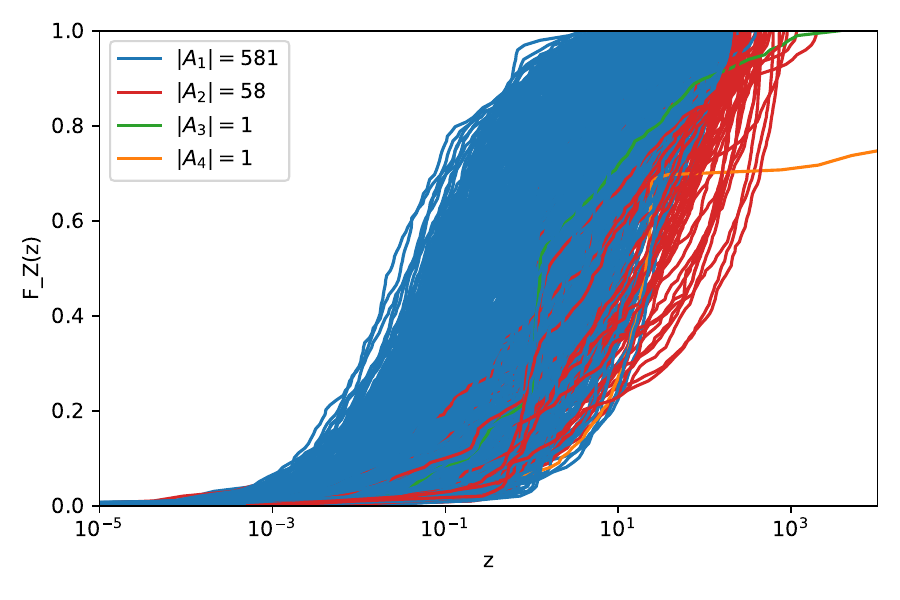}
    \caption{Partitions obtained with the distance $d_Z$.}
    \label{fig:conditional/partition_reddit_with_z_4}
  \end{subfigure}
  \hfill
  \begin{subfigure}[b]{0.48\linewidth}
    \includegraphics[width=\linewidth]{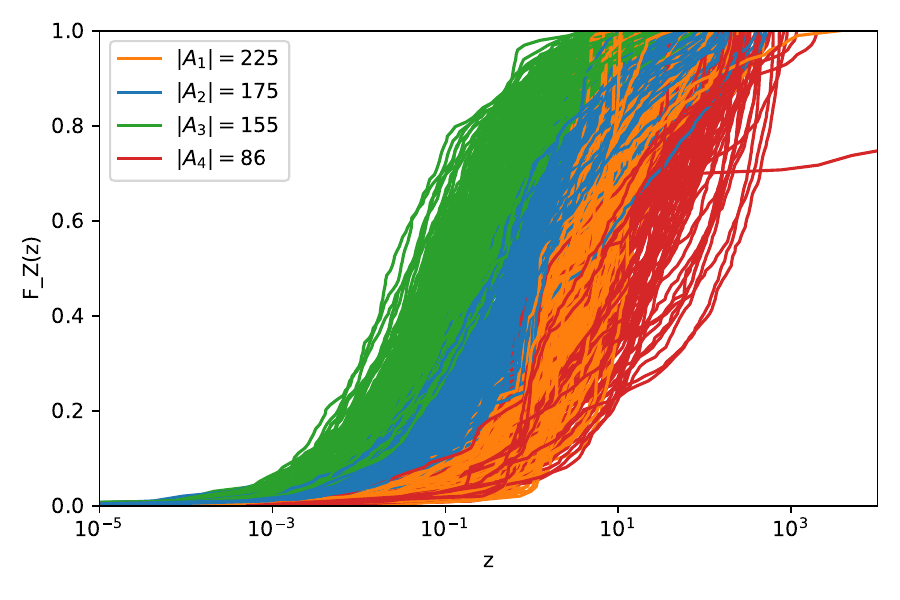}
    \caption{Partitions obtained with the distance $d_{\log Z}$.}
    \label{fig:conditional/partition_reddit_with_logz_4}
  \end{subfigure}
  \caption{The two subfigures show the CDF of $Z$ for all instances in the calibration dataset of the Reddit dataset. The colors determine the partition in which an instance falls according to the distance $d_Z$ on the left and $d_{\log Z}$ on the right, as introduced in \cref{sec:metrics}. The legend denotes the size of each cluster of the partition.}
  \label{fig:conditional/partition_reddit}
\end{figure}

In this section, we elaborate on the choice of distance function to create the partitions for the metric CCE introduced in \cref{sec:metrics}.
On \cref{fig:conditional/partition_reddit}, we show the CDF of $Z$ for all instances in the calibration dataset of the Reddit dataset.
As shown in \cref{fig:conditional/partition_reddit_with_z_4}, instances where the distributions of $Z$ have the longest tails exhibit extreme distances from other distributions, resulting into their isolation into small clusters.
\cref{fig:conditional/partition_reddit_with_logz_4} shows that, by instead focusing on the random variable $\log Z$, we achieve more balanced cluster sizes, which is crucial to have an accurate estimation of coverage within each partition.

\section{Computation time}

In \cref{table:compute_time}, we present the computation times for evaluating the scores and regions across each conformal method utilized in our experiments. Except for \texttt{C-Const}, which incurs a minimal computation time primarily due to data loading, the computational demands of the other methods are of a comparable magnitude.

For all methods excluding \texttt{C-Const}, computation time is primarily governed by the calculation of the CDF of the time and the joint PDF of the time and mark. Specifically, the most resource-intensive tasks involve computing the quantiles of the time or generating samples from the time distribution, as these operations require inverting the CDF using the bisection method, typically necessitating around 30 evaluations.

In the cases of \texttt{C-QR} and \texttt{C-QRL}, the computation time is dominated by computing the quantiles of the time. For \texttt{C-HDR} (time and joint), \texttt{C-PROB}, \texttt{C-APS}, and \texttt{C-RAPS}, the primary computational load comes from generating time samples. More specifically, for \texttt{C-HDR}, these samples are needed to compute HPD values.
For \texttt{C-PROB}, \texttt{C-APS}, and \texttt{C-RAPS}, computing the marginal PMF of the mark relative to the time involves averaging the joint density over the time across these samples.

    \begin{table}[h]
    \footnotesize
    \centering
    \begin{tabular}{ll|rrrr|rrr|rrr}
        \toprule
         & & \multicolumn{4}{l}{\textbf{Time}} & \multicolumn{3}{l}{\textbf{Mark}} & \multicolumn{3}{l}{\textbf{Joint}} \\
        \textbf{Dataset} & \textbf{\makecell{Compute\\type}} & C-Const & C-QR & C-QRL & C-HDR & C-PROB & C-APS & C-RAPS & \makecell{C-QRL\\-RAPS} & \makecell{C-HDR\\-RAPS} & C-HDR \\
        \midrule
        \multirow[c]{2}{*}{LastFM} & Score & 0.07 & 15.01 & 8.10 & 8.63 & 8.40 & 8.40 & 8.36 & 16.56 & 17.11 & 8.71 \\
         & Region & 0.10 & 9.97 & 5.42 & 10.51 & 5.61 & 5.62 & 5.61 & 10.34 & 15.55 & 11.70 \\
        \multirow[c]{2}{*}{MOOC} & Score & 0.38 & 93.17 & 47.64 & 51.70 & 49.68 & 49.33 & 49.71 & 96.90 & 102.07 & 51.88 \\
         & Region & 0.75 & 62.32 & 32.06 & 66.46 & 33.30 & 33.36 & 33.58 & 64.24 & 99.40 & 74.98 \\
        \multirow[c]{2}{*}{Reddit} & Score & 0.25 & 56.47 & 29.04 & 31.51 & 30.34 & 30.23 & 30.19 & 59.48 & 61.97 & 31.93 \\
         & Region & 0.48 & 38.31 & 19.84 & 40.60 & 20.72 & 20.78 & 20.73 & 39.48 & 60.66 & 45.68 \\
        \multirow[c]{2}{*}{Retweets} & Score & 0.60 & 159.38 & 80.30 & 85.84 & 84.54 & 84.27 & 84.13 & 165.46 & 171.16 & 86.88 \\
         & Region & 0.56 & 106.49 & 53.93 & 112.55 & 56.77 & 56.43 & 56.60 & 110.03 & 169.60 & 114.00 \\
        \multirow[c]{2}{*}{\makecell{Stack\\Overflow}} & Score & 0.45 & 105.68 & 53.33 & 57.20 & 55.86 & 55.92 & 55.76 & 113.25 & 112.95 & 57.55 \\
         & Region & 0.58 & 71.12 & 35.91 & 75.17 & 37.90 & 38.14 & 37.88 & 79.21 & 111.91 & 79.11 \\
        \bottomrule
    \end{tabular}

    \caption{Time to compute the scores and regions for all considered conformal methods on real world datasets using the CLNM model, averaged over 5 runs, in seconds.}
    \label{table:compute_time}
    \end{table}

\section{Additional examples of joint prediction regions}

\begin{figure}[H]
    \centering
    \includegraphics[width=0.9\textwidth]{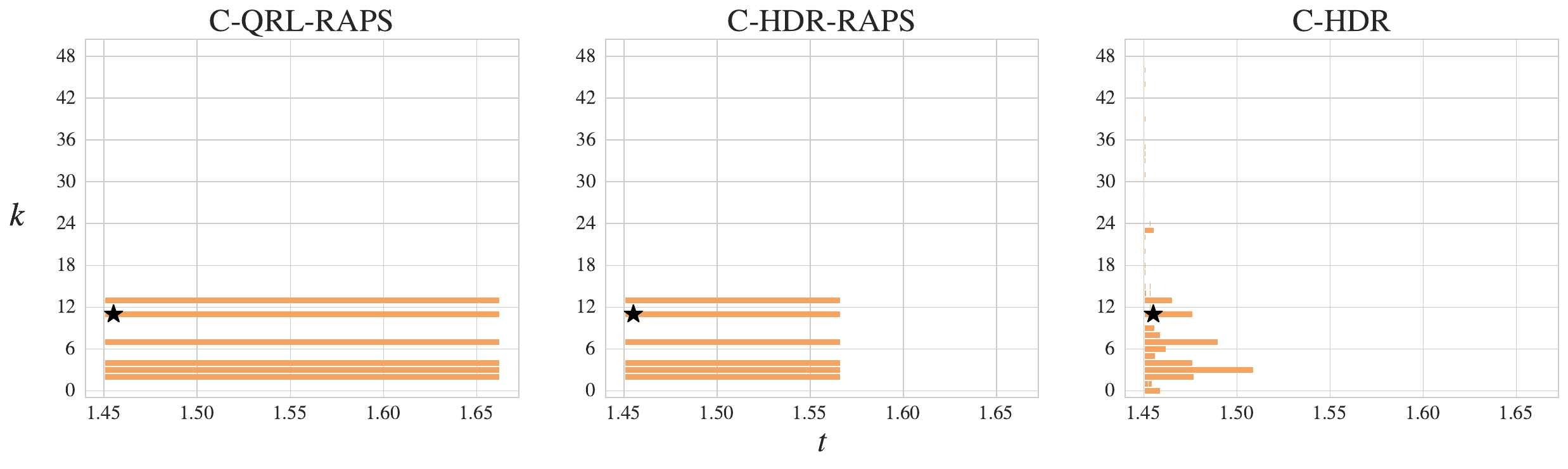}
    \caption{Example of joint prediction regions generated for the last event of a test sequence in the MOOC dataset.}
    \label{fig:intervals_mooc}
\end{figure}

\begin{figure}[H]
    \centering
    \includegraphics[width=0.9\textwidth]{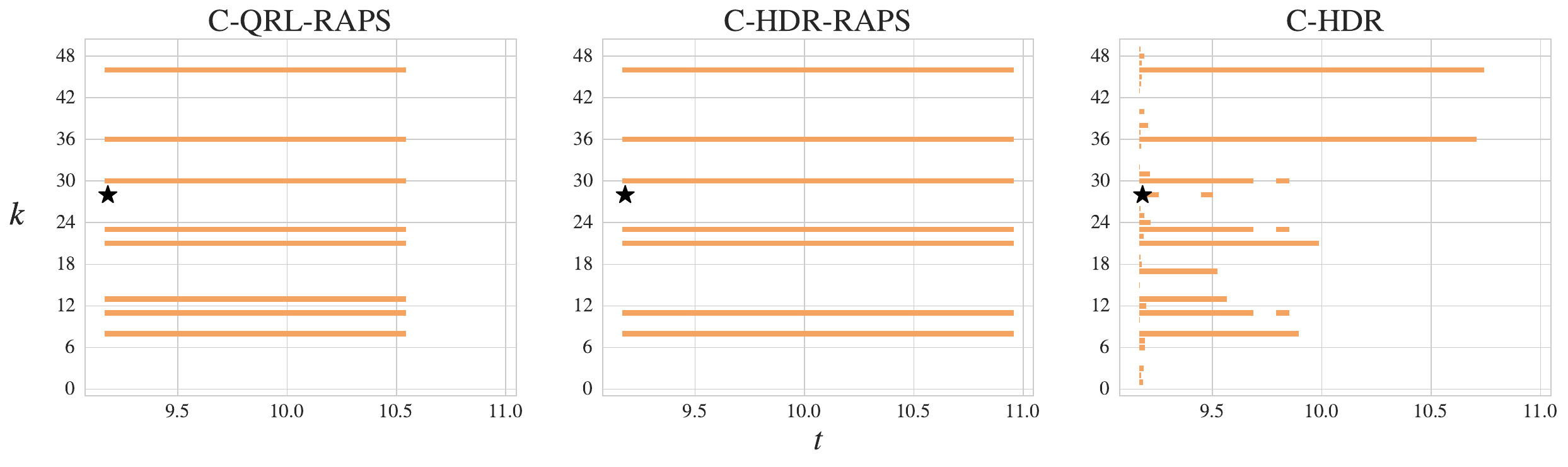}
    \caption{Example of joint prediction regions generated for the last event of a test sequence in the Reddit dataset.}
    \label{fig:intervals_reddit}
\end{figure}

\begin{figure}[H]
    \centering
    \includegraphics[width=0.9\textwidth]{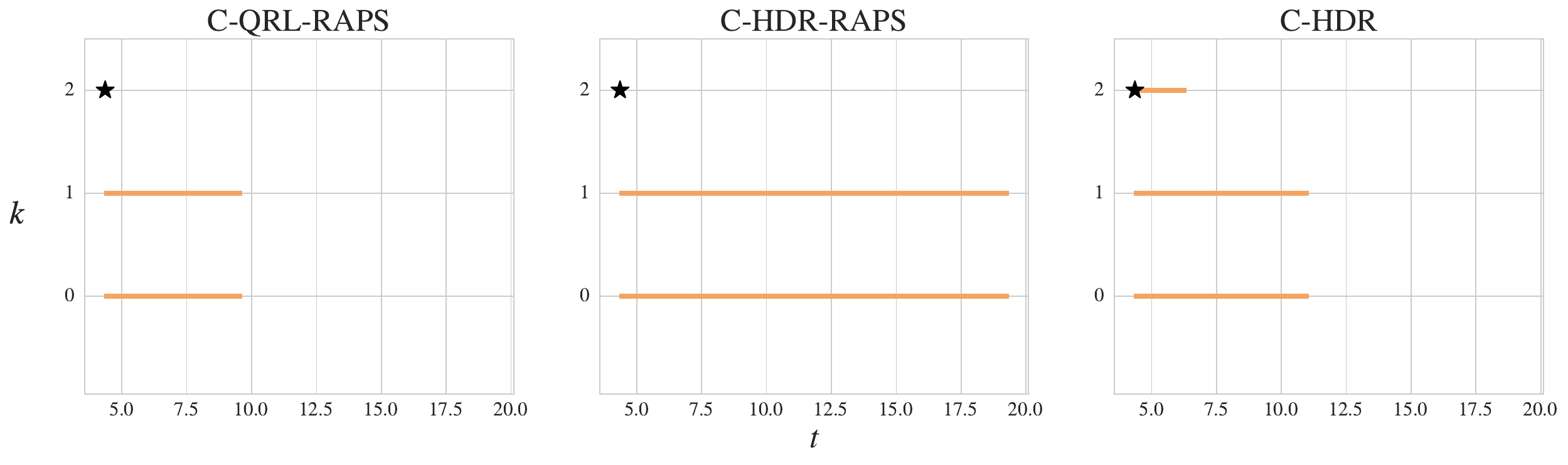}
    \caption{Example of joint prediction regions generated for the last event of a test sequence in the Retweets dataset.}
    \label{fig:intervals_retweets}
\end{figure}

\begin{figure}[H]
    \centering
    \includegraphics[width=0.9\textwidth]{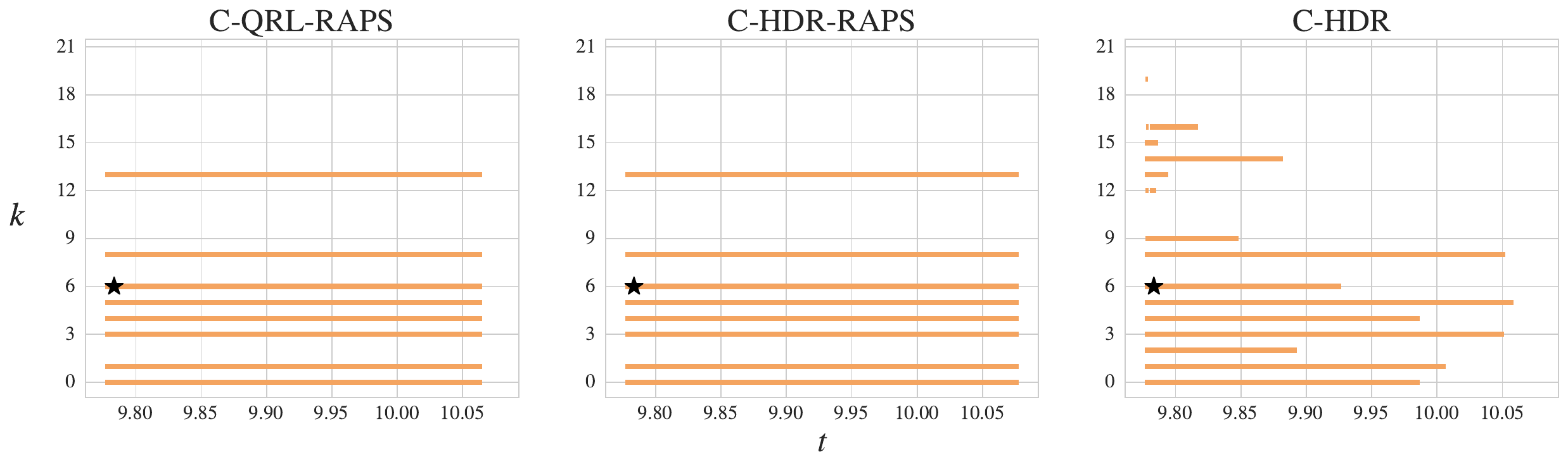}
    \caption{Example of joint prediction regions generated for the last event of a test sequence in the Stack Overflow dataset.}
    \label{fig:intervals_so}
\end{figure}

\cref{fig:intervals_mooc,fig:intervals_retweets,fig:intervals_reddit,fig:intervals_so} present additional predictions regions generated by conformal methods on the datasets MOOC, Reddit, Retweets and Stack Overflow, respectively.
We observe that C-HDR generally selects more marks than C-QRL-RAPS and C-HDR-RAPS. However, the joint region produced by C-HDR is usually smaller.

\end{document}